\documentclass{article}

\usepackage{algorithm}
\usepackage{algpseudocode}
\usepackage{subcaption}
\usepackage{adjustbox}
\usepackage{amsmath}
\usepackage{amssymb}
\usepackage{mathtools}
\usepackage{amsthm}
\usepackage{diagbox}
\usepackage{tikz}
\usetikzlibrary{arrows, arrows.meta, positioning}
\usepackage{wrapfig}
\usepackage{float}
\usepackage{caption}
\usepackage{enumitem}
\usepackage{booktabs}

\DeclareMathOperator*{\argmax}{arg\,max}
\DeclareMathOperator*{\argmin}{arg\,min}
\DeclareMathOperator*{\softmax}{softmax}

\newtheorem{theorem}{Theorem}[section]
\newtheorem{proposition}[theorem]{Proposition}
\newtheorem{lemma}[theorem]{Lemma}
\newtheorem{corollary}[theorem]{Corollary}

\newtheorem{assumption}[theorem]{Assumption}
\newtheorem{remark}[theorem]{Remark}

\def\tcr{\textcolor{red}}	

\usepackage[numbers]{natbib}
\usepackage[final]{neurips_2025}

\usepackage[utf8]{inputenc} 
\usepackage[T1]{fontenc}    
\usepackage{hyperref}       
\usepackage{url}            
\usepackage{booktabs}       
\usepackage{amsfonts}       
\usepackage{nicefrac}       
\usepackage{microtype}      
\usepackage{xcolor,verbatim}         

\title{Multi-Objective Reinforcement Learning with Max-Min Criterion:  A Game-Theoretic Approach}

\author{%
  Woohyeon Byeon\textsuperscript{1}\\
  \And
    Giseung Park\textsuperscript{2}\\
  \And
   Jongseong Chae\textsuperscript{1}\\
  \And
   Amir Leshem\textsuperscript{3}\\
  \And
   Youngchul Sung\textsuperscript{1}*
   \\
}

\begin{document}

\maketitle

\footnotetext[1]{School of Electrical Engineering, KAIST, Republic of Korea, $^2$University of Toronto Robotics Institute, Canada, $^3$Faculty of Engineering, Bar-Ilan University,  Israel. *Correspondence to: Youngchul Sung <ycsung@kaist.ac.kr>.}

\begin{abstract}
In this paper, we propose a provably convergent and practical framework for multi-objective reinforcement learning with max-min criterion. From a game-theoretic perspective, we reformulate max-min multi-objective reinforcement learning as a two-player zero-sum regularized continuous game and introduce an efficient algorithm based on mirror descent. Our approach simplifies the policy update while ensuring global last-iterate convergence. 
We provide a comprehensive theoretical analysis on our algorithm, including iteration complexity under both exact and approximate policy evaluations, as well as sample complexity bounds. 
To further enhance performance, we modify the proposed algorithm with adaptive regularization.
Our experiments demonstrate the convergence behavior of the proposed algorithm in tabular settings, and our implementation for deep reinforcement learning significantly outperforms  previous baselines in many MORL environments. \\

\end{abstract}

\section{Introduction}
\label{introduction}
Reinforcement Learning (RL) focuses on  sequential decision-making in Markov Decision Processes (MDPs), which have been extensively studied both theoretically and practically.
However, in several practical decision-making problems such as autonomous vehicles, resource allocation and communication, multiple objectives should be simultaneously optimized instead of a conventional single objective \cite{autonomous_vehicles_morl_1,autonomous_vehicles_morl_2,resource_allocation_morl_1,resource_allocation_morl_2}. 
These scenarios motivate the development of Multi-Objective Reinforcement Learning (MORL), an extension of RL in which the reward function yields a vector rather than a scalar.
In recent years, MORL has seen significant progresses \cite{yang,mdqn,pareto_convexity,park,mo_gym}.
A widely-adopted approach to MORL is the 
 utility-based approach  \cite{onestate,practical_guide_morl}, which  is formulated as follows: for a utility function $u$ (also called scalarization function), solve $\max_{\pi} u(V^\pi_1,\ldots,V^\pi_K)$, where $\pi$ is a policy and $V^\pi_k$ is the value function of the $k$-th objective.
As long as the utility function is non-decreasing in the sense of Pareto dominance, the optimal policy for $u(V^\pi_1,\ldots,V^\pi_K)$ is also Pareto optimal \cite{onestate}.

Although many previous works on MORL considered the weighted sum utility function \cite{yang, pd-morl, pareto_convexity}, the weighted sum utility function is not the desired design metric in many cases, especially when considering the application of MORL to resource allocation where fairness is an important issue. 
For example, consider scheduling of cloud computing resources, in which a computing job is typically divided into multiple subtasks and each subtask is performed by a different resource and completed subtasks are combined to finish the overall job~\cite{resource_allocation_1,resource_allocation_2}. In this case, the overall time for finishing the job is determined by the maximum of subtask completion times.  Consider another example of traffic signal light control at an intersection combining multiple roads. One can consider the waiting time at each road and  may want to minimize the maximum of the waiting times of all roads for fairness on the drivers at all roads rather than to minimize the sum of the waiting times.  Hence, the min-max (or equivalently max-min with negation of objectives) criterion naturally arises in many real-world resource allocation problems as well as  other application 
 domains \citep{intro_ecology, intro_tsc,equalizer_rule, network_morl_1, network_morl_2,hanMME, primal_dual_4}, where fairness is important. (Please see Appendix~\ref{append:fairness} for comparison with other scalarizartion criteria including proportional fairness.)

In this paper, we consider this \textbf{max-min MORL}, where the utility function is the minimum function: $u(x_1,\ldots,x_K)=\min_{k=1,\ldots,K} x_k$ and the resulting problem is given by  
 $\max_{\pi} \min_{k=1,\ldots,K} V^\pi_k$.  With the increasing relevance of max-min MORL,  \citet{park} recently  proposed a \textbf{model-free}  algorithm for max-min MORL through a convex formulation to circumvent the non-differentiability of min operation. Although their work provides a model-free algorithm to solve exact max-min MORL  in contrast to previous approximation methods \cite{ESR_example_1,ESR_nonlinear,mdqn} or model-based approaches \cite{cousins_model_based} to max-min MORL, their algorithm  suffers from high memory requirement and computational costs and only guarantees average-iterate convergence. 
In this paper, we further advance  max-min MORL and propose a  fast algorithm for max-min MORL with last-iterate convergence and significantly improved efficiency in computing time and required memory by reformulating max-min MORL as a {\em two-player zero-sum game}.

Our contributions are summarized as follows:

$\bullet$  We propose a single-loop algorithm for entropy-regularized max-min MORL by leveraging appropriate regularization to derive a closed-form update.

$\bullet$ Our algorithm can be viewed as a primal-dual algorithm  and we provide a theoretical analysis of the proposed algorithm in the  tabular case, including global last-iterate convergence, iteration complexity, and sample complexity.

$\bullet$ We demonstrate convergence behavior through numerical simulations in the tabular case and show that our deep reinforcement learning implementation significantly outperforms existing baselines in several MORL environments including a realistic traffic signal control task.

\section{Background}

\paragraph{Multi-objective Markov decision process}

A multi-objective Markov decision process (MOMDP) extends a standard MDP by incorporating a vector-valued reward function. 
Formally, a MOMDP is defined as $\langle S,A,P,\mu,\mathbf{r},\gamma\rangle$ with state space $S$, action space $A$, transition $P:S\times A\to \Delta(S)$, where $\Delta(X)$ denotes the set of probability distributions on set $X$, initial state distribution $\mu\in\Delta(S)$ and discount factor $\gamma\in(0,1)$. Unlike single-objective MDPs, the reward function $\mathbf{r}: S\times A\to \mathbb{R}^K$ is vector-valued  with $\mathbf{r}(s,a)=(r_1(s,a),\ldots,r_K(s,a))$, where $K$ is the number of objectives. An agent sequentially interacts with an environment by taking an action from its stationary policy $\pi: S\to \Delta(A)$.
Unlike single-objective settings, the agent seeks to optimize all components of the reward vector simultaneously in some manner.

The multi-objective state value of a policy $\pi$ is defined as 
$\mathbf{V}^\pi = \mathbb{E}_{\mu,\pi}\left[\sum_{t=0}^\infty \gamma^t \mathbf{r}(s_t,a_t)\right] \in \mathbb{R}^K$, 
and $V^\pi_k$ denotes the $k$-th element of $\mathbf{V}^\pi$.
The multi-objective state value with entropy regularization is defined as 
$\mathbf{V}^\pi_\tau=\mathbb{E}_{\mu,\pi}\left[\sum_t \gamma^t \left(\mathbf{r}(s_t,a_t)-\tau\log\pi(a_t|s_t)\mathbf{1}_K\right)\right]\in\mathbb{R}^K,$
where $\tau$ is a regularization coefficient,
$\mathbf{1}_K$ is a $K$-dimensional vector all of which entries are 1,
and $V^\pi_{k,\tau}$ denotes the $k$-th element of vector $\mathbf{V}^\pi_\tau$.
Throughout this paper, $w=(w(1),\ldots,w(K))\in\Delta^K$ denotes a weight vector, where $\Delta^K$ is the $(K-1)$-simplex, i.e., $\Delta^K = \{ w| w(i) \ge 0 ~\forall i,~ w(1) + \cdots +w(K)=1\}$.
For a weight vector $w$, we define the weighted value as $V^\pi_{w}=\langle w, \mathbf{V}^\pi\rangle$ and the weighted soft value as $V^\pi_{w,\tau}=\langle w, \mathbf{V}^\pi_\tau\rangle$, where $\langle\cdot,\cdot\rangle$ denotes inner product. The soft values can be decomposed as 
$V^\pi_{k,\tau} = V_k^\pi + \tau \tilde{H}(\pi),~\forall k$ and $V^\pi_{w,\tau} = V_w^\pi + \tau \tilde{H}(\pi)$,
where
$\tilde{H}(\pi)=\mathbb{E}_{\mu,\pi}\left[-\sum_t \gamma^t \log\pi(a_t|s_t)\right]$ is the expected cumulative  entropy of policy $\pi$.  
A summary of notations is provided in Appendix~\ref{glossary}.




\textbf{Max-min MORL} ~~Since the value function for MORL is  vector-valued, optimizing the value function $\mathbf{V}^\pi$ over $\pi$ cannot be defined as in single-objective RL. Thus, \textbf{max-min MORL} considers the following max-min optimization:
$  \max_\pi \min_{k=1,\ldots,K} V_k^\pi$.
However, it is known that max-min MORL may yield the indeterminacy of its solution and entropy regularization  resolves the possible indeterminacy of max-min MORL solution \citep{park}. So, in this paper, instead of max-min MORL, we can consider the following  \textbf{entropy-regularized max-min MORL} problem:
\begin{equation}\label{maxmin_ent_target_problem}
    \max_\pi \min_{k=1,\ldots,K} [~\underbrace{V_k^\pi +\tau \tilde{H}(\pi)}_{=V_{k,\tau}^\pi} ~]
\end{equation}
which converges to max-mix MORL as $\tau \rightarrow 0$. It can be shown that the value gap between the entropy unregularized and regularized problems is proportional to the coefficent $\tau$.  
Due to the non-linearity of the $\min$ operator, max-min MORL or entropy-regularized max-min MORL cannot be  solved directly using standard RL methods, requiring a specialized framework such as \cite{park} or approximate approaches  \cite{ESR_example_1,ESR_nonlinear,mdqn}.

\textbf{Mirror descent and natural policy gradient}  ~Mirror descent (MD)~\cite{mirror_descent} is a general optimization framework that subsumes popular algorithms such as gradient descent and the multiplicative weights update~\cite{mwu}. It has been widely applied in learning dynamics across various games~\cite{apmd,learning_in_games_1,learning_in_games_2,learning_in_games_3}.
Formally, the update rule for MD can be expressed in primal domain as 
\begin{equation}
    x_{t+1} = \max_{x\in \mathcal{X}} ~\lambda \langle \nabla f(x_t), x\rangle - D_\psi(x,x_t),
\end{equation}
where  $f(x)$ is the objective function,  $\lambda$ is a step size, and 
$D_\psi$ is the Bregman divergence induced by a continuously differentiable and strictly convex function $\psi$, defined as
$D_\psi(x,y) = \psi(x) - \psi(y) - \langle \nabla\psi(y), x-y\rangle,~\forall x,y\in \mathcal{X}$. MD  basically  implements  steepest descent in a manifold of which Riemannian metric is given by the Hessian of $D_\psi$, capturing relevant geometry of the optimization variable space. 
In reinforcement learning, the natural policy gradient (NPG) can be interpreted as a MD variant for MDPs with the Fisher information metric \cite{linear_convergence_npg, mirror_descent,trpo}.

\section{The Proposed Method}\label{section_method}

Our goal in this paper is to solve the general problem \eqref{maxmin_ent_target_problem}. We here develop a single-loop algorithm by interpreting entropy-regularized max-min MORL (\ref{maxmin_ent_target_problem}) as learning a Nash equilibrium in a {\em two-player zero-sum game} and exploiting theory of learning in games.

\subsection{Max-min versus min-max}\label{subsection:starting_point}
In this section, we derive Theorem~\ref{thm:saddle_is_maxminmorl}, which serves as the initial step for our reformulation of the entropy-regularized max-min MORL problem~\eqref{maxmin_ent_target_problem} as a two-player zero-sum game.

Our key idea begins with the  fact that the entropy-regularized max-min MORL problem  \eqref{maxmin_ent_target_problem}
is equivalent to the following problem \citep{park}:
\begin{equation}\label{eq:P2'_again}
    \text{Minimization Reformulation}: \min_{w \in \Delta^K} V^*_{w,\tau}
\end{equation}
where $V^*_{w,\tau}$ is the optimal soft value under the scalarized reward $\langle w, \mathbf{r}\rangle$.
Here, the equivalence of problems means that both problems have the same optimal value and an optimal solution of one problem yields an optimal solution of the other. The equivalence was first shown in \citep{park} and a brief explanation of this equivalence is provided in Appendix~\ref{append:prior_work} for readers' convenience.

The next enabling step is that the problem  \eqref{maxmin_ent_target_problem} involving  minimization over discrete variable $k$ can be converted to minimization over continuous variable $w\in\Delta^K$ as follows \cite{boyd2004convex}:
\begin{equation}\label{target_maxmin}
    \max_\pi \min_{k=1,\ldots,K} V_{k,\tau}^\pi 
    = \max_\pi \min_{w\in\Delta^K}  \langle w, \mathbf{V}_\tau^\pi \rangle,
\end{equation}
where $\langle w, \mathbf{V}_\tau^\pi \rangle = \sum_{k=1}^K w(k) V_{k,\tau}^\pi$, and $w(k)$ is the $k$-th element of weight vector $w$. 
This can easily be seen by considering the case in which we have a single minimum $V_{k',\tau}^\pi$ for some $k'$. In this case, the minimizing $w$ is given by $(0,\cdots,0,1,0,\cdots,0)$ with 1 at the $k'$th position. 
Note that while the two problems in \eqref{target_maxmin}  yield the same optimal value, the minimizers are not always in one-to-one correspondence. When the minimum is attained at multiple $k$ indices, the corresponding optimal weights $w$ form a convex combination of those indices and lie on the boundary of the simplex. 
Theorem~\ref{thm:saddle_is_maxminmorl} below formally bridges this gap by proving equivalence of optimal solutions.

Furthermore, due to $ V^*_{w,\tau} \triangleq \max_\pi \langle w, \mathbf{V}_\tau^\pi\rangle$ by its definition, 
the problem  (\ref{eq:P2'_again}) can be rewritten as 
\begin{equation}\label{target_minmax}
    \min_{w\in\Delta^K} \max_\pi ~\langle w, \mathbf{V}_\tau^\pi\rangle.
\end{equation}
Combining these relations, we have the following equivalence between the max-min problem  
(\ref{target_maxmin}) (=\eqref{maxmin_ent_target_problem}) and the min-max  problem (\ref{target_minmax}) (=\eqref{eq:P2'_again}), i.e., 
\begin{equation}\label{maxmin_equals_minmax}
    \max_\pi \min_{w\in\Delta^K} \langle w, \mathbf{V}_\tau^\pi\rangle 
    = \min_{w\in\Delta^K} \max_\pi \langle w, \mathbf{V}_\tau^\pi\rangle .
\end{equation}
Note that the max-min value is not equal to the min-max value in general. Since  value functions in RL are non-concave in policy even with direct parameterization \cite{agarwal}, the equality (\ref{maxmin_equals_minmax}) cannot be directly induced from the minimax theorem \cite{rockafellar1997convex}.

From this special relation that the max-min value equals the min-max value in our case, the existence of a saddle point is guaranteed in Proposition~\ref{prop:maxmin_equals_minmax_implies_saddle} in Appendix~\ref{append:existence_saddle}. Once the existence of a saddle point is guaranteed, Theorem~\ref{thm:saddle_is_maxminmorl} establishes the equivalence of this saddle point (i.e., Nash equilibrium) in (\ref{maxmin_equals_minmax}) and the solution of entropy-regularized max-min MORL \eqref{maxmin_ent_target_problem}; i.e., it suffices to find a Nash equilibrium in (\ref{maxmin_equals_minmax}) to solve entropy-regularized max-min MORL.

\begin{theorem}\label{thm:saddle_is_maxminmorl}
    Let $(\Bar{\pi},\Bar{w})$ be a saddle point in (\ref{maxmin_equals_minmax}). Then, $\Bar{\pi}$ is a solution to the entropy-regularized max-min MORL \eqref{maxmin_ent_target_problem}.
\end{theorem}
{\em Proof:} See Appendix~\ref{section:saddle_pf}.

Now, based on Theorem~\ref{thm:saddle_is_maxminmorl}, we will reformulate 
entropy-regularized max-min MORL as learning a Nash equilibrium achieving equality of \eqref{maxmin_equals_minmax} in the two-player zero-sum game whose payoff functions are given by $\langle w, \mathbf{V}_\tau^\pi\rangle$ and $-\langle w, \mathbf{V}_\tau^\pi\rangle$ for efficient algorithm construction.

\subsection{Two-plaer zero-sum regularized continuous game formulation}\label{subsection_2p0s}


We first define a two-player zero-sum continuous game $\mathcal{G}$ as follows.
There are two players; $Learner$ and $Adversary$. 
The player $Learner$ corresponds to the RL agent in the MOMDP who learns the policy parameter $\theta$. 
We define the strategy space of $Learner$ as $X_{Learner} = \Theta\subset \mathbb{R}^d$, which is the space of policy parameters in the MOMDP. In other words, $Learner$ takes continuous strategy $\theta\in\Theta$, which indicates the parameter of policy. 
The player $Adversary$ has the strategy space $X_{adv}=\Delta^K$ and takes continuous strategy $w\in\Delta^K$.
The utility functions of the two players are defined as
\begin{equation}
    u_{Learner}^\mathcal{G}(\theta,w) = \langle w, \mathbf{V}_\tau^{\pi_\theta}\rangle = -u_{Adv}^\mathcal{G}(\theta,w).
\end{equation}
In this game, $Learner$ tries to maximize its utility $\langle w, \mathbf{V}_\tau^{\pi_\theta}\rangle$ and $Adversary$ tries to maximize its utility $-\langle w, \mathbf{V}_\tau^{\pi_\theta}\rangle$, i.e., minimize $\langle w, \mathbf{V}_\tau^{\pi_\theta}\rangle$. In a NE of this game, max-min and min-max values become the same and \eqref{maxmin_equals_minmax} is achieved to yield our policy solution.

Instead of learning a NE in $\mathcal{G}$, however,  we solve the following  regularized game $\mathcal{RG}$, which is judiciously designed to our advantage for efficient algorithm construction. 
The regularized game $\mathcal{RG}$ has the same players and strategy spaces as the original game $\mathcal{G}$. Instead, $\mathcal{RG}$ has utility functions regularized from the utility functions of $\mathcal{G}$. 
Formally, the utility functions of $\mathcal{RG}$  are defined as 
\begin{equation}  \label{eq:RGutil}
    u_{Learner}^\mathcal{RG}(\theta,w) = \underbrace{\langle w, \mathbf{V}^{\pi_\theta}\rangle + \tau\tilde{H}(\pi_\theta)}_{=\langle w, \mathbf{V}_\tau^{\pi_\theta}\rangle} - \tau_w H(w) = - u_{Adv}^\mathcal{RG}(\theta,w),
\end{equation}
where  $\tilde{H}(\pi)=\mathbb{E}_{\mu,\pi}\left[-\sum_t \gamma^t \log\pi(a_t|s_t)\right]$ as already defined,  $H(w) := -\sum_{k=1}^K w(k)\log w(k)$,  and $\tau$ and $\tau_w$ are positive coefficients.

Now, let us explain why we solve the new regularized game $\mathcal{RG}$ instead of solving $\mathcal{G}$.
For $Learner$, the main reason for adding entropy regularization $\tilde{H}(\pi_\theta)$ is 
to avoid the indeterminacy problem as already explained \citep{park}. 
In short, stationary deterministic policies are insufficient for max-min MORL since there exists an MOMDP that has only a stochastic optimal policy for max-min MORL \cite{onestate}.  

For $Adversary$, adding $-H(w)$ regularization, $Adversary$  minimizes $\langle w, \mathbf{V}_\tau^{\pi_\theta}\rangle - \tau_w H(w)$. That is, $Adversary$  minimizes $\langle w, \mathbf{V}_\tau^{\pi_\theta}\rangle$ while increasing the entropy $H(w)$. Note that $Learner$  maximizes $\min_{w\in\Delta^K} \langle w, \mathbf{V}_\tau^{\pi_\theta}\rangle$ with its policy $\pi_\theta$. Without $-H(w)$ regularization, $w$ achieving $\min_{w\in\Delta^K} \langle w, \mathbf{V}_\tau^\pi\rangle$ would be one-hot vector focusing only on the worst dimension of $\mathbf{V}_\tau^{\pi_\theta}$. However, with  $-H(w)$ regularization,  the elements of the  $w$ vector will be spread out to incorporate multiple objective dimensions for $Learner$'s optimization. This speeds up learning and 
we can achieve last-iterate convergence rather than average-iterate convergence, which will be proven in  Section~\ref{section_tabular}.
In addition, unlike other strongly convex regularizations such as squared $l_2$-norm, our choice of negative entropy regularization enables a {\em closed-form solution} for the MD update of $w$ shown in (\ref{eq:md_w}) in Section~\ref{subsection_proposed_algorithm}.
This closed-form update due to our $-H(w)$ regularization  significantly simplifies the optimization process and reduces computational overhead.




\subsection{The ERAM algorithm: Entropy-regularized adversary for max-min MORL}\label{subsection_proposed_algorithm}

We propose \textbf{ERAM} (Entropy-Regularized Adversary for Max-min MORL), an efficient  algorithm to solve  max-min MORL via a game-theoretic perspective. Based on the reformulation of the problem as a two-player zero-sum game, we leverage equilibrium learning methods and introduce a variant of the MD algorithm to find a Nash equilibrium in $\mathcal{RG}$.

First, we update the policy parameter with the MD objective on $u_{Learner}^\mathcal{RG}$ with step size $\eta$ as follows: 
\begin{align}
    \theta_{t+1} = & \argmax_\theta~ \{\eta\langle \nabla_\theta~ u_{Learner}^\mathcal{RG}(\theta_t,w_t), ~\theta\rangle - D_\psi(\theta,\theta_t)\}.\label{eq:md_theta}
\end{align}
Note that the objective in \eqref{eq:md_theta} can be viewed as the Lagrangian of the optimization of the linear approximation of $Learner$'s utility around current $\theta_t$ (i.e., $u_{Learner}^\mathcal{RG}(\theta_t,w_t) + \langle \nabla_\theta~ u_{Learner}^\mathcal{RG}(\theta_t,w_t), ~\theta-\theta_t\rangle$) under a constraint on $D_\psi(\theta,\theta_t)$.  
Here, we choose the convex function $\psi$ for the Bregman divergence to be  the negative Shannon entropy to    make the Bregman divergence  the  KL-divergence, which yields the Fisher information matrix (FIM) as Hessian matrix and 
is relevant to statistical manifolds: That is, $\pi_\theta$ is a probablity distribution with coordinate $\theta$, and $\{\pi_\theta\}$ forms a statistical manifold where FIM is an invariant metric \cite{AmariBook}.  Then, 
the optimization with 
such conservatism in RL leads to natural policy gradient (NPG) \cite{AmariBook,mirror_descent,linear_convergence_npg,trpo}.
The NPG update rule is given by \cite{npg_ent}
\begin{equation}
    \theta_{t+1} = \theta_t + \eta F^\dagger(\theta_t)
    \sum_{k=1}^K w_t(k) \nabla_\theta V^{\pi_{\theta_t}}_{k,\tau},
\end{equation}
where 
$F(\theta)=\mathbb{E}_{(s,a)\sim d^{\pi_\theta}}[\nabla_\theta\log\pi_\theta(a|s)\nabla_\theta\log\pi_\theta(a|s)^T]$ is the FIM, $\eta$ is a step size for NPG, and $\dagger$ denotes the Moore–Penrose pseudo-inverse \cite{horn2012matrix}.
In the discrete tabular case, with softmax policy parameterization, i.e., $\pi_\theta(a|s) = \frac{e^{\theta_{sa}}}{\sum_{a'} e^{\theta_{sa'}}}$, $\theta\in\mathbb{R}^{|S||A|}$, which is general enough to cover non-negative stochastic policies,  we have the closed-form for NPG update  \cite{npg_ent}:
\begin{equation}  
    \pi_{\theta_{t+1}}(a|s) 
    =  \frac{1}{Z_{\pi}(t,s)} (\pi_{\theta_{t}}(a|s))^\alpha 
    \exp \left(\frac{1-\alpha}{\tau}Q_{w_t,\tau}^{\pi_{\theta_t}}(s,a)\right), \label{eq:NPG111}
\end{equation}
where $\alpha = 1 - \frac{\eta \tau}{1 - \gamma}$, and $Z_\pi(t,s)$ is a normalization constant.  
In the case of linear function approximation for $\pi_\theta$, NPG can easily be implemented with compatible function approximation~\cite{suttonNPG}.
In the case of general nonlinear neural network parameterization, the computation of FIM is difficult but the NPG update for $\theta$ can readily be replaced  with TRPO \cite{trpo} or PPO \cite{ppo} of which purpose is to solve such divegence-constrained policy optimization. We will use PPO for deep implementation.

For the update of the strategy $w$ of $Adversary$, we use a variant of MD objective on $u_{Adv}^{\mathcal{RG}}$ with step size $\lambda$ as follows:
\begin{align}
    w_{t+1} 
     & =\argmax_{w\in\Delta^K} ~ \lambda \langle \nabla_w (-\langle w, \mathbf{V}^{\pi_{\theta_t}}\rangle)|_{w=w_t}, ~w \rangle  + \lambda\tau_w H(w) - D_\psi(w,w_t).  \label{eq:md_w}
\end{align}
Note that $\tau \tilde{H}(\pi_\theta)$ is irrelevant to $w$ update, and $\tau_wH(w)$ is outside of the gradient.  Again, we choose the Bregman divergence to be the KL divergence, considering $\{w\}$ forms a weight simplex  $\Delta^K$, a manifold identical to the probability simplex of $K$ outcomes.   With this choice, we have the following closed-form solution to  \eqref{eq:md_w} \cite{boyd2004convex}: 
\begin{equation}
    w_{t+1} = \softmax\left(-\frac{1-\beta}{\tau_w}\mathbf{V}^{\pi_{\theta_t}} + \beta\log w_t\right) \label{eq:w_closedform}
\end{equation}
where $\beta = \frac{1}{\lambda \tau_w + 1}$.
In this way, we circumvent any complicated iterative procedure to update $w_{t}$.  
Note that we could have derived another closed-form solution to $w_{t+1}$ for conventional MD on $u_{Adv}^{\mathcal{RG}}$, which yields the same solution as in \eqref{eq:w_closedform} with $\beta=1-\lambda\tau_w$.
We prefer the modified MD update  because the resulting closed-form solution employs $\beta = \frac{1}{\lambda\tau_w + 1}$, which  is  guaranteed to lie in $(0, 1)$ for any choice of hyperparameters $\lambda$ and $\tau_w$.
We note that \eqref{eq:NPG111} and \eqref{eq:w_closedform} together lead to an NE of the game $\mathcal{RG}$, as shown in Section \ref{section_tabular}.

\subsection{The ARAM algorithm: Adaptively-regularized adversary for max-min MORL}

Although ERAM provides an efficient model-free algorithm based on PPO and closed-form update of $w_t$, the performance can further be improved by detailing the $w_t$ update. We here develop Adaptively Regularized Adversary for Max-min MORL (\textbf{ARAM}), a variant of ERAM. 
In ARAM, we adopt  an adaptive regularization method that extends the standard entropy regularization. Note that
the regularizer $H(w)$ in the $Adversary$'s update rule~\eqref{eq:md_w} can be expressed as the KL divergence from the uniform reference weight
$\tfrac{1}{K}\mathbf{1}_K$, i.e.,  
$H(w) = -D_{\mathrm{KL}}\left(w \,\middle\|\, \tfrac{1}{K}\mathbf{1}_K\right) + \log K$. Thus, increasing entropy is equivalent to minimizing the distance from the uniform reference. In ARAM, we replaced  the uniform reference $\tfrac{1}{K}\mathbf{1}_K$  
 with a dynamically computed vector $c \in \Delta^K$, 
leading to the regularizer $-D_{\mathrm{KL}}(w \| c)$ instead of $H(w)$, where 
 $c$ captures the correlation between each reward component and the  worst-performing objective in the previous iteration. 
That is,  the $i$-th element of the reference vector $c$ is obtained as  
\begin{equation}
c_i=\mbox{softmax}(\mathbb{E}_{s,a} [ r_i(s,a) r_{i'}(s,a)]),~ i=1,\cdots,K,
\end{equation}
where $i'$ is the index of the worst-performing objective at the previous iteration batch and the expectation is replaced with sample expectation.
Thus, ERAM minimizes $\langle w, \mathbf{V}_\tau^{\pi_\theta}\rangle$ while trying to keep $w$ close to $\tfrac{1}{K}\mathbf{1}_K$, i.e., considering all objective dimensions equally. On the other hand, ARAM minimizes  $\langle w, \mathbf{V}_\tau^{\pi_\theta}\rangle$ while putting more emphasis on poorly-performing objective dimensions but not solely on the worst dimension, enabling joint optimization of multiple objective dimensions unlike previous GGF-PPO \citep{mdqn} which optimizes only the worst dimension at each batch.  Again, a closed-form formula for the $w$ update in ARAM can be derived and more on  ARAM  is provided in Appendix~\ref{append:corr}.

Summarizing the above, the pseudo-codes of the proposed algorithms are in Appendix~\ref{append:pseudo_code}.
Our source code is provided at \url{https://github.com/whbyeon/ERAM-ARAM}.

\section{Theoretical Analysis}\label{section_tabular}

In this section, we provide theoretical analysis of ERAM with respect to convergence, iteration complexity, and sample complexity under the assumption of tabular MOMDP in which the closed-form update \eqref{eq:NPG111} for $\theta$ is available with the softmax policy parameterization of  
$\pi_\theta(a|s)$, while  leaving theoretical analysis of more complicated ARAM as a possible future work. 
Note that the term "global convergence" means "convergence regardless of initial condition", and the term "last-iterate convergence of sequence $x_t$ to $x^*$" means "$x_t\to x^*$ as $t\to\infty$", whereas the term "average-iterate convergence" means  "$\frac{1}{T}\sum_{t=1}^T x_t\to x^*$ as $t\to\infty$",  which does not imply $x_t\to x^*$ in general. 

Note that for \eqref{eq:NPG111} and  \eqref{eq:w_closedform} we need to know the action value function since the state value function can readily be obtained from the action value function by summing over actions. We consider two cases regarding the knowledge of the action value function, exact and approximate policy evaluation cases, for proof of convergence. 

\textbf{Global convergence with exact policy evaluation:} ~~Theorem~\ref{thm:convergence_exact} shows the last-iterate convergence of Algorithm~\ref{alg:pseudo_code} when policy evaluation is exact.

\begin{theorem}\label{thm:convergence_exact}
Let $\{\theta_t\}_t$ and $\{w_t\}_t$ are the sequences generated by Algorithm~\ref{alg:pseudo_code} and let $\pi_t=\pi_{\theta_t}$. Then, the optimality gaps satisfy the following: 
\begin{align}
    \|\log\pi^*-\log\pi_t\|_\infty 
    \le{}& C_1 [\rho(\eta,\lambda)]^t\\
    \|w^*-w_t\|_\infty 
    \le{}& C_2 [\rho(\eta,\lambda)]^t\\
    \|Q^{\pi^*}_{w^*,\tau} - Q^{\pi_{t}}_{w_{t},\tau}\|_\infty 
    \le {}& C_3 [\rho(\eta,\lambda)]^t
\end{align}
for some $C_1,C_2,C_3$,
where $0<\rho(\eta,\lambda)\le 1-\frac{\epsilon^2}{2}< 1$  
with $\eta=\frac{\epsilon(1-\gamma)}{\tau}$, 
$\tau_w\ge\frac{12K(\max_{s,a,k}|r_k(s,a)|+\tau\log|A|)^2}{\tau(1-\gamma)^4}>0$
and $\epsilon\in(0,\epsilon_0)$ for some $\epsilon_0$. 
\end{theorem}

{\em Proof:}   See Appendix~\ref{convergence_pf_exact}.

For last-iterate convergence of policy and weight, the step size of weight-update $\lambda=O(\epsilon^2)$ should have a smaller scale than the step size of NPG $\eta=O(\epsilon)$. 
We note that if $\eta=O(\epsilon)$, then it suffices for $\lambda$ to have a smaller scale at least $O(\epsilon^{p}),~p>1$. Our choice $\lambda=O(\epsilon^2)$ is one possible choice that satisfies this condition.
In an intuitive sense, the policy is required to be updated faster than the weight.

\begin{corollary}  \label{cor:Cor42}
    Let the desired accuracy error tolerance be denoted by $\epsilon_{acc}$.
    To achieve 
    $\|\log\pi^*-\log\pi_t\|_\infty\le \epsilon_{acc}, ~
    \|w^*-w_t\|_\infty\le \epsilon_{acc},~
    \|Q^{\pi^*}_{w^*,\tau} - Q^{\pi_{t}}_{w_{t},\tau}\|_\infty
    \le \epsilon_{acc}$ with $\eta=\frac{\epsilon(1-\gamma)}{\tau}$, $\lambda = \frac{\epsilon^2}{\tau_w(1-\epsilon^2)}$, softmax policy and  exact policy evaluation, Algorithm~\ref{alg:pseudo_code} requires at most $O(\frac{1}{\epsilon^2}\log\frac{1}{\epsilon_{acc}})$ iterations.
\end{corollary}

\textbf{Global convergence with approximate policy evaluation:} ~~When exact policy evaluation is not available, our algorithm can be adapted to use approximate value estimates with bounded error. 
We establish that the last-iterate convergence guarantee still holds under this relaxed setting. 



\begin{theorem}\label{thm:convergence_inexact}
Assume that the estimated values $\widehat{Q}^{\pi}_{w,\tau}$ and $\widehat{Q}_k^{\pi}$ satisfy 
$\|\widehat{Q}^{\pi}_{w,\tau} - {Q}^{\pi}_{w,\tau}\|_\infty<\delta$ and $\|\widehat{Q}_k^{\pi} - {Q}_k^{\pi}\|_\infty<\delta$ for any $\pi,w,k$.  
Let $\{\theta_t\}_t$ and $\{w_t\}_t$ be the sequences generated by Algorithm~\ref{alg:pseudo_code_inexact}, a modified version of Algorithm 1 for approximate policy evaluation provided in Appendix~\ref{append:pseudo_code}, and let $\pi_t=\pi_{\theta_t}$. Then, the optimality gaps satisfy the following:
\begin{align}
    \|\log\pi^*-\log\pi_t\|_\infty 
    \le{}& \widehat{C}_1 [\widehat{\rho}(\eta,\lambda)]^t +\widehat{D}_1{\delta}/{\epsilon^2}\\
    \|w^*-w_t\|_\infty 
    \le{}& \widehat{C}_2 [\widehat{\rho}(\eta,\lambda)]^t +\widehat{D}_2{\delta}/{\epsilon^2}\\
    \|Q^{\pi^*}_{w^*,\tau} - Q^{\pi_{t}}_{w_{t},\tau}\|_\infty 
    \le {}& \widehat{C}_3 [\widehat{\rho}(\eta,\lambda)]^t +\widehat{D}_3{\delta}/{\epsilon^2}
\end{align}
where $0<\widehat{\rho}(\eta,\lambda)<1-\frac{\epsilon^2}{2}<1$ for $\epsilon\in(0,\epsilon_0)$
with the same condition in Theorem~\ref{thm:convergence_exact} except for the values of 
$\widehat{D}_1,\widehat{D}_2,\widehat{D}_3$ and $\epsilon_0$.
\end{theorem}

Proof is similar to that of Theorem~\ref{thm:convergence_exact} and is provided in Appendix~\ref{convergence_pf_inexact}.

\begin{corollary}\label{coro:sample_complexity}
    Let the accuracy error tolerance be denoted by $\epsilon_{acc}$.
    Assume that the estimation error $\delta$ is sufficiently small to satisfy $\delta\le\frac{\epsilon^2\epsilon_{acc}}{\widehat{D}_i},~i=1,2,3$.
    To achieve 
    $\|\log\pi^*-\log\pi_t\|_\infty\le 2\epsilon_{acc},~ 
    \|w^*-w_t\|_\infty\le 2\epsilon_{acc},~\text{and}~
    \|Q^{\pi^*}_{w^*,\tau} - Q^{\pi_{t}}_{w_{t},\tau}\|_\infty
    \le 2\epsilon_{acc}$ with $\eta=\frac{\epsilon(1-\gamma)}{\tau}$, $\lambda = \frac{\epsilon^2}{\tau_w(1-\epsilon^2)}$, softmax policy with approximate policy evaluation, Algorithm~\ref{alg:pseudo_code_inexact} requires at most $O(\frac{1}{\epsilon^2}\log\frac{1}{\epsilon_{acc}})$ iterations.
    Furthermore, with estimation error $\delta=O(\epsilon^2\epsilon_{acc})$ under Assumptions~\ref{assumption:generative_model} and \ref{assumption:estimate_error} provided in Appendix~\ref{append:sample_complexity_inexact}, by employing fresh samples for the policy evaluation for each objective at every iteration and taking the union bound for all objectives and all $O(\frac{1}{\epsilon^2}\log\frac{1}{\epsilon_{acc}})$ iterations, Algorithm~\ref{alg:pseudo_code_inexact} requires at most $\Tilde{O}(\frac{K}{(1-\gamma)^3\epsilon^4\epsilon_{acc}^2})\times O(\frac{1}{\epsilon^2}\log\frac{1}{\epsilon_{acc}}) = \Tilde{O}(\frac{K}{(1-\gamma)^3\epsilon^6\epsilon_{acc}^2})$ samples per each state-action pair. 
\end{corollary}

\begin{remark}
    It can be shown that the difference between the optimal value function induced  the regularized game $\mathcal{RG}$ and that of the  unregularized counterpart (i.e., $\mathcal{RG}$ with $\tau = \tau_w = 0$) is upper-bounded linearly in the regularization coefficients $\tau$ and $\tau_w$.
    This implies that the optimal solution of the regularized game yields a value function close to that of the original max–min MORL problem (i.e.,~\eqref{maxmin_ent_target_problem} with $\tau = 0$), as long as the regularization coefficients are sufficiently small.
\end{remark}

Proofs of Corollaries~\ref{cor:Cor42} and~\ref{coro:sample_complexity} are provided in Appendix~\ref{append:coro_pf}, and details for the sample complexity analysis are provided in Appendix~\ref{append:sample_complexity_inexact}.

\section{Related Works}
\label{related_works}

\textbf{Max-min MORL} ~~The utility-based approach is a key aspect of MORL, as it incorporates fairness and reflects user preferences~\cite{onestate,practical_guide_morl}.  
To capture broader notions of fairness, recent studies have adopted non-linear utility functions.   \citet{cousins_model_based} considered a model-based approach to fair RL but this method is  applicable to small finite problems. 
\citet{ESR_example_1} analyzed fairness through the lens of the Nash social welfare function, while \citet{ESR_nonlinear} proposed a reward-aware value iteration framework for general non-linear welfare functions, including the $\min$ operator.  
These works focused on optimizing the objective $\mathbb{E}_{\pi}\left[\min_k \sum_t \gamma^t r_k(s_t,a_t)\right]$, which is not the true min value  $\min_k\mathbb{E}_{\pi}\left[\sum_t \gamma^t r_k(s_t,a_t)\right]$, to simplify the problem.
\citet{mdqn} studied fair policy learning in MORL using the generalized Gini social welfare function, which includes max-min fairness as a special case.   GGF-DQN, their DQN-based method, optimizes again the surrogate objective $\mathbb{E}_\pi\left[\min_k \sum_t \gamma^t r_k(s_t,a_t)\right]$
due to the difficulty in constructing a Bellman operator under the non-linearity of the $\min$ operator.  GGF-PPO, their policy-based method, performs PPO update every iteration batch based on the current minimum objective value dimension.  Note that if we remove $H(w)$ by setting $\tau_w=0$ and remove $D_\psi(w,w_t)$ in our $w$ update \eqref{eq:md_w}, then the strategy $w$ of $Adversary$ is the one-hot vector with element 1 at  the minimum dimension of $\mathbf{V}_\tau^{\pi_\theta}$ and this case corresponds to GGF-PPO. Hence, our work can be considered as  a generalization of GGF-PPO. 
However, when $w$ is constrained to switching among one-hot vectors as in GGF-PPO, only average-iterate convergence is guaranteed~\cite{mwu}.
Note that our method has $H(w)$ encouraging to consider multiple dimensions simultaneously and $D_\psi(w,w_t)$ preventing rapid jumps in $w$ for last-iterate convergence. Performance improvement will be shown in Section \ref{section_experiment}.

For max-min fairness based on direct optimization of  $\min_k\mathbb{E}_{\pi}\left[\sum_t \gamma^t r_k(s_t,a_t)\right]$,  \citet{park} proposed a theoretical framework based on primal-dual convex programming for maximum-entropy reinforcement learning.  
To implement this framework, they introduced a model-free double-loop algorithm: the inner loop computes a stochastic zeroth-order gradient estimator using Gaussian smoothing, and the outer loop performs projected gradient descent with the estimated gradient.  The complexity of this method is far beyond that of our current method.

\textbf{Game-theoretic learning with regularization} ~~Regularization techniques have been widely used to compute Nash equilibria in game-theoretic settings.  
APMD~\cite{apmd} analyzes MD under smooth monotone game assumptions, which do not hold in our RL-based setting due to the non-monotonicity of the value gradients. In addition, APMD applies MD to all agents, whereas we combine RL-oriented NPG for the learner and MD for the adversary, which better suits deep RL implementations.  
\citet{r-nad} studied two-player zero-sum Markov games based on replicator dynamics and provided an asymptotic convergence analysis using Lyapunov techniques.  
\citet{lq-game} focused on regularized linear-quadratic games and established non-asymptotic convergence guarantees.  
\citet{2p0smg} proposed a regularized gradient descent–ascent method for two-player zero-sum Markov games and provided non-asymptotic analysis.
In contrast to these approaches, our method addresses a heterogeneous setting with one RL learner and a non-RL adversary, and is specifically tailored to max-min criterion for MORL. 
We also provide non-asymptotic convergence  for our method.
For a detailed discussion of zero-sum Markov game literature, 
we refer the reader to Appendix~\ref{append:2p0sMG}.

\textbf{Primal-dual methods and distributionally-robust RL} ~~
Several constrained RL methods adopt a primal--dual framework to solve constrained MDPs, where a Lagrangian function is formulated and an alternating optimization procedure is applied to maximize over the policy and minimize over the Lagrange multipliers~\cite{primal_dual_1,truly_no_regret,primal_dual_3,primal_dual_4,Efroni}. 
Although our algorithm can be viewed as one instance of primal-dual algorithms, our work differs from these approaches in that the weight $w$ lies in a probability simplex, whereas Lagrange multipliers of these works reside in the non-negative quadrant. 
In addition, due to $w \in \Delta^K$, we employ entropy regularization $H(w)$ instead of the $\|w\|^2$ regularization used in~\citet{truly_no_regret}, which allows us to obtain a closed-form solution for the $w$-update without requiring projected gradient descent used in \citet{Efroni,truly_no_regret}. 

Distributionally-robust reinforcement learning (DR-RL) typically addresses transition uncertainty~\cite{drrl_1,drrl_2,drrl_3}. 
Analogously, in reward-uncertain MDPs~\cite{rumdp}, one can define an uncertainty set in the reward space and apply a max--min formulation to achieve robustness. 
Our setting corresponds to a finite uncertainty set $\{r_1,\ldots,r_K\}$, where the regularization term captures such internalized reward uncertainty, in a manner similar to how DR-RL handles transition uncertainty. 
For infinite uncertainty sets, DR-RL could potentially be extended to analyze robustness in terms of a distribution over the reward space.


\section{Experiments}\label{section_experiment}

\subsection{Numerical results in tabular setting}\label{subsection:tabular}

\begin{wrapfigure}[9]{r}{0.35\textwidth}
\vspace{-3pt}
  \centering  \includegraphics[width=0.9\linewidth]{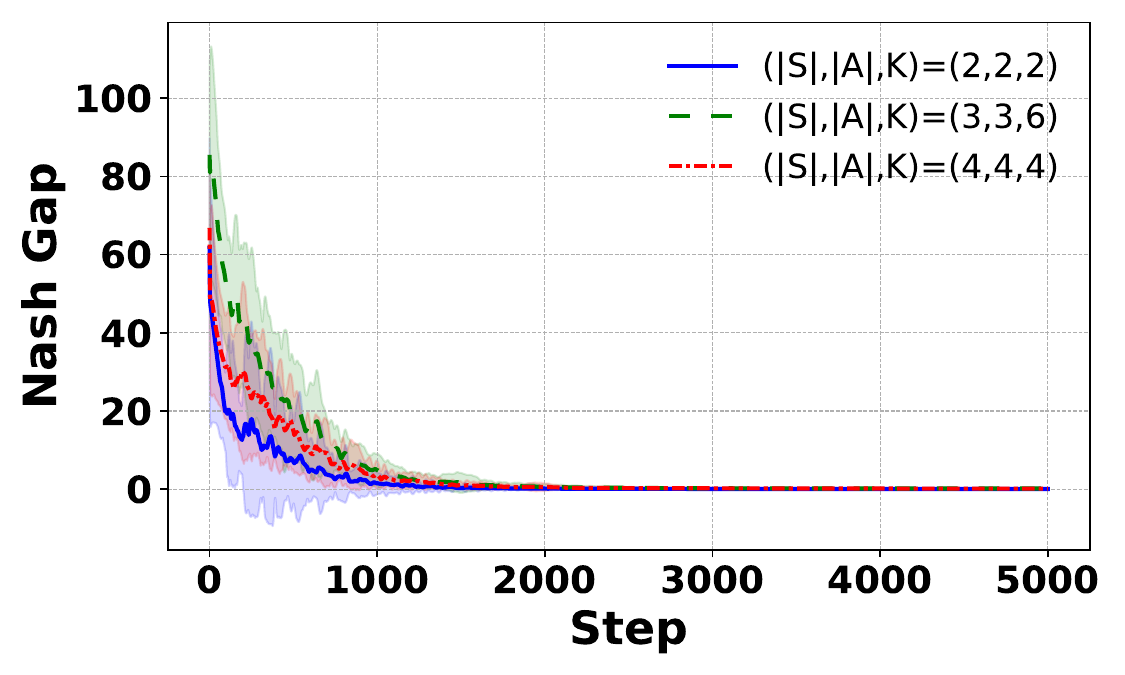}
  \captionsetup{aboveskip=2pt, belowskip=0pt}
  \caption{Nash gap}
  \label{fig:tabular_ng}
\end{wrapfigure}
We empirically demonstrate the convergence of our algorithm in tabular MOMDP settings.
We considered three types of tabular MOMDPs with different sizes: $(|S|,|A|,K) = (2,2,2)$, $(3,3,6)$, and $(4,4,4)$.
As the suboptimality measure, we used the Nash gap, a common metric in game theory~\cite{nash_gap_example_1,nash_gap_example_2}, defined as
$\text{Nash\_Gap}(\theta,w) = (\max_{\theta'} \langle w,\mathbf{V}^{\pi_{\theta'}}\rangle - \langle w,\mathbf{V}^{\pi_\theta}\rangle) + (\langle w,\mathbf{V}^{\pi_\theta}\rangle - \min_{w'}\langle w',\mathbf{V}^{\pi_\theta}\rangle)$.
This measures how far the strategy tuple $(\theta,w)$ is from a Nash equilibrium in the unregularized game.

Figure~\ref{fig:tabular_ng} shows that the Nash gap of Algorithm~\ref{alg:pseudo_code_inexact} decreases quickly over time in all three tabular settings.
Each curve represents the average over 50 randomly generated instances, with shaded areas showing standard deviation.
We also present the convergence behavior of ARAM with approximate policy evaluation in the tabular settings in Figure~\ref{fig:aram_ng} in Appendix~\ref{append:tabular}.

\subsection{Experimental results in traffic signal control}\label{subsection:tsc_result}

To evaluate the effectiveness of our algorithm in real-world multi-objective problems, we conducted experiments in the traffic signal control simulation environment~\cite{sumorl}.
At a four-road intersection, the agent controlled traffic signals based on traffic state information and received a reward vector composed of the negative total waiting times.
Rewards were defined either per road (4 objectives) or per lane (16 objectives). We considered three scenarios:
Base-4, Asym-4, and Asym-16, which differ in the number of objectives and traffic flow symmetry, as explained in Appendix~\ref{append:tsc_env}.

\begin{wrapfigure}[12]{r}{0.3\textwidth}
\vspace{-11pt}
  \centering
  \includegraphics[width=0.9\linewidth]{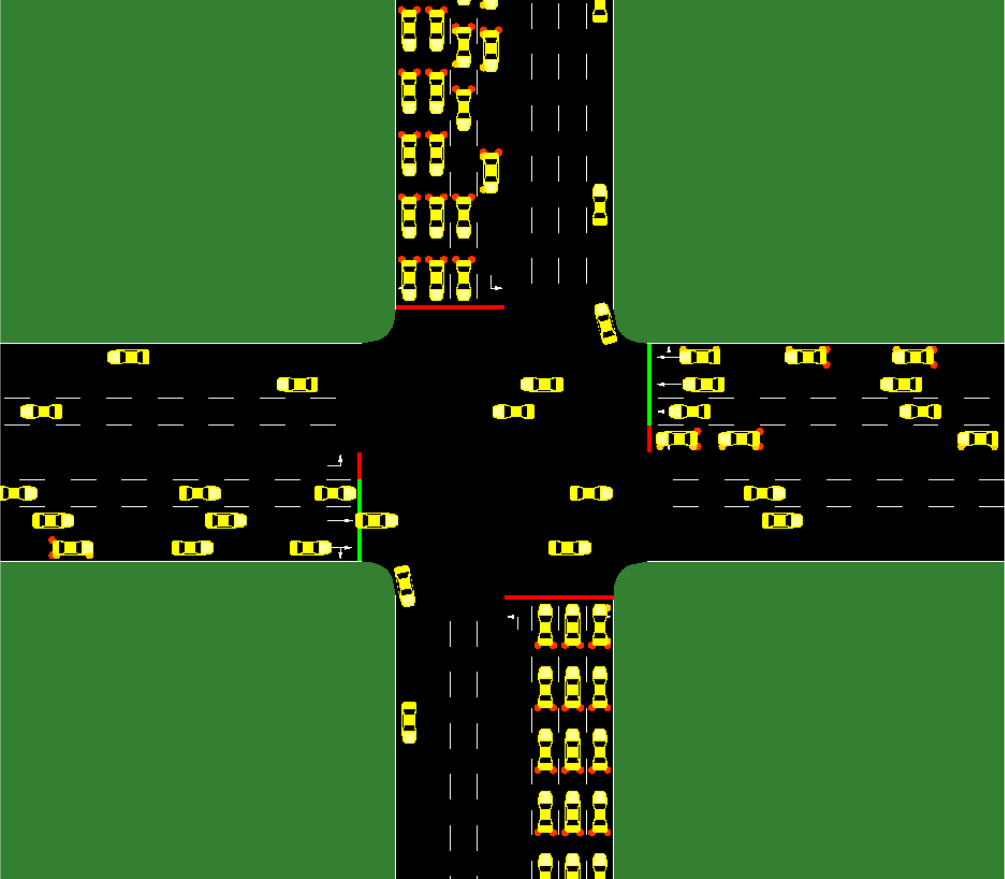}
  \captionsetup{aboveskip=2pt, belowskip=0pt}
  \caption{Traffic signal control environment \cite{sumorl}}
  \label{fig:traffic_env}
\end{wrapfigure}
We compared our method against GGF-DQN, GGF-PPO~\cite{mdqn}, \citet{park}, and Avg-DQN. 
Avg-DQN  optimizes the average reward $\frac{1}{K}\sum_k r_k(s,a)$, reflecting simple sum approaches.  
GGF-DQN optimizes a surrogate objective,
$\mathbb{E}_\pi\left[\min_k \sum_t \gamma^t r_k(s_t,a_t)\right]$,  
which yields a lower bound on the true max-min objective $\min_k\mathbb{E}_\pi[\sum_t \gamma^t r_k(s_t,a_t)]$ via Jensen's inequality.  
GGF-PPO performs a PPO update on the objective of minimum value at each iteration.
In contrast, the method of~\citet{park} directly solves the target problem~\eqref{maxmin_ent_target_problem} without relying on a surrogate, using projected gradient descent with a smoothed gradient estimate via Gaussian smoothing.
We evaluated the max-min performance as the minimum of the empirical return vector, i.e.,
$\min_k \hat{R}_k = \frac{1}{N} \sum_{i=1}^N \sum_t \gamma^t r^i_k(s_t, a_t)$,
averaged over $N=32$ episodes and five random seeds.
(Simulation details  are provided in Appendix~\ref{append:tsc_env}.)


\begin{table}[!h]
\vspace{-5pt}
\centering
\begin{tabular}{cccccccc}
\toprule
Environments & ARAM & ERAM & \citet{park} & GGF-PPO & GGF-DQN & Avg-DQN\\ 
\midrule
Base-4  & \textbf{-1160}  & \underline{-1387} & -1681    & -1731    & -1838      & -2774 & \\ 
Asym-4  & \textbf{-2696}  & \underline{-2732} & -3510    & -3501    & -3053      & -4245 & \\ 
Asym-16  & \textbf{-15043} & \underline{-17334} & -23663   & -21663   & -17792     & -27499 & \\ 
\bottomrule
\end{tabular}
\captionsetup{aboveskip=2pt, belowskip=0pt}
\caption{Max-min performance in traffic signal control. Bold: best; underline: second-best.}
\label{table:experiment_tsc}
\vspace{-10pt} 
\end{table}
Table~\ref{table:experiment_tsc} reports the max-min performance across the traffic signal control environments.   
ARAM consistently outperforms all baselines across all environments.  
ERAM achieves comparable results while maintaining architectural simplicity, and both methods directly optimize the target objective without relying on surrogate losses. 

More experimental results on other environments such as the species conservation environment~\cite{species_conservation}, MO-Reacher environment~\cite{mo_gym}, and Four-Room environment~\cite{mo_gym} are provided in Appendix~\ref{append:additional_experiments}, showing the superior max-min performance of our algorithms.

\subsection{Complexity comparison}\label{subsection:comparison_prior}

\begin{wrapfigure}[18]{r}{0.42\textwidth} 
\vspace{-12pt}
  \centering  \includegraphics[width=0.85\linewidth]{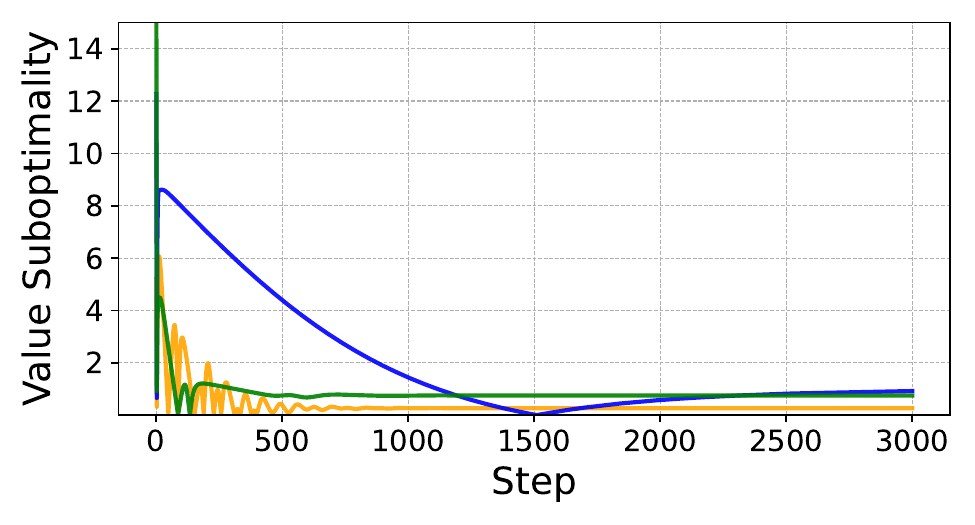}
  \vspace{-1pt}
\includegraphics[width=0.85\linewidth]{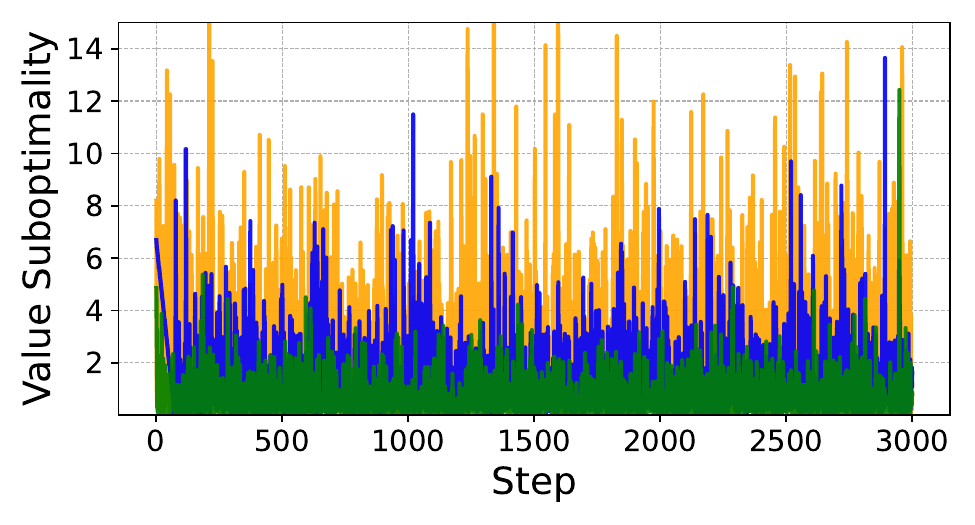}
  \vspace{-5pt} 
  \caption{Convergence comparison in random tabular MOMDPs: ERAM (top) vs. \citet{park} (bottom). Colors indicate different random MOMDPs.}
  \label{fig:convergence_compare}
\end{wrapfigure}
\textbf{Convergence behavior} ~~Our method guarantees last-iterate convergence, unlike \citet{park}, which ensures only average-iterate convergence. Figure~\ref{fig:convergence_compare} illustrates this difference in simple MOMDPs with two states, two actions, and two objectives. We present results on three representative randomly generated MOMDPs to ensure visual clarity; the same color is used to indicate the same MOMDP instance across the two plots. In each case, ERAM consistently approaches the true optimal value, up to a gap determined by the regularization coefficient, while the baseline exhibits oscillatory behavior.


\textbf{Memory efficiency} ~~To assess memory efficiency, we compared the number of model parameters used in each weight update. 
As in the original implementation, \citet{park} employs 20 copies of the Q-network, resulting in 274,084 parameters per update in the Base-4 and Asym-4 environments (4-dimensional reward), and 274,096 parameters in Asym-16 (16-dimensional reward). 
In contrast, our method uses only 13,704 and 14,484 parameters, respectively—corresponding to reductions of approximately 95\% and 94.7\%.
Despite this drastic reduction, our method achieves superior max-min performance, as shown in Table~\ref{table:experiment_tsc}.

\begin{wraptable}{r}{0.65\textwidth}
  \centering
  \small  
  \begin{tabular}{lllll}
    \toprule
    Envs & ERAM & ARAM & \citet{park} & GGF-PPO \\
    \midrule
    Base-4& \textbf{111} ± 2.6 & 120 ± 3.9 & 346 ± 14 & {122 ± 4.0}\\
    Asym-4 & 87.2 ± 2.4 & 87.4 ± 2.4 & 241 ± 6.3 & {\textbf{84.8 ± 2.2}}\\
    Asym-16 & \textbf{356} ± 27 & 365 ± 20 & 1125 ± 95 & {394 ± 5.5}\\
    \bottomrule
  \end{tabular}
  \caption{Training wall time (minutes), averaged over five seeds.}
  \label{table:training_time}
  \vspace{-6pt}
\end{wraptable}
\textbf{Computational cost efficiency} ~~To evaluate the computational complexity of our method, we compared the training wall-time across different environments. \citet{park} relies on an extensive soft Q-learning procedure for each weight update, leading to significantly longer training times. In contrast, our single-loop approach updates the weight and policy simultaneously, substantially reducing overall training time.
ERAM achieved training time reductions of approximately 67.8\% in Base-4, 63.8\% in Asym-4, and 68.4\% in Asym-16. 
Similarly, ARAM achieved reductions of approximately 65.5\% in Base-4, 63.7\% in Asym-4, and 67.6\% in Asym-16.
Although ARAM includes an additional step for computing correlation vectors, this overhead does not noticeably increase the total runtime. Overall, the computational complexity of ARAM is almost the same as that of ERAM. All experiments were conducted independently on the same hardware to ensure a fair comparison. See Table~\ref{table:training_time} for a full summary. 
Note that the memory requirement and computational complexity of ERAM and ARAM are at the same level of GGF-PPO since our $w$ update uses a closed-form solution  and the size of $w$ is only the number of objectives.

\subsection{Ablation study}
Since $Learner$ of our algorithm is PPO in deep RL cases, ablation study regarding $Learner$ can refer to PPO, and we focus on ablation study of $Adversary$ whose update is given by \eqref{eq:w_closedform} for ERAM or \eqref{eq:appendARAMwt} for ARAM with two hyperparameters $\lambda$ and $\tau_w$. 
The algorithms are not so sensitive when $\beta$ is not near 0 in ERAM or not near 1 in ARAM. 
When $\beta \approx 0$, ERAM effectively omits the mirror descent term, allowing us to observe the impact of MD. 
When $\beta \approx 1$, ARAM ignores the adaptive regularizer, highlighting its contribution to performance.
Please see Appendix~\ref{append:ablation} and Figure~\ref{fig:ablation_star_heatmaps} for more.

\section{Conclusion}
In this paper, we have considered MORL with max-min criterion. Exploiting the special max-min and min-max equivalence in this problem, we have reformulated max-min MORL as a two-player zero-sum regularized continuous game. We have proven the existence of a NE of this game, which yields a max-min MORL policy. Then, we have proposed  efficient algorithms to find a NE of this game, where the learner can use conventional PPO to update its strategy and the adversary updates its strategy based on a closed-form formula. We have proven last-iterate convergence of the algorithms in the tabular case and demonstrated that the proposed algorithms significantly outperform existing max-min MORL methods.  The proposed max-min MORL algorithm has the complexity of PPO and can be  used practically for many real-world resource allocation  problems. 

\section{Acknowledgments}
This work was supported by the National Research Foundation of Korea (NRF) grant
funded by the Korea government (MSIT) (No. RS-2025-
00557589, Generative Model Based Efficient Reinforcement Learning Algorithms for Multi-modal Expansion in
Generalized Environments)
and by 
the Ministry of Innovation, Science \& Technology, Israel grant 1001556423 and ISF grant 2197/22. 

\newpage

\bibliography{refer.bib}

\section*{NeurIPS Paper Checklist}

\begin{enumerate}

\item {\bf Claims}
    \item[] Question: Do the main claims made in the abstract and introduction accurately reflect the paper's contributions and scope?
    \item[] Answer: \answerYes{} 
    \item[] Justification: {The claims in the abstract and introduction clearly reflect the theoretical contributions and experimental findings presented in the paper.}


\item {\bf Limitations}
    \item[] Question: Does the paper discuss the limitations of the work performed by the authors?
    \item[] Answer: \answerYes{} 
    \item[] Justification: We discuss the limitations in Appendix~\ref{append:limitation}.

\item {\bf Theory assumptions and proofs}
    \item[] Question: For each theoretical result, does the paper provide the full set of assumptions and a complete (and correct) proof?
    \item[] Answer: \answerYes{} 
    \item[] Justification: 
    We provide the assumptions and proofs for Theorem~\ref{thm:saddle_is_maxminmorl}, Theorem~\ref{thm:convergence_exact}, Theorem~\ref{thm:convergence_inexact}, Corollary~\ref{cor:Cor42}, and Corollary~\ref{coro:sample_complexity} in Appendix~\ref{section:saddle_pf}, Appendix~\ref{convergence_pf_exact}, Appendix~\ref{convergence_pf_inexact}, Appendix~\ref{append:coro_pf}, and Appendix~\ref{append:sample_complexity_inexact}.


    \item {\bf Experimental result reproducibility}
    \item[] Question: Does the paper fully disclose all the information needed to reproduce the main experimental results of the paper to the extent that it affects the main claims and/or conclusions of the paper (regardless of whether the code and data are provided or not)?
    \item[] Answer: \answerYes{} 
    \item[] Justification: 
    The experimental environments are described in Appendices~\ref{append:tabular},~\ref{append:traffic}, and~\ref{append:additional_experiments}, and the hyperparameters are provided in Appendix~\ref{append:hyperparam}.

\item {\bf Open access to data and code}
    \item[] Question: Does the paper provide open access to the data and code, with sufficient instructions to faithfully reproduce the main experimental results, as described in supplemental material?
    \item[] Answer: \answerYes{} 
    \item[] Justification: We provide the source code for the traffic signal control experiment, which constitutes the main experimental result of the paper.

\item {\bf Experimental setting/details}
    \item[] Question: Does the paper specify all the training and test details (e.g., data splits, hyperparameters, how they were chosen, type of optimizer, etc.) necessary to understand the results?
    \item[] Answer: \answerYes{} 
    \item[] Justification: 
    The experimental setups and environments are described in Appendix~\ref{append:tabular} and \ref{append:traffic}.
    The source code is provided and hyperparameters are listed in Appendix~\ref{append:hyperparam}.

\item {\bf Experiment statistical significance}
    \item[] Question: Does the paper report error bars suitably and correctly defined or other appropriate information about the statistical significance of the experiments?
    \item[] Answer: \answerYes{} 
    \item[] Justification: We used five random seeds for the deep RL benchmarks in Section~\ref{subsection:tsc_result} and Appendix~\ref{append:additional_experiments}, and fifty random seeds for the tabular settings in Section~\ref{subsection:tabular} and Appendix~\ref{append:tabular}. Figure~\ref{fig:tabular_ng} and Table~\ref{table:training_time} report the standard deviation across seeds, with shaded areas in the figure and numerical values in the table. Figure~\ref{fig:convergence_compare} shows results from three random seeds for improved visual clarity.


\item {\bf Experiments compute resources}
    \item[] Question: For each experiment, does the paper provide sufficient information on the computer resources (type of compute workers, memory, time of execution) needed to reproduce the experiments?
    \item[] Answer: \answerYes{} 
    \item[] Justification: Device specifications are provided in Appendix~\ref{append:hyperparam}.
    
\item {\bf Code of ethics}
    \item[] Question: Does the research conducted in the paper conform, in every respect, with the NeurIPS Code of Ethics \url{https://neurips.cc/public/EthicsGuidelines}?
    \item[] Answer: \answerYes{} 
    \item[] Justification: The research fully complies with the NeurIPS Code of Ethics.

\item {\bf Broader impacts}
    \item[] Question: Does the paper discuss both potential positive societal impacts and negative societal impacts of the work performed?
    \item[] Answer: \answerYes{} 
    \item[] Justification: We discuss the social impacts in Appendix~\ref{append:limitation}.
    
\item {\bf Safeguards}
    \item[] Question: Does the paper describe safeguards that have been put in place for responsible release of data or models that have a high risk for misuse (e.g., pretrained language models, image generators, or scraped datasets)?
    \item[] Answer: \answerNA{} 
    \item[] Justification: The paper does not pose risks of misuse requiring special safeguards.

\item {\bf Licenses for existing assets}
    \item[] Question: Are the creators or original owners of assets (e.g., code, data, models), used in the paper, properly credited and are the license and terms of use explicitly mentioned and properly respected?
    \item[] Answer: \answerYes{} 
    \item[] Justification: We use Stable Baselines3 (MIT license), SUMO-RL (MIT license), and MO-Gymnasium (MIT license), all of which are properly cited and used in compliance with their respective licenses.

\item {\bf New assets}
    \item[] Question: Are new assets introduced in the paper well documented and is the documentation provided alongside the assets?
    \item[] Answer: \answerNA{} 
    \item[] Justification: We do not release new assets.

\item {\bf Crowdsourcing and research with human subjects}
    \item[] Question: For crowdsourcing experiments and research with human subjects, does the paper include the full text of instructions given to participants and screenshots, if applicable, as well as details about compensation (if any)? 
    \item[] Answer: \answerNA{} 
    \item[] Justification: The paper does not involve crowdsourcing or human-subject research.

\item {\bf Institutional review board (IRB) approvals or equivalent for research with human subjects}
    \item[] Question: Does the paper describe potential risks incurred by study participants, whether such risks were disclosed to the subjects, and whether Institutional Review Board (IRB) approvals (or an equivalent approval/review based on the requirements of your country or institution) were obtained?
    \item[] Answer: \answerNA{} 
    \item[] Justification: The research does not involve human subjects and therefore does not require IRB approval.

\item {\bf Declaration of LLM usage}
    \item[] Question: Does the paper describe the usage of LLMs if it is an important, original, or non-standard component of the core methods in this research? Note that if the LLM is used only for writing, editing, or formatting purposes and does not impact the core methodology, scientific rigorousness, or originality of the research, declaration is not required.
    \item[] Answer: \answerNA{} 
    \item[] Justification: The research does not involve LLMs as an important or original component of the core methodology.

\end{enumerate}

\appendix
\newpage

\section{ Glossary }\label{glossary}
\begin{table}[!ht]
	\centering
	\begin{adjustbox}{width=\textwidth}
		\begin{tabular}{ll}
			\toprule
			Notations & Descriptions   \\
			\hline
                \textbf{MOMDP} & {}\\ 
                \hline
			$S$ & State space \\
			$A$ & Action space  \\
			$P$ & Transition dynamics  \\
			$\mu$  & Initial state distribution \\
			$\mathbf{r}$ & Multi-objective reward in the MOMDP  \\
                $K$ & number of objectives, i.e., dimension of multi-objective reward \\
                $r_k$ & $k$-th coordinate of vector reward $\mathbf{r}$, $k=1,\ldots,K$ \\
			$\gamma$ & Discount factor \\
			$\pi,\Pi$ & Policy, Policy space \\
                $\theta,\Theta$ & Policy parameter, Policy parameter space\\
                $w$ & weight, action of $Adversary$\\
                \hline
                \textbf{Values} & {}\\ 
                \hline
                $\mathbf{V}^\pi$ & $\mathbb{E}_{\mu,\pi}\left[\sum_{t=0}^\infty \gamma^t \mathbf{r}(s_t,a_t)\right]\in\mathbb{R}^K$, Value vector under policy $\pi$ in the MOMDP \\
                $V_k^\pi$ & $\mathbb{E}_{\mu,\pi}\left[\sum_{t=0}^\infty \gamma^t r_k(s_t,a_t)\right]\in\mathbb{R}$, $k$-th coordinate of value vector under policy $\pi$ in the MOMDP, $k=1,\ldots,K$ \\
                $V^\pi_w$ & $\langle w, \mathbf{V}^\pi\rangle = \sum_{k=1}^K w(k) V_k^\pi$, the weighted value for any $w\in\Delta^K$.\\
                $\mathbf{V}^\pi_\tau$ & $\mathbb{E}_{\mu,\pi}\left[\sum_t \gamma^t (\mathbf{r}(s_t,a_t)-\tau\log\pi(a_t|s_t)\mathbf{1}_K)\right]\in\mathbb{R}^K$, soft value vector with $K$ objectives and entropy coefficient $\tau$\\
                $V^\pi_{k,\tau}$ & $k$-th coordinate of soft value vector\\
                $V^\pi_{w,\tau}$ & $\langle w, \mathbf{V}^\pi_\tau\rangle = \mathbb{E}_{\mu,\pi}\left[\sum_t \gamma^t (\langle w, \mathbf{r}(s_t,a_t)\rangle-\tau\log\pi(a_t|s_t))\right]\in\mathbb{R}$, soft value with scalar reward $\langle w,\mathbf{r}\rangle$\\
                $V^*_{w,\tau}, Q^*_{w,\tau}$ & Soft optimal values with scalar reward $\langle w,\mathbf{r}\rangle$\\
                \hline
                \textbf{Constants} & {}\\ 
                \hline
                $\tau$ & Entropy coefficient for NPG\\
                $\tau_w$ & Entropy coefficient for mirror descent on $w$\\
                $\eta$ & Step size for NPG\\
                $\lambda$ & Step size for mirror descent on $w$\\
                $\alpha$ & $1-\frac{\eta\tau}{1-\gamma}$\\
                $\beta$ & $\frac{1}{\lambda\tau_w + 1}$\\
                \hline
                \textbf{Mathematical Notations} & {}\\ 
                \hline
                $\Delta^K$ & $(K-1)$-Simplex, i.e., $\{w\in\mathbb{R}^K|\sum_{k=1}^K w(k) =1, w(k)\ge0,~k=1,\ldots,K\}$ \\
                $\langle x,y \rangle$ & $\sum_{i=1}^d x_i y_i$, inner product for any $x,y\in\mathbb{R}^d$. \\
                $H(w)$ & $-\sum_{k=1}^K w(k)\log w(k)$, Shannon entropy for any $w\in\Delta^k$\\
                $\tilde{H}(\pi)$ &  $\mathbb{E}_{\mu,\pi}\left[-\sum_t \gamma^t \log\pi(a_t|s_t)\right]$\\
                $\softmax(x)$ & $\left(\frac{e^{x_1}}{\sum_{i=1}^d e^{x_i}},\ldots,\frac{e^{x_d}}{\sum_{i=1}^d e^{x_i}}\right)\in\mathbb{R}^d$ for any $x\in\mathbb{R}^d$\\
                $\mathbf{1}_d$ & all-one-vector with dimension $d$\\ 
                $D_{KL}$ & Kullback Leibler divergence\\
                $\| \cdot \|_1$ & $l_1$-norm, i.e., $\|x\|_1=\sum_{i=1}^d |x_i|$ for $x\in\mathbb{R}^d$\\
                $\| \cdot \|_\infty$ & $l_\infty$-norm, i.e., $\|x\|_\infty=\max_i |x_i|$\\
                \bottomrule
		\end{tabular}
	\end{adjustbox}
	\caption{ Notations in the main paper  }
	\label{tab:notations}
\end{table}


\section{Relationship between Max-Min Fairness and Other Fairness Criteria in MORL}\label{append:fairness}

We can consider  three major scalarization methods for MORL, as follows:

\begin{itemize}
    \item Sum return maximization
     \begin{equation}
         \max_\pi \sum_{i=1}^K w_i V_i
     \end{equation}
     \item Max-min optimization
     \begin{equation}
         \max_\pi \min \{w_1 V_1, \cdots, w_K V_K\}
     \end{equation}
     \item Proportionally-fair (PF) optimization
        \begin{equation}  \label{eq:PFform}
         \max_\pi \sum_{i=1}^K w_i \log V_i
     \end{equation}
\end{itemize}
Here, $w_1,\cdots,w_K$ are some given weighting factors. We consider the weighted version of all three methods above for full generality. 

The sum return maximization for given weights is a simple single-objective RL for which there are many existing algorithms. Hence, the problem is rather simple.  

Consider the PF problem. Since the logarithm function is differentiable, the policy gradient of the PF objective  \eqref{eq:PFform} can easily be obtained. That is, let us use  $V_i^{\pi_\theta}$ instead of $V_i$ to show the dependence of return on the policy parameter $\theta$ explicitly.
Then, 
\begin{eqnarray}
    \frac{\partial}{ \partial \theta } \sum_i w_i \log V_i^{\pi_\theta} &=&  \sum_i w_i \frac{1}{V_i^{\pi_\theta}}  \frac{\partial V_i^{\pi_\theta}}{ \partial \theta }  .
\end{eqnarray}
Note that $\frac{\partial V_i^{\pi_\theta}}{ \partial \theta }$ is nothing but the conventional policy gradient, which is given by  $\nabla_\theta 
\log \pi_\theta (s,a)Q_i (s,a)$ \cite{sutton}.  Hence, it is rather straightforward to solve the PF problem for MORL and we only need to track individual Q functions. 

In contrast, the max-min problem does not allow such direct differentiation and becomes more complicated. Hence, a specialized method is needed as in \cite{park} or approximate approaches were considered as in \cite{ESR_example_1,ESR_nonlinear,mdqn}. In this paper, we solve the exact max-min MORL problem and provide a very efficient algorithm for this problem. 

\begin{figure}[H]
    \centering
    \includegraphics[width=0.5\textwidth]{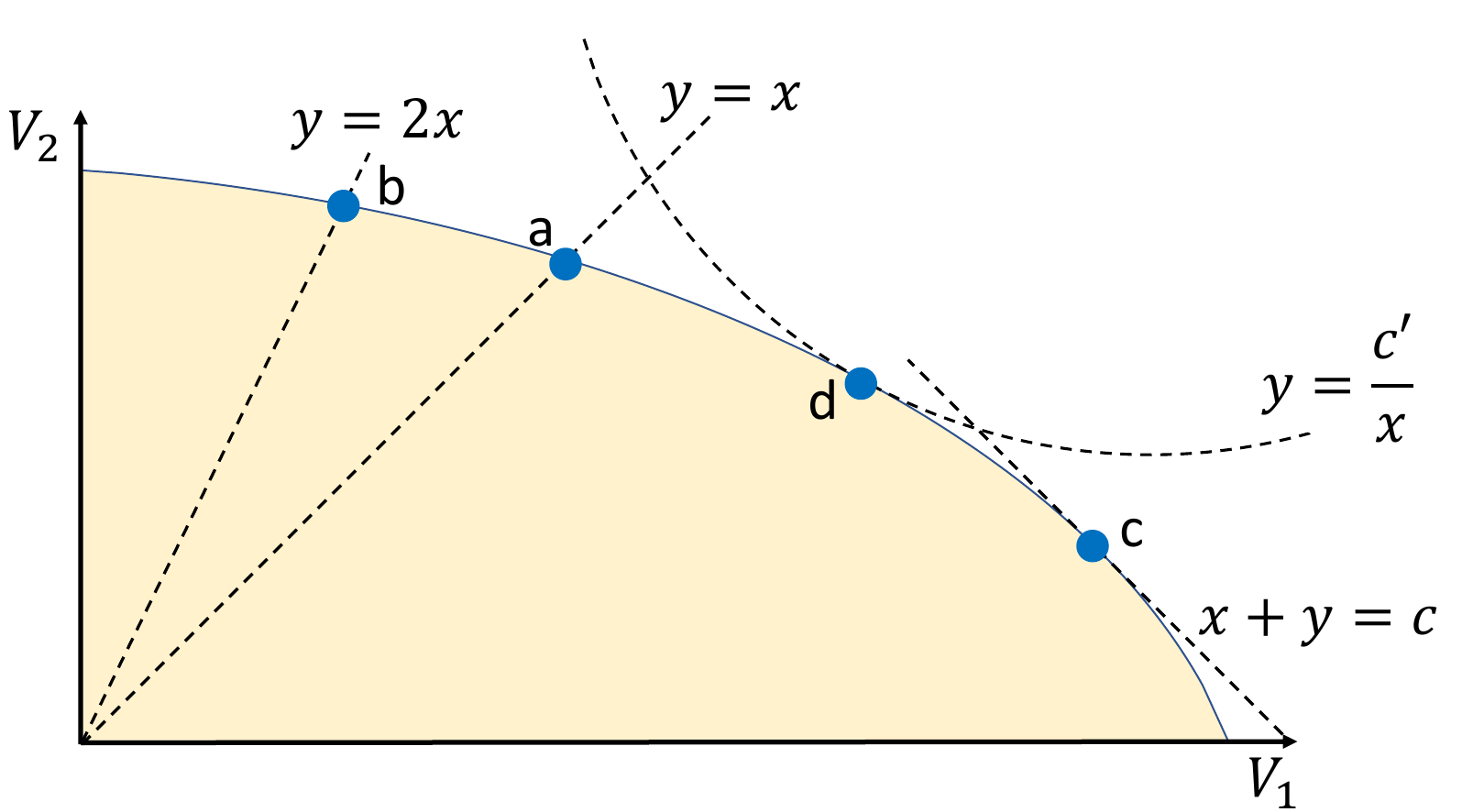}
    \captionsetup{aboveskip=2pt, belowskip=0pt}
    \caption{Relationship between max-min fairness and other fairness notions in MORL.}
    \label{fig:fairness_diagram}
\end{figure}

Figure~\ref{fig:fairness_diagram} illustrates the solution of each method for equal weights with two objectives. 
In the figure, all of points \textbf{a}, \textbf{b}, \textbf{c}, and \textbf{d} are Pareto optimal. 
The max-min  solution, point \textbf{a}, corresponds to the Pareto-optimal point that achieves Pareto efficiency with $V_1=V_2$ due to equalizer rule \cite{equalizer_rule}. 
Note that by weighting $w_1$ and $w_2$, we can achieve the Pareto-boundary point on which $w_1 V_1=w_2 V_2$ due to the same equalizer rule, e.g., Point \textbf{b} in the figure. 

Point \textbf{c} is the sum return maximization point on which the Pareto-boundary is tangential with the straight line $V_1 + V_2 =c$ for some $c$. Again, we can change the slope of the line by considering weights as $w_1 V_1 + w_2 V_2 =c$.

Point \textbf{d} is the PF point on which Pareto-boundary is tangential to the curve $V_1V_2 = c'$ for some $c'$ because $\sum_i \log V_i = \log \prod_i V_i$.  Note that the PF point is a compromised point between the sum maximization point and the max-min point.

\section{Brief Summary of the Prior Work: Convex Optimization Formulation of Max-Min MORL \cite{park}} \label{append:prior_work}

Based on the fact that value function is linear in state-action occupancy measure \cite{sutton,puterman}, 
\citet{park} rewrote the max-min MORL without entropy regularization in terms of state-action occupancy measure (\textbf{P0} in \citet{park}).

\begin{equation} \label{PO}
\textbf{P0}:     ~~~\max_{d} \min_{k=1,\ldots,K}  \sum_{s,a} d(s,a) r_k(s,a) 
\end{equation}
\begin{align} 
    \sum_{a'} d(s',a') =& \mu(s') + \gamma \sum_{s,a} P(s' | s,a) d(s,a), ~~ \forall s'\\
    d(s,a) \geq& 0, ~~ \forall s,a.
\end{align}

By taking dual problem and linear programming formulation of value iteration \cite{puterman}, 
\citet{park} has shown that 
solving \textbf{P0} is equivalent to solving the following convex optimization problem \textbf{P2}.
The convexity of the objective function in \textbf{P2}, and the equivalence of \textbf{P0} and \textbf{P2} are addressed in Theorem 3.1 and Theorem 3.2 in \citet{park}, respectively.

\begin{equation}\label{P2}
    \textbf{P2}\text{ in \citet{park}}: \min_{w \in \Delta^K} \sum_s \mu(s)V^{*}_{w}(s)
\end{equation}
where $V^{*}_{w}(\cdot)$ is optimal value function with linearly scalarized reward $\langle w, \mathbf{r}(s,a)\rangle$.\\

However, this value iteration based method can fail to find optimal policy in max-min MORL even in a simple one-state MOMDP \cite{onestate,park}. 
This issue is called indeterminacy; that is, there exists an MOMDP which has no max-min optimal policy, which is stationary and deterministic. 
To circumvent this issue,
\cite{park} adopted entropy regularization to ensure the policy stochastic and established convex programming reformulation similar to above:
\begin{equation}\label{P0'}
    \textbf{P0':}~
    \max_{d} \min_{k=1,\ldots,K}  \sum_{s,a} d(s,a) \left\{ r_k(s,a) + \tau H( \pi^d(\cdot | s)) \right\} 
\end{equation}
\begin{align} 
    \sum_{a'} d(s',a') =& \mu(s') + \gamma \sum_{s,a} P(s' | s,a) d(s,a), ~~ \forall s'\\
    d(s,a) \geq& 0, ~~ \forall s,a
\end{align}
where $\pi^d(a|s):=\frac{d(s,a)}{\sum_{a'}d(s,a')}$.

Similar to above, solving \textbf{P0'} is equivalent to solving the following convex optimization problem \textbf{P2'}.
The convexity of the objective function in \textbf{P2'}, and the equivalence of \textbf{P0'} and \textbf{P2'} are addressed in Theorem 4.1 and Theorem 4.2 in \citet{park}, respectively.

\begin{equation}\label{P2'}
    \textbf{P2'}\text{ in \citet{park}}: \min_{w \in \Delta^K} \sum_s \mu(s)V^{*}_{w,\tau}(s)
\end{equation}
where $V^{*}_{w,\tau}(\cdot)$ is the soft optimal value function with scalar reward $\langle w, \mathbf{r}(s,a)\rangle$ which is the solution of soft Bellman equation in $v$.
\begin{equation*}
    v(s) = \tau \log \sum_a \exp [ \frac{1}{\tau}\{ \langle w, \mathbf{r}(s,a)\rangle + \gamma \sum_{s'}P(s'|s,a)v(s') \} ],~\forall s
\end{equation*}

\section{More on Adaptive Regularization of $Adversary$}\label{append:corr}

With the adaptive regularizer $-D_{KL}(w \| c_t)$, where
\begin{equation}
    c_t := c(\pi_t) = \softmax\left( \mathbb{E}_{(s,a)\sim d^{\pi_t}} \left[ r_k(s,a) \cdot r_{i'_t}(s,a) \right] \right)_{k=1}^K,
\end{equation}
and where $i'_t$ denotes the index of the worst-performing objective at iteration $t$,
i.e., $V^{\pi_t}_{i'_t} \le V^{\pi_t}_k$ for all $k = 1, \ldots, K$,
the $Adversary$'s update rule in Equation~\eqref{eq:md_w} becomes:
\begin{align}
    w_{t+1} =&\argmax_{w\in\Delta^K} ~ \lambda \langle \nabla_w (-\langle w, \mathbf{V}^{\pi_{\theta_t}}\rangle)|_{w=w_t}, ~w \rangle  - \lambda\tau_w D_{KL}(w\|c_t) - D_{KL}(w\|w_t)\\
    =&\argmax_{w\in\Delta^K} ~ \langle -\lambda\mathbf{V}^{\pi_{\theta_t}} + \lambda\tau_w \log c_t + \log w_t,~w\rangle + (\lambda \tau_w + 1)H(w)\\
    =&\argmax_{w\in\Delta^K} ~ \langle -\frac{1-\beta}{\tau_w}\mathbf{V}^{\pi_{\theta_t}} + (1-\beta) \log c_t + \beta\log w_t,~w\rangle + H(w).
\end{align}
This leads to the following closed-form solution~\cite{boyd2004convex}:
\begin{equation}\label{eq:corr_update}
w_{t+1} = \softmax\left( -\frac{1 - \beta}{\tau_w} \mathbf{V}^{\pi_t} + \beta \log w_t + (1 - \beta) \log c(\pi_{t}) \right).
\end{equation}

The pseudo-code of the deep RL implementation of ARAM combined with PPO is provided in Section \ref{sec:ARAMcode}.

\section{Pseudo Codes}
\label{append:pseudo_code}
\subsection{ERAM}\label{append:eramm_pseudo_code}

\begin{algorithm}[H] 
   \caption{Adversary with regularizer for max-min MORL employing exact policy evaluation}
   \label{alg:pseudo_code}
\begin{algorithmic}
   \State {\bfseries Input:} initial policy parameter $\theta_0$, initial weight $w_0$, number of iterations $T$, NPG step size $\eta$, MD step size $\lambda$, regularizer coefficients: $\tau,\tau_w$
   \For{$t=1$ {\bfseries to} $T$}
   \State Evaluate current policy $\pi_t$ and obtain $\mathbf{V}^{\pi_t}$
   \State $\theta_{t+1} \leftarrow \theta_t + \eta F^\dagger(\theta_t)
    \sum_{k=1}^K w_t(k) \nabla_\theta V^{\pi_{\theta_t}}_{k,\tau}$
   \Comment{update $Learner$'s strategy}
   \State $w_{t+1} \leftarrow \softmax\left(-\frac{1-\beta}{\tau_w}\mathbf{V}^{\pi_t} + \beta\log w_t\right)$ 
   \Comment{update $Adversary$'s strategy in closed form}
   \EndFor
   \State {\bfseries Return: } $\pi_T, w_T$
\end{algorithmic}
\end{algorithm}

In the tabular case with softmax policy, the $\theta$ update  is simplified as 
\begin{equation}\label{eq:pseudo_code_update_example}
    \pi_{\theta_{t+1}}(a|s) 
    \propto (\pi_{\theta_{t}}(a|s))^\alpha \exp (\frac{1-\alpha}{\tau}{Q}_\tau^{\pi_{\theta_t}}(s,a)).
\end{equation}

In the deep learning setting, the $\theta$ update can be replaced with PPO \cite{ppo}, and the corresponding $\theta$ update is seen in Appendix  \ref{sec:ARAMcode}.

\subsection{ERAM with Approximate Policy Evaluation}
\begin{algorithm}[H]
   \caption{ERAM employing approximate policy evaluation}
   \label{alg:pseudo_code_inexact}
\begin{algorithmic}
   \State {\bfseries Input:} initial policy parameter $\theta_0$, initial weight $w_0$, number of iterations $T$, NPG step size $\eta$, MD step size $\lambda$, regularizer coefficients: $\tau,\tau_w$
   \For{$t=1$ {\bfseries to} $T$}
   \State Obtain approximate policy evaluation $\widehat{\mathbf{V}}^{\pi_t}$
   for current policy $\pi_t$ 
   \State $\theta_{t+1} \leftarrow \theta_t + \eta F^\dagger(\theta_t)
    \sum_{k=1}^K w_t(k) \nabla_\theta V^{\pi_{\theta_t}}_{k,\tau}$ 
    \Comment{update $Learner$'s strategy with $\widehat{\mathbf{V}}^{\pi_t}$, e.g.,~\eqref{eq:pseudo_code_update_example}}
   \State $w_{t+1} \leftarrow softmax(-\frac{1-\beta}{\tau_w}\widehat{\mathbf{V}}^{\pi_t} + \beta\log w_t)$  
   \Comment{update $Adversary$'s strategy}
   \EndFor
   \State {\bfseries Return: } $\pi_T, w_T$
\end{algorithmic}
\end{algorithm}

In the tabular case with softmax policy, the $\theta$ update with approximate policy evaluation is simplified as 
\begin{equation}\label{eq:pseudo_code_update_example}
    \pi_{\theta_{t+1}}(a|s) 
    \propto (\pi_{\theta_{t}}(a|s))^\alpha \exp (\frac{1-\alpha}{\tau}\widehat{Q}_\tau^{\pi_{\theta_t}}(s,a)).
\end{equation}

In the deep learning setting, the $\theta$ update can be replaced with PPO \cite{ppo}, and the corresponding $\theta$ update is seen in Appendix  \ref{sec:ARAMcode}.

\subsection{ARAM with PPO for the Learner Update} \label{sec:ARAMcode}
\begin{algorithm}[H]    
   \caption{Adversary with adaptive regularizer for max-min MORL with multi-objective variant of PPO}
\begin{algorithmic}
   \State {\bfseries Input:} policy network $\pi_\theta$ and state-value vector network $\mathbf{V}_\phi$, initial parameters $\theta_0,\phi_0$, clip coefficient $\epsilon_{clip}$, initial weight $w_0$, number of iterations $T$, MD step size $\lambda$, regularizaion coefficients: $\tau_{PPO}, \tau_w$
   \For{$t=1$ {\bfseries to} $T$}
   \State Collect samples from $\pi_{\theta_t}$ and update critic buffer $\mathcal{B}_{critic}$
   \State $\widehat{\mathbf{V}}_t := \mathbf{V}_{\phi_t}$
   \Comment{Get estimate of value for current policy with current critic network}
   \State Compute $c_t$:\[
  c(\pi_{t}) = \softmax\left( \mathbb{E}_{(s,a)\sim d^{\pi_{\theta^t}}}[r_k(s,a) r_{i'_t}(s,a)] \right)_{k=1}^K
  \]
   \State Update $Learner$'s strategy with PPO
   \State ~~~~~~~~~~~Optimize $L_{clip}(\theta;\theta_t,w_t)$ to obtain $\theta_{t+1}$
   \State ~~~~~~~~~~~Optimize $L_{critic}(\phi)$ to obtain $\phi_{t+1}$
   \State Update $Adversary$'s strategy
   \begin{equation} \label{eq:appendARAMwt}
       w_{t+1} \leftarrow \softmax\left(-\frac{1-\beta}{\tau_w}\widehat{\mathbf{V}}_t + \beta\log w_t + (1-\beta)\log c_t\right)
   \end{equation}
   \Comment{}
   \EndFor
   \State {\bfseries Return: } $\pi_T, w_T$
\end{algorithmic}
\end{algorithm}
Here, $L_{clip}(\theta;\theta_t,w_t) = -\mathbb{E}_{(s,a)\sim d^{\pi_{\theta_t}}}[\min\{r_t(\theta)(s,a)\langle w_t,\widehat{\mathbf{A}}(s,a)\rangle, clip(r_t(\theta)(s,a),1-\epsilon_{clip},1+\epsilon_{clip})\langle w_t, \widehat{\mathbf{A}}(s,a)\rangle\} + \tau_{PPO} H(\pi_{\theta}(\cdot|s))]$ with estimated vector advantage $\widehat{\mathbf{A}}(s,a)\in\mathbb{R}^K$,
and $L_{critic}(\phi)=\mathbb{E}_{(s,\mathbf{V}_{s,target})\sim \mathcal{B}_{critic}}[\|\mathbf{V}_\phi(s)-\mathbf{V}_{s,target}\|^2]$ with the estimated return vector at $s$ replacing $\mathbf{V}_{s,target}$.

The closed-form update formula for $Adversary$'s strategy $w$ with the $D(w||c_t)$ regularizer is shown above.

\section{Supplementary Materials for Section~\ref{section_method}}\label{append:supp_section4}

\subsection{Existence of a saddle point in \eqref{maxmin_equals_minmax}}\label{append:existence_saddle}

\begin{proposition}\label{prop:maxmin_equals_minmax_implies_saddle}
    If $\max_\pi \min_{w\in\Delta^K} \langle w, \mathbf{V}_\tau^\pi\rangle 
    = \min_{w\in\Delta^K} \max_\pi \langle w, \mathbf{V}_\tau^\pi\rangle$ holds, then a saddle point of $\langle w, \mathbf{V}_\tau^\pi\rangle$ exists.
\end{proposition}
\begin{proof}
    By assumption, $\max_{\pi} \min_{w\in\Delta^K} \langle w, V_\tau^\pi\rangle 
    = \min_{w\in\Delta^K} \max_\pi \langle w, V_\tau^\pi\rangle$ -(1) holds. 
    By extreme value theorem, $w^* := \argmin_{w\in\Delta^K}\max_\pi \langle w, \mathbf{V}^\pi_\tau\rangle$ -(2) and $\pi^* := \argmax_{\pi} \min_{w\in\Delta^K} \langle w, \mathbf{V}^\pi_\tau\rangle$ -(3) exist. (Note that the policy space $\Pi = {\Delta(A)}^{|S|}$ is compact due to Tychonoff's theorem.)
    Then, 
    \begin{align}
        \langle w^*, \mathbf{V}^{\pi^*}_\tau\rangle
        \ge & \min_w \langle w, \mathbf{V}^{\pi^*}_\tau\rangle\\
        = & \max_{\pi} \min_{w} \langle w, \mathbf{V}_\tau^\pi\rangle ~(\because~(3))\\
        = & \min_{w} \max_{\pi} \langle w, \mathbf{V}_\tau^\pi\rangle ~(\because~(1))\\
        = & \max_\pi \langle w^*, \mathbf{V}^{\pi}_\tau\rangle ~(\because~(2))\\
        \ge & \langle w^*, \mathbf{V}^{\pi^*}_\tau\rangle.
    \end{align}
    Therefore, 
    \begin{equation}
        \langle w^*, \mathbf{V}^{\pi}_\tau\rangle 
        \le \max_\pi \langle w^*, \mathbf{V}^{\pi}_\tau\rangle 
        = \langle w^*, \mathbf{V}^{\pi^*}_\tau\rangle
        = \min_w \langle w, \mathbf{V}^{\pi^*}_\tau\rangle
        \le \langle w, \mathbf{V}^{\pi^*}_\tau\rangle,~\forall w,\pi.
    \end{equation}
    This implies that $(w^*,\pi^*)$ is a saddle point of $\langle w, \mathbf{V}^{\pi}_\tau\rangle$, which concludes the existence of a saddle point of $\langle w, \mathbf{V}^{\pi}_\tau\rangle$.
\end{proof}


\subsection{Proof for Theorem \ref{thm:saddle_is_maxminmorl}}\label{section:saddle_pf}
\begin{proof}
    Let $(\Bar{\pi},\Bar{w})$ be a saddle point in (\ref{maxmin_equals_minmax}) and for any $w\in\Delta^K$, define $\pi^*(w):=argmax_\pi \langle w,\mathbf{V}^\pi_\tau\rangle$. 
    Let $w^*$ be the optimal solution of minimization reformulation of entropy-regularized max-min MORL (\ref{eq:P2'_again}). 
    
    By Theorem 4.2 in \citet{park}, if $w$ is any optimal solution of (\ref{eq:P2'_again}), whose objective function is $V^*_{\cdot,\tau}\triangleq V^{\pi^*(\cdot)}_{\cdot,\tau}=\langle \cdot, V^{\pi^*(\cdot)}_\tau \rangle$ with the domain $\Delta^K$,     
    the policy defined by $\pi_w(a|s):=\exp\left(\frac{Q^*_{w,\tau}(s,a)-V^*_{w,\tau}(s)}{\tau}\right)\triangleq\exp\left(\frac{Q^{\pi^*(w)}_{w,\tau}(s,a)-V^{\pi^*(w)}_{w,\tau}(s)}{\tau}\right)$ is an optimal solution of entropy-regularized max-min MORL (\ref{maxmin_ent_target_problem}) where  
    $Q^*_{w,\tau}$ and $V^*_{w,\tau}$ are soft optimal values for scalar reward $\langle w,\mathbf{r}\rangle$. 
    Therefore, we show that $\Bar{w}$ is an optimal solution of (\ref{eq:P2'_again}).
    This concludes that the induced policy $\pi_{\Bar{w}}(a|s):=\exp\left(\frac{Q^*_{\Bar{w},\tau}(s,a)-V^*_{\Bar{w},\tau}(s)}{\tau}\right)$ is an optimal solution of entropy-regularized max-min MORL (\ref{maxmin_ent_target_problem}).
    
    By the definition of saddle point,
    \begin{equation}\label{eq:def_saddle}
        \langle \Bar{w},\mathbf{V}^\pi_\tau\rangle 
        \le \langle \Bar{w},\mathbf{V}^{\Bar{\pi}}_\tau\rangle 
        \le \langle w,\mathbf{V}^{\Bar{\pi}}_\tau\rangle, ~~\forall w,\pi.
    \end{equation}
    In particular, $\Bar{\pi} = \pi^*(\Bar{w})$ by (\ref{eq:def_saddle}), i.e. $\Bar{\pi}$ is soft optimal policy for scalar reward $\langle\Bar{w},\mathbf{r}\rangle$.
    Then, 
    \begin{align}
        \langle \Bar{w}, \mathbf{V}^{\Bar{\pi}}_\tau \rangle 
        = & \min_{w'} \langle w', \mathbf{V}^{\Bar{\pi}}_\tau \rangle~(\because (\ref{eq:def_saddle}))\\
        = & \min_{w'} \langle w', \mathbf{V}^{\pi^*(\Bar{w})}_\tau \rangle~(\because \Bar{\pi} = \pi^*(\Bar{w}))\\
        \le & \langle w^*, \mathbf{V}^{\pi^*(\Bar{w})}_\tau \rangle\\
        \le & \langle w^*, \mathbf{V}^{\pi^*(w^*)}_\tau \rangle\\
        \le & \langle \Bar{w}, \mathbf{V}^{\pi^*(\Bar{w})}_\tau \rangle\\
        &(\because w^*\text{ is an optimal solution of (\ref{eq:P2'_again}) with objective function $V^*_{\cdot,\tau}\triangleq V^{\pi^*(\cdot)}_{\cdot,\tau}=\langle \cdot, V^{\pi^*(\cdot)}_\tau \rangle$ })\\
        = & \langle \Bar{w}, \mathbf{V}^{\Bar{\pi}}_\tau \rangle~(\because \Bar{\pi} = \pi^*(\Bar{w})).
    \end{align}
    Therefore, $\langle \Bar{w}, \mathbf{V}^{\Bar{\pi}}_\tau \rangle=\langle w^*, \mathbf{V}^{\pi^*(w^*)}_\tau \rangle$, which means that $\Bar{w}$ is an optimal solution of (\ref{eq:P2'_again}).
    Then, for any $s,a$,
    \begin{align}
        \pi_{\Bar{w}}(a|s)=&\exp\left(\frac{Q^*_{\Bar{w},\tau}(s,a)-V^*_{\Bar{w},\tau}(s)}{\tau}\right)\\
    = &\exp\left(\frac{Q^{\pi^*(\Bar{w})}_{\Bar{w},\tau}(s,a)-V^{\pi^*(\Bar{w})}_{\Bar{w},\tau}(s)}{\tau}\right)~(\because\text{definition of }\pi^*(\cdot))\\
    = & \exp\left(\frac{Q^{\Bar{\pi}}_{\Bar{w},\tau}(s,a)-V^{\Bar{\pi}}_{\Bar{w},\tau}(s)}{\tau}\right)~(\because \Bar{\pi} = \pi^*(\Bar{w}))\\
    = & \Bar{\pi}(a|s)~(\because \Bar{\pi}\text{ is soft optimal policy for scalar reward }\langle\Bar{w},\mathbf{r}\rangle).
    \end{align}
    Therefore, $\Bar{\pi}$ is an optimal solution of entropy-regularized max-min MORL (\ref{maxmin_ent_target_problem}).
\end{proof}

\subsection{Remark for replacing MD with NPG}

\begin{remark}\label{rmk:npg_resembles_md}
    The MD update rule (\ref{eq:w_closedform}) can be written as 
    \begin{equation}
        w_{t+1}{(k)} \propto (w_{t}{(k)})^\beta \exp{\left(-\frac{1-\beta}{\tau_w}V^{\pi_t}_k\right)},~~\forall k=1,\ldots,K .\label{eq:npg_md_equiv}
    \end{equation}
    Recall that the NPG update with softmax policy \eqref{eq:NPG111} is 
    \begin{equation}
        \pi_{\theta_{t+1}}(a|s) 
        \propto (\pi_{\theta_{t}}(a|s))^\alpha \exp \left(\frac{1-\alpha}{\tau}Q_\tau^{\pi_{\theta_t}}(s,a)\right).
    \end{equation}
    We can observe that the update rules for MD and NPG have similar structures under the softmax policy family. This gives an intuition for why replacing MD on $\theta$ (\ref{eq:md_theta}) with NPG does not appear to be absurd.
\end{remark}

\section{ Proof of Theorem \ref{thm:convergence_exact}}
\label{convergence_pf_exact}

Note that the strategy of $Learner$ is a policy parameter $\theta_t$, it suffice to analyze the convergence of resulting policy $\pi_t:=\pi_{\theta_t}$. 
Similar to \cite{npg_ent}, we define auxiliary variables that are needed for our proof.
\begin{align}
    \xi_0(s,a) := &\|\exp(Q^{\pi^*}_{w^*,\tau}(s,\cdot)/\tau)\|_1\pi_0(a|s)\\
    \xi_{t+1}(s,a) := & (\xi_{t}(s,a))^\alpha e^{\frac{1-\alpha}{\tau}Q^{\pi_t}_{w_t,\tau}(s,a)} \label{eq:xi_update}\\ 
    \kappa_0(k)  = &\|\exp(-\mathbf{V}^{\pi^*}_\tau/\tau_w)\|_1 w_0(k)\\
    \kappa_{t+1}(k)  = &(\kappa_t(k))^\beta e^{-\frac{1-\beta}{\tau_w}\mathbf{V}^{\pi_t}_\tau} \label{eq:kappa_update}\\
    (\pi^*,w^*):& \text{ Nash equilibrium in }\mathcal{RG}\\
    Q^*_{w,\tau} := & Q^{\pi^*(w)}_{w,\tau} \triangleq \max_\pi Q^{\pi}_{w,\tau}~(\text{soft optimal $Q$-value w.r.t. reward $\langle w,\mathbf{r}\rangle$})\\
    V^*_{w,\tau} := &V^{\pi^*(w)}_{w,\tau} \triangleq \max_\pi V^{\pi}_{w,\tau}~(\text{soft optimal $V$-value w.r.t. reward $\langle w,\mathbf{r}\rangle$})
\end{align}

In words, $\xi_t(s,\cdot)$ and $\kappa_t$ are unnormalized $\pi_t(\cdot|s)$ and $w$, respectively. That is, $\pi_t(a|s) = \xi_t(s,a)/\|\xi_t(s,\cdot)\|_1,~\forall t,s,a$ and $w_t(k) = \kappa_t(k)/\|\kappa_t\|_1,~\forall t,k$.

\subsection{Introducing optimality gaps}

\textbf{Optimality gap for $\pi$}\\
We note the 1-Lipschitzness of log-sum-exponential (lse) function, which was also mentioned in \cite{npg_ent}.\\
Define the function $lse(x):=\log\sum_{i=1}^d e^{x_i}$. For $x,y\in\mathbb{R}^d$,
\begin{align*}
    |lse(x)-lse(y)| = & |\langle \nabla lse(z), x-y\rangle|\\
    & (\text{where $z=tx+(1-t)y$ for some $t\in(0,1)$, by mean value theorem})\\
    = & |\langle \softmax(z), x-y\rangle| \\
    \le & \|x-y\|_\infty \sum_{i=1}^d \frac{e^{z_i}}{\sum_{j=1}^d e^{z_j}} = \|x-y\|_\infty.
\end{align*}

Then, for each $s,a$, 
\begin{align*}
    |\log\pi^*(a|s) - \log \pi_{t}(a|s)| 
    = & |(Q^{\pi^*}_{w^*,\tau}(s,a)/\tau-V^{\pi^*}_{w^*,\tau}(s)/\tau) - (\log \xi_{t}(s,a) - \log \|\xi_{t}(s,\cdot)\|_1)|\\
    = & |(Q^{\pi^*}_{w^*,\tau}(s,a)/\tau-\log \xi_{t}(s,a)) - (V^{\pi^*}_{w^*,\tau}(s)/\tau - \log \|\xi_{t}(s,\cdot)\|_1)|\\
    \le & |Q^{\pi^*}_{w^*,\tau}(s,a)/\tau-\log \xi_{t}(s,a)|\\
    +& | lse(Q^{\pi^*}_{w^*,\tau}(s,\cdot)/\tau) - lse(\log \xi_{t}(s,a))|\\
    & (\because V^{\pi^*}_{w^*,\tau}(s)/\tau = lse(Q^{\pi^*}_{w^*,\tau}(s,\cdot)/\tau)\text{ holds for soft optimal value})\\
    \le & |Q^{\pi^*}_{w^*,\tau}(s,a)/\tau-\log \xi_{t}(s,a)| + | Q^{\pi^*}_{w^*,\tau}(s,a)/\tau - \log \xi_{t}(s,a)|\\
    &(\because \text{1-Lipschitzness of }lse)\\
    = & 2|Q^{\pi^*}_{w^*,\tau}(s,a)/\tau-\log \xi_{t}(s,a)|
\end{align*}
where the first equality utilized the fact that $\pi^*=RBR(w^*)$ (regularized best-response), i.e. soft optimal policy with respect to reward $\langle w^*,\mathbf{r}\rangle$, and
the property of soft value function and soft optimal policy $\pi^*(a|s) = 
\exp\left(\frac{Q^{\pi^*}_{w^*,\tau}(s,a) - V^{\pi^*}_{w^*,\tau}(s)}{\tau}\right)$ established in \cite{soft_opt_val_pol_rel}.

Therefore,
\begin{align}
    \|\log\pi^* - \log \pi_{t}\|_\infty\le 
    \frac{2}{\tau}\|Q^{\pi^*}_{w^*,\tau}-\tau\log \xi_{t}\|_\infty. \label{eq:policy_bound_exact}
\end{align}

In words, the term $\|Q^{\pi^*}_{w^*,\tau}-\tau\log \xi_t\|_\infty$ bounds the gap between policy $\pi_{t}$ in $t$-th iteration and Nash policy $\pi^*$.

\textbf{Optimality gap for $w$}

\begin{lemma}\label{lem:log_bd}
    $\|w-w'\|_\infty \le \|\log w - \log w'\|_\infty,~\forall w,w'\in \text{int}(\Delta^K)$ where the logarithm is taken element-wisely and $\text{int}(D)$ denotes interior of set $D$.
\end{lemma}
\begin{proof}
    Let $x,y\in int(\Delta^K)$, interior of simplex, and define $f(x):=[\log x_1,\ldots,\log x_K]^T$ for any $x\in int(\Delta^K)$.
    Let $a = f(x)$, $b = f(y)$ and $g(x):=[e^{x_1},\ldots,e^{x_K}]^T$ for any $x\in\mathbb{R}^K$.
    By the mean value theorem for multi-variate function $g$,
    \begin{align*}
        g(b)-g(a) = J_g(c)(b-a)
    \end{align*}
    where $J_g(c)$ is Jacobian matrix of $g$ at $c$ and $c=ta+(1-t)b,~t\in(0,1)$. Specifically, the Jacobian matrix becomes a diagonal matrix $J_g(c) = diag(e^{c_1},\ldots,e^{c_K})$.
    Then, 
    \begin{align*}
        \|g(b)-g(a)\|_\infty = \|J_g(c)(b-a)\|_\infty
        \le & \|J_g(c)\|_\infty\|b-a\|_\infty\\
        = & \max\{e^{c_1},\ldots,e^{c_K}\}\|b-a\|_\infty
    \end{align*}
    Here, since $x_i,y_i<1~(\because x,y\in int(\Delta^K))$, $a_i = \log x_i<0$ and $b_i = \log y_i<0$. Hence, $c_i=ta_i+(1-t)b_i<0$.
    These results in $\max\{e^{c_1},\ldots,e^{c_K}\}<1$. By plugging in $a = f(x)$, $b = f(y)$ above, we conclude that $\|y-x\|_\infty < \|\log y -\log y\|_\infty$ for $x,y\in int(\Delta^K)$. 
\end{proof}

Due to the lemma \ref{lem:log_bd}, it suffice to analyze $\|\log w^* - \log w_t\|_\infty$ instead of $\|w^* - w_t\|_\infty$. \\
Note that $w^*$ is the best-response to $\pi^*$ with respect to $u^\mathcal{RG}$ since $(\pi^*,w^*)$ is Nash equilibrium in $\mathcal{RG}$, and $w^*$ can be written as below. 
\begin{align}
    w^* = & \argmax_{w\in\Delta^K}~ -\langle w,\mathbf{V}^{\pi^*}_\tau\rangle + \tau_w H(w)\\
    = & \softmax\left(-\frac{1}{\tau_w}\mathbf{V}^{\pi^*}_\tau\right)
\end{align}

Using this,
\begin{align}
    \log w^{*}(k) - \log w_t(k)
    = & \log \frac{\exp\left(-\frac{1}{\tau_w}V^{\pi^*}_{k,\tau}\right)}{\|\exp\left(-\frac{1}{\tau_w}\mathbf{V}^{\pi^*}_{\tau}\right)\|_1} - \log \frac{\kappa_t(k)}{\|\kappa_t\|_1}\\
    = & -\frac{1}{\tau_w}V^{\pi^*}_{k,\tau} - \log \kappa_t(k)
    -(lse(-\frac{1}{\tau_w}\mathbf{V}^{\pi^*}_{\tau}) - lse(\log \kappa_t))
\end{align}
By taking absolute value for both side and applying triangular inequality, 
\begin{align}
    |\log w^{*}(k) - \log w_t(k)|
    \le & ~|-\frac{1}{\tau_w}V^{\pi^*}_{k,\tau} - \log \kappa_t(k)|
    + |lse(-\frac{1}{\tau_w}\mathbf{V}^{\pi^*}_{\tau}) - lse(\log \kappa_t)|\\
    \le & ~|-\frac{1}{\tau_w}V^{\pi^*}_{k,\tau} - \log \kappa_t(k)| + \|-\frac{1}{\tau_w}\mathbf{V}^{\pi^*}_{\tau} - \log \kappa_t\|_\infty\\
    &(\because \text{1-Lipschitzness of }lse)
\end{align}
By taking $max_k$ for both sides, obtain the following bound. 
\begin{align}
    \|\log w^* - \log w_t\|_\infty \le \frac{2}{\tau_w}\|-\mathbf{V}^{\pi^*}_{\tau} - \tau_w\log \kappa_t\|_\infty \label{eq:w_bound_exact}
\end{align}
In particular, combining this bound with the lemma \ref{lem:log_bd}, obtain the following bounds.
\begin{align}
    \|w^* - w_t\|_\infty 
    \le \|\log w^* - \log w_t\|_\infty 
    \le \frac{2}{\tau_w}\|-\mathbf{V}^{\pi^*}_{\tau} - \tau_w\log \kappa_t\|_\infty \label{eq:w_bound_exact_full}
\end{align}

\textbf{Optimality gap for soft $Q$}\\
The optimality gap for soft $Q$ at iteration $t$ can be expressed as follows.
\begin{equation}
    \|Q^{\pi^*}_{w^*,\tau} - Q^{\pi_t}_{w_t,\tau}\|_\infty
\end{equation}

\textbf{Supplementary term}\\
Following \cite{npg_ent}, we use the following supplementary term which helps to analyze the optimality gap for soft $Q$.
\begin{equation}
    \max\{0,-\min_{s,a} (Q^{\pi_t}_{w_t,\tau} - \tau\log\xi_t)\}
\end{equation}

\subsection{Recursive bounds for optimality gaps and supplementary term}
We establish recursive bounds for the following three optimality gaps and the supplementary term:\\
$\|Q^{\pi^*}_{w^*,\tau}-\tau\log \xi_{t}\|_\infty$ (optimality gap for policy (\ref{eq:policy_bound_exact})), $\|Q^{\pi^*}_{w^*,\tau}-Q^{\pi_t}_{w_t,\tau}\|_\infty$ (optimality gap for soft $Q$ function),
 $\|-\mathbf{V}^{\pi^*}_{\tau} - \tau_w\log \kappa_t\|_\infty$ (optimality gap for $w$ (\ref{eq:w_bound_exact_full}))
 and $\max\{0,-\min_{s,a} (Q^{\pi_t}_{w_t,\tau} - \tau\log\xi_t)\}$ (supplementary term, used for optimality gap for soft $Q$ function).
After then, following \cite{npg_ent}, we establish linear system with these optimality gaps to show convergence.

\textbf{Recursive bounds: optimality gap for policy}
For each $(s,a)$,
\begin{align}
    Q^{\pi^*}_{w^*,\tau}(s,a)-\tau\log\xi_{t+1}(s,a) & = Q^{\pi^*}_{w^*,\tau}(s,a) - \tau\alpha\log\xi_{t}(s,a) - (1-\alpha)Q^{\pi_t}_{w_t,\tau}(s,a)~(\because (\ref{eq:xi_update}))\\
     & = \alpha(Q^{\pi^*}_{w^*,\tau}(s,a) - \tau\log\xi_{t}(s,a))\\
    & + (1-\alpha)(Q^{\pi^*}_{w^*,\tau}(s,a) - Q^{\pi_t}_{w_t,\tau}(s,a))
\end{align}
Thus, 
\begin{align}
    \|Q^{\pi^*}_{w^*,\tau}-\tau\log\xi_{t+1}\|_\infty & \le \alpha\|Q^{\pi^*}_{w^*,\tau} - \tau\log\xi_t\|_\infty + (1-\alpha)\|Q^{\pi^*}_{w^*,\tau} - Q^{\pi_t}_{w_t,\tau}\|_\infty \label{eq:pi_gap_exact}
\end{align}

\textbf{Recursive bounds: optimality gap for soft $Q$ function}
Since $\pi$ (thus, $\theta$) is a max-player and $w$ is a min-player in our two-player zero-sum regularized game, we cannot guarantee monotonicity of $Q^{\pi_{t}}_{w_{t},\tau}(s,a)$ in $t$.
Thus, we establish both upper and lower bounds for $Q^{\pi^*}_{w^*,\tau}(s,a) - Q^{\pi_{t+1}}_{w_{t+1},\tau}(s,a)$ and these will offer an upper bound for $|Q^{\pi^*}_{w^*,\tau}(s,a) - Q^{\pi_{t+1}}_{w_{t+1},\tau}(s,a)|$, thus upper bound for $\|Q^{\pi^*}_{w^*,\tau}(s,a) - Q^{\pi_{t+1}}_{w_{t+1},\tau}(s,a)\|_\infty$.

1) Upper bound\\
For each $s,a$,
\begin{align}
    &Q^{\pi^*}_{w^*,\tau}(s,a) - Q^{\pi_{t+1}}_{w_{t+1},\tau}(s,a)\\
    =& \langle w^*, \mathbf{r}(s,a)\rangle + \gamma \mathbb{E}_{s'\sim P(\cdot|s,a)}[V^{\pi^*}_{w^*,\tau}(s')] - \langle w_{t+1}, r(s,a)\rangle - \gamma \mathbb{E}_{s'\sim P(\cdot|s,a)}[V^{\pi_{t+1}}_{w_{t+1},\tau}(s')]\\
    =& \langle w^*-w_{t+1}, \mathbf{r}(s,a)\rangle + \gamma \mathbb{E}_{s'\sim P(\cdot|s,a)}[V^{*}_{w^*,\tau}(s')] \\
     - & \gamma \mathbb{E}_{s'\sim P(\cdot|s,a)}\mathbb{E}_{a'\sim\pi_{t+1}(\cdot|s')}[Q^{\pi_{t+1}}_{w_{t+1},\tau}(s',a')-\tau\log\pi_{t+1}(s',a')]\\
    =& \langle w^*-w_{t+1}, \mathbf{r}(s,a)\rangle + \gamma \mathbb{E}_{s'\sim P(\cdot|s,a)}[\tau\log\|e^{Q^{*}_{w^*,\tau}(s',\cdot)/\tau}\|_1]\\
    -& \gamma \mathbb{E}_{s'\sim P(\cdot|s,a)}\mathbb{E}_{a'\sim\pi_{t+1}(\cdot|s')}\left[Q^{\pi_{t+1}}_{w_{t+1},\tau}(s',a')-\tau\log\frac{\xi_{t+1}(s',a')}{\|\xi_{t+1}(s',\cdot)\|_1}\right] \label{eq:for_inexact_q_ub}\\
    = & \langle w^*-w_{t+1}, \mathbf{r}(s,a)\rangle \label{eq:rwd_term}\\
    +& \gamma \mathbb{E}_{s'\sim P(\cdot|s,a)}[\tau\log\|e^{Q^{\pi^*}_{w^*,\tau}(s',\cdot)/\tau}\|_1 - \tau\log \|\xi_{t+1}(s',\cdot)\|_1]~~-(A) \label{eq:A}\\
    -& \gamma \mathbb{E}_{s'\sim P(\cdot|s,a),a'\sim\pi_{t+1}(\cdot|s')}[Q^{\pi_{t+1}}_{w_{t+1},\tau}(s',a') - \tau\log\xi_{t+1}(s',a')]~~-(B) \label{eq:B}
\end{align}
In the third line, 
the second term  used $V^{\pi^*}_{w^*,\tau} = V^{*}_{w^*,\tau}$ since 
$\pi^*$ is the best-response to $w^*$ with respect to $u^\mathcal{RG}$
and the third term used soft Bellman equation. 
In the fourth line, 
used $V^{*}_{w^*,\tau}(s')=\tau\log\|e^{Q^{*}_{w^*,\tau}(s',\cdot)/\tau}\|_1$ which holds for soft optimal value function.

We can easily upper bound the line (\ref{eq:rwd_term}), 
$\langle w^*-w_{t+1}, \mathbf{r}(s,a)\rangle$, by 
$Kr_{max}\|w^*-w_{t+1}\|_\infty$ due to $H\Ddot{o}lder$ inequality and $\|\mathbf{r}(s,a)\|_1\le Kr_{max}~\forall s,a$ where $r_{max} = \max_{s,a,k} |r_k(s,a)|$.
By (\ref{eq:w_bound_exact_full}), 
\begin{align}
    & \langle w^*-w_{t+1}, \mathbf{r}(s,a)\rangle\\
    \le & Kr_{max}\|w^*-w_{t+1}\|_\infty\\
    \le & \frac{2Kr_{max}}{\tau_w}\|-\mathbf{V}^{\pi^*}_{\tau} - \tau_w\log \kappa_{t+1}\|_\infty \label{eq:r_bound}
\end{align}

From now, we derive the upper bound for $(A)$.

For each $s$,
\begin{align}
    &\tau\log\|e^{Q^{\pi^*}_{w^*,\tau}(s,\cdot)/\tau}\|_1 - \tau\log \|\xi_{t+1}(s,\cdot)\|_1\\
    =&\tau( lse(Q^{\pi^*}_{w^*,\tau}(s,\cdot)/\tau) - lse(\log\xi_{t+1}(s,\cdot)) )\\
    \le& \tau \| Q^{\pi^*}_{w^*,\tau}(s,\cdot)/\tau - \log\xi_{t+1}(s,\cdot) \|_\infty~(\because \text{1-Lipschitzness of }lse)\\
    =&\| Q^{\pi^*}_{w^*,\tau}(s,\cdot) - \tau\log\xi_{t+1}(s,\cdot) \|_\infty\\
    \le& \| Q^{\pi^*}_{w^*,\tau} - \tau\log\xi_{t+1}\|_\infty 
\end{align} 
Therefore, 
\begin{align}
    (A) = &\gamma \mathbb{E}_{s'\sim P(\cdot|s,a)}[\tau\log\|e^{Q^{\pi^*}_{w^*,\tau}(s',\cdot)/\tau}\|_1 - \tau\log \|\xi_{t+1}(s',\cdot)\|_1]\\
    \le & \gamma \mathbb{E}_{s'\sim P(\cdot|s,a)}[\| Q^{\pi^*}_{w^*,\tau} - \tau\log\xi_{t+1}\|_\infty]\\
    = & \gamma\| Q^{\pi^*}_{w^*,\tau} - \tau\log\xi_{t+1}\|_\infty \label{eq:A_bound}
\end{align}

From now, we derive the upper bound for $(B)$.

To establish the upper bound, following \citet{npg_ent} and \citet{fednpg}, we use supplementary term. For each $s,a$,
\begin{align}
    & Q^{\pi_{t+1}}_{w_{t+1},\tau}(s,a) - \tau\log\xi_{t+1}(s,a)\\
    \ge & \min_{s,a}\{Q^{\pi_{t+1}}_{w_{t+1},\tau}(s,a) - \tau\log\xi_{t+1}(s,a)\},
\end{align}
then,
\begin{align}
    &-(Q^{\pi_{t+1}}_{w_{t+1},\tau}(s,a) - \tau\log\xi_{t+1}(s,a))\\
    \le & - \min_{s,a}\{Q^{\pi_{t+1}}_{w_{t+1},\tau}(s,a) - \tau\log\xi_{t+1}(s,a)\}\\
    \le & \max\{0,- \min_{s,a}\{Q^{\pi_{t+1}}_{w_{t+1},\tau}(s,a) - \tau\log\xi_{t+1}(s,a)\}\}.
\end{align}
Using the last term, establish the upper bound for (B) as follows.
\begin{align}
    (B) = & - \gamma \mathbb{E}_{s'\sim P(\cdot|s,a),a'\sim\pi_{t+1}(\cdot|s')}[Q^{\pi_{t+1}}_{w_{t+1},\tau}(s',a') - \tau\log\xi_{t+1}(s',a')]\\
    \le & \gamma \mathbb{E}_{s'\sim P(\cdot|s,a),a'\sim\pi_{t+1}(\cdot|s')}[
    \max\{0,- \min_{s,a}\{Q^{\pi_{t+1}}_{w_{t+1},\tau}(s,a) - \tau\log\xi_{t+1}(s,a)\}\}
    ]\\
    = & \gamma \max\{0,- \min_{s,a}\{Q^{\pi_{t+1}}_{w_{t+1},\tau}(s,a) - \tau\log\xi_{t+1}(s,a)\}\} \label{eq:B_bound}
\end{align}

Combining (\ref{eq:r_bound}), (\ref{eq:A_bound}) and (\ref{eq:B_bound}), obtain the upper bound for optimality gap for soft $Q$ as follows. \\
For each $s,a$,
\begin{align}
     &Q^{\pi^*}_{w^*,\tau}(s,a) - Q^{\pi_{t+1}}_{w_{t+1},\tau}(s,a) \label{eq:Q_ub}\\
     \le & \frac{2Kr_{max}}{\tau_w}\|-\mathbf{V}^{\pi^*}_{\tau} - \tau_w\log \kappa_{t+1}\|_\infty
     + \gamma\| Q^{\pi^*}_{w^*,\tau} - \tau\log\xi_{t+1}\|_\infty\\
     + &\gamma \max\{0,- \min_{s,a}\{Q^{\pi_{t+1}}_{w_{t+1},\tau}(s,a) - \tau\log\xi_{t+1}(s,a)\}\}\\
      =:& UB_1 
\end{align}

As mentioned, we derive the lower bound for $Q^{\pi^*}_{w^*,\tau}(s,a) - Q^{\pi_t}_{w_t,\tau}(s,a)$ and finally combine with the upper bound (\ref{eq:Q_ub}) to conclude the upper bound for $\|Q^{\pi^*}_{w^*,\tau}(s,a) - Q^{\pi_t}_{w_t,\tau}(s,a)\|_\infty$.

\begin{align}
    &Q^{\pi^*}_{w^*,\tau}(s,a) - Q^{\pi_{t+1}}_{w_{t+1},\tau}(s,a) \label{eq:Q_lb}\\
    =&\langle w^*-w_{t+1}, \mathbf{r}(s,a)\rangle + \gamma \mathbb{E}_{s'}[V^{\pi^*}_{w^*,\tau}(s') - V^{\pi_{t+1}}_{w_{t+1},\tau}(s')]\\
    =&\langle w^*-w_{t+1}, \mathbf{r}(s,a)\rangle + \gamma \mathbb{E}_{s'}[\langle w^*, V^{\pi^*}_{\tau}(s')\rangle - \langle w_{t+1}-w^*+w^*, V^{\pi_{t+1}}_{\tau}(s')\rangle]\\
    =&\langle w^*-w_{t+1}, \mathbf{r}(s,a)\rangle + \gamma \mathbb{E}_{s'}[\langle w^*, V^{\pi^*}_{\tau}(s')-V^{\pi_{t+1}}_{\tau}(s')\rangle + \langle w^* - w_{t+1}, V^{\pi_{t+1}}_{\tau}(s')\rangle]\\
    =&\langle w^*-w_{t+1}, \mathbf{r}(s,a)+\gamma \mathbb{E}_{s'}[V^{\pi_{t+1}}_{\tau}(s')]\rangle + \gamma \mathbb{E}_{s'}[V^{\pi^*}_{w^*,\tau}(s')-V^{\pi_{t+1}}_{w^*,\tau}(s')]\\
    =&\langle w^*-w_{t+1}, Q_\tau^{\pi_{t+1}}(s,a)\rangle + \gamma \mathbb{E}_{s'}[V^{\pi^*}_{w^*,\tau}(s')-V^{\pi_{t+1}}_{w^*,\tau}(s')]\\
    \ge&-Q_{\tau, max,l_1}\|w^*-w_{t+1}\|_\infty+0\\
    \ge & -Q_{\tau, max,l_1} \frac{2}{\tau_w} \|-\mathbf{V}^{\pi^*}_{\tau} - \tau_w\log \kappa_{t+1}\|_\infty ~ (\because (\ref{eq:w_bound_exact_full}))\\
    =: & -UB_2 
\end{align}
where $Q_{\tau, max,l_1} = \max_{\pi,s,a}\|Q_{\tau}^\pi(s,a)\|_1 = \frac{K(r_{max} + \tau\log|A|)}{1-\gamma}$. 
In the first inequality, the first term is from H$\Ddot{o}$lder inequality and the second term is from $V^{\pi^*}_{w^*,\tau}(s)\ge V^{\pi_{t+1}}_{w^*,\tau}(s),~\forall s$ since $(\pi^*,w^*)$ is a Nash equilibrium in $\mathcal{RG}$.
For simplicity, we denote $Q_{\tau,max} = \frac{r_{max} + \tau\log|A|}{1-\gamma}$.

From $-UB_2\le Q^{\pi^*}_{w^*,\tau}(s,a) - Q^{\pi_{t+1}}_{w_{t+1},\tau}(s,a) \le UB_1$ (from (\ref{eq:Q_ub}) and (\ref{eq:Q_lb})),\\
obtain $\|Q^{\pi^*}_{w^*,\tau} - Q^{\pi_{t+1}}_{w_{t+1},\tau}\|_\infty\le\max\{UB_1,UB_2\}$. 

Note that since $r_{max}\le Q_{\tau,max}$, $UB_2$ dominates the first term in $UB_1$.
By combining the upper bound (\ref{eq:Q_ub}) and the lower bound (\ref{eq:Q_lb}), conclude the recursive bound for optimality gap of soft $Q$ function as follows.
\begin{align}
    &\|Q^{\pi^*}_{w^*,\tau} - Q^{\pi_{t+1}}_{w_{t+1},\tau}\|_\infty \label{eq:q_gap_exact}\\
    \le & \frac{2KQ_{\tau,max}}{\tau_w}\|-\mathbf{V}^{\pi^*}_{\tau} - \tau_w\log \kappa_{t+1}\|_\infty
     + \gamma\| Q^{\pi^*}_{w^*,\tau} - \tau\log\xi_{t+1}\|_\infty\\
     +& \gamma \max\{0,- \min_{s,a}\{Q^{\pi_{t+1}}_{w_{t+1},\tau}(s,a) - \tau\log\xi_{t+1}(s,a)\}\}
\end{align}
Note that the terms in the right-hand side, which are the terms for $t+1$, will be boiled down to the terms for $t$ by using recursive bounds.

\textbf{Recursive bounds: optimality gap for $w$}
For each $k=1,\ldots,K$,
\begin{align}
    & -V^{\pi^*}_{k,\tau} - \tau_w\log \kappa_{t+1}(k)\\
    = & \beta(-V^{\pi^*}_{k,\tau}) + (1-\beta)(-V^{\pi^*}_{k,\tau}) - \tau_w\beta\log\kappa_t(k) -(1-\beta)(-V^{\pi_t}_{k,\tau})~(\because (\ref{eq:kappa_update}))\\
    = & \beta(-V^{\pi^*}_{k,\tau} - \tau_w\log\kappa_t(k)) + (1-\beta)(-V^{\pi^*}_{k,\tau}+V^{\pi_t}_{k,\tau}).
\end{align}
Thus,
\begin{align}
    & |-V^{\pi^*}_{k,\tau} - \tau_w\log \kappa_{t+1}(k)|\\
    \le & \beta|-V^{\pi^*}_{k,\tau} - \tau_w\log\kappa_t(k)| + (1-\beta)|-V^{\pi^*}_{k,\tau}+V^{\pi_t}_{k,\tau}|\\
    \le & \beta |-V^{\pi^*}_{k,\tau} - \tau_w\log\kappa_t(k)| 
    + (1-\beta)\frac{M}{\tau}\|Q^{\pi^*}_{w^*,\tau}-\tau\log\xi_t\|_\infty
\end{align}
where the last inequality is due to the lemma 15 in \citet{fednpg}, applying with 
$\pi^*(\cdot|s) = \softmax(Q^{\pi^*}_{w^*,\tau}(s,\cdot)/\tau)$ and $\pi_t(\cdot|s) = \softmax(\log\xi_t(s,\cdot))$ and $M = \frac{r_{max}(1+\gamma) + 2\tau(1-\gamma)\log|A|}{(1-\gamma)^2}$ following \citet{fednpg}.

By taking $\max_k$, we obtain the upper bound for optimality gap of $w$ as follows.
\begin{align}
    \|-\mathbf{V}^{\pi^*}_{\tau} - \tau_w\log \kappa_{t+1}\|_\infty
    \le \beta \|-\mathbf{V}^{\pi^*}_{\tau} - \tau_w\log \kappa_{t}\|_\infty + (1-\beta)\frac{M}{\tau}\|Q^{\pi^*}_{w^*,\tau}-\tau\log\xi_t\|_\infty \label{eq:w_gap_exact}
\end{align}

\textbf{Recursive bounds: supplementary term}
\begin{align}
    &Q^{\pi_{t+1}}_{w_{t+1},\tau}(s,a) - \tau\log\xi_{t+1}(s,a)\\
    = & Q^{\pi_{t+1}}_{w_{t+1},\tau}(s,a) - \tau\alpha\log\xi_{t}(s,a) - (1-\alpha)Q^{\pi_{t}}_{w_{t},\tau}(s,a)~(\because(\ref{eq:xi_update}))\\
    = &  Q^{\pi_{t+1}}_{w_{t+1},\tau}(s,a) -  Q^{\pi_{t}}_{w_{t},\tau}(s,a)
    + \alpha (Q^{\pi_{t}}_{w_{t},\tau}(s,a) -\tau \log\xi_{t}(s,a)) \label{eq:supp_bd_intermediate}
\end{align}

To establish a lower bound for the term $Q^{\pi_{t+1}}_{w_{t+1},\tau}(s,a) -  Q^{\pi_{t}}_{w_{t},\tau}(s,a)$,
\begin{align}
    &Q^{\pi_{t+1}}_{w_{t+1},\tau}(s,a) -  Q^{\pi_{t}}_{w_{t},\tau}(s,a) \label{eq:q_difference_start}\\
    = & (Q^{\pi_{t+1}}_{w_{t+1},\tau}(s,a) - Q^{\pi_{t+1}}_{w_{t},\tau}(s,a))
    + (Q^{\pi_{t+1}}_{w_{t},\tau}(s,a)
    -  Q^{\pi_{t}}_{w_{t},\tau}(s,a))\\
    \ge & Q^{\pi_{t+1}}_{w_{t+1},\tau}(s,a) - Q^{\pi_{t+1}}_{w_{t},\tau}(s,a)\\
    = & \langle w_{t+1}-w_t, Q^{\pi_{t+1}}_\tau(s,a) \rangle\\
    \ge & - Q_{\tau,max,l_1}\|w_{t+1}-w_t\|_\infty~(\because H\Ddot{o}lder \text{ inequality})\\
    \ge & - Q_{\tau,max,l_1}\|\log w_{t+1}-\log w_t\|_\infty~(\because \text{lemma \ref{lem:log_bd}})\\
    \ge & - 2Q_{\tau,max,l_1}\|\log \kappa_{t+1}-\log \kappa_t\|_\infty~(\because w_t = \frac{\kappa_t}{\|\kappa_t\|_1}~\forall t\text{ and 1-Lipschitzness of lse})\\
    = & -\frac{2Q_{\tau,max,l_1}(1-\beta)}{\tau_w}\|-\mathbf{V}^{\pi_t}_\tau - \tau_w \log \kappa_t \|_\infty \\
    \ge & -\frac{2Q_{\tau,max,l_1}(1-\beta)}{\tau_w} (\|-\mathbf{V}^{\pi_t}_\tau + \mathbf{V}^{\pi^*}_\tau \|_\infty
    +
    \|-\mathbf{V}^{\pi^*}_\tau - \tau_w \log \kappa_t \|_\infty)\\
    \ge & -\frac{2Q_{\tau,max,l_1}(1-\beta)}{\tau_w} (\frac{M}{\tau}\|Q^{\pi^*}_{w^*,\tau} - \tau\log\xi_t \|_\infty
    +
    \|-\mathbf{V}^{\pi^*}_\tau - \tau_w \log \kappa_t \|_\infty). \label{eq:q_difference}
\end{align}
The third line is due to performance improvement lemma of NPG with fixed reward (\citet{npg_ent}, lemma 1), which implies $Q^{\pi_{t+1}}_{w_{t},\tau}(s,a)
    -  Q^{\pi_{t}}_{w_{t},\tau}(s,a)\ge0,~\forall t,s,a$. 
For completeness, we provide the statement of the lemma below.\\

\begin{lemma}\label{lem:performance_improvement}(Performance improvement by NPG with fixed scalar reward; adaptation of Lemma 1 in \citet{npg_ent})\\
    For $0<\eta\le \frac{1-\gamma}{\tau}$,\\
    $V^{\pi_{t+1}}_{w_{t+1},\tau}-V^{\pi_{t}}_{w_{t},\tau}
    = \mathbb{E}_{s\sim d^{\pi_{t+1}}_\mu}[(\frac{1}{\eta}-\frac{\tau}{1-\gamma})D_{KL}(\pi_{t+1}(\cdot|s)\|\pi_{t}(\cdot|s))
    + \frac{1}{\eta}D_{KL}(\pi_{t}(\cdot|s)\|\pi_{t+1}(\cdot|s))]$\\
    where $d^{\pi_{t+1}}_\mu$ is a stationary distribution induced by policy $\pi_{t+1}$ and initial state distribution $\mu$,\\
    i.e., $d^{\pi_{t+1}}_\mu(s) = (1-\gamma)\sum_{s_0}\sum_{t=0}^\infty \gamma^t Pr^{\pi_{t+1}}(s_t=s|s_0)\mu(s_0)$.\\
    As a result,
    $Q^{\pi_{t+1}}_{w_{t+1},\tau}(s,a)-Q^{\pi_{t}}_{w_{t},\tau}(s,a)\ge0$ for any $s,a$.
\end{lemma}

The last inequality \eqref{eq:q_difference} is due to the lemma 15 in \citet{fednpg}, applying with 
$\pi^*(\cdot|s) = softmax(Q^{\pi^*}_{w^*,\tau}(s,\cdot)/\tau)$ and $\pi_t(\cdot|s) = softmax(\log\xi_t(s,\cdot))$.

Plugging (\ref{eq:q_difference}) into (\ref{eq:supp_bd_intermediate}) results in the following bound.
\begin{align}
    &Q^{\pi_{t+1}}_{w_{t+1},\tau}(s,a) - \tau\log\xi_{t+1}(s,a)\\
    = &  Q^{\pi_{t+1}}_{w_{t+1},\tau}(s,a) -  Q^{\pi_{t}}_{w_{t},\tau}(s,a)
    + \alpha (Q^{\pi_{t}}_{w_{t},\tau}(s,a) -\tau \log\xi_{t}(s,a))\\
    \ge & -\frac{2Q_{\tau,max,l_1}(1-\beta)}{\tau_w} (\frac{M}{\tau}\|Q^{\pi^*}_{w^*,\tau} - \tau\log\xi_t \|_\infty
    +
    \|-\mathbf{V}^{\pi^*}_\tau - \tau_w \log \kappa_t \|_\infty)\\
    + &\alpha (Q^{\pi_{t}}_{w_{t},\tau}(s,a) -\tau \log\xi_{t}(s,a)),
\end{align}
which implies 
\begin{align}
     & -(Q^{\pi_{t+1}}_{w_{t+1},\tau}(s,a) - \tau\log\xi_{t+1}(s,a))\\
    \le & \frac{2Q_{\tau,max,l_1}(1-\beta)}{\tau_w} (\frac{M}{\tau}\|Q^{\pi^*}_{w^*,\tau} - \tau\log\xi_t \|_\infty
    +
    \|-\mathbf{V}^{\pi^*}_\tau - \tau_w \log \kappa_t \|_\infty)\\
    - &\alpha (Q^{\pi_{t}}_{w_{t},\tau}(s,a) -\tau \log\xi_{t}(s,a)).
\end{align}

By taking $\max_{s,a}$ for both sides, 
\begin{align}
    & -\min_{s,a}\{Q^{\pi_{t+1}}_{w_{t+1},\tau}(s,a) - \tau\log\xi_{t+1}(s,a)\}\\
    \le & \frac{2Q_{\tau,max,l_1}(1-\beta)}{\tau_w} (\frac{M}{\tau}\|Q^{\pi^*}_{w^*,\tau} - \tau\log\xi_t \|_\infty
    +
    \|-\mathbf{V}^{\pi^*}_\tau - \tau_w \log \kappa_t \|_\infty)\\
    - &\alpha \min_{s,a}\{Q^{\pi_{t}}_{w_{t},\tau}(s,a) -\tau \log\xi_{t}(s,a)\}\\
    \le & \frac{2Q_{\tau,max,l_1}(1-\beta)}{\tau_w} (\frac{M}{\tau}\|Q^{\pi^*}_{w^*,\tau} - \tau\log\xi_t \|_\infty
    +
    \|-\mathbf{V}^{\pi^*}_\tau - \tau_w \log \kappa_t \|_\infty)\\
     +&\alpha\max\{0,-\min_{s,a}\{Q^{\pi_{t}}_{w_{t},\tau}(s,a) -\tau \log\xi_{t}(s,a)\}\}.
\end{align}
Since the last line is non-negative, 
\begin{align}
    & \max\{0,- \min_{s,a}\{Q^{\pi_{t+1}}_{w_{t+1},\tau}(s,a) - \tau\log\xi_{t+1}(s,a)\}\} \label{eq:supp_exact}\\
    \le & \alpha\max\{0,-\min_{s,a}\{Q^{\pi_{t}}_{w_{t},\tau}(s,a) -\tau \log\xi_{t}(s,a)\}\} \\
     + &\frac{2Q_{\tau,max,l_1}(1-\beta)}{\tau_w} (\frac{M}{\tau}\|Q^{\pi^*}_{w^*,\tau} - \tau\log\xi_t \|_\infty
    +
    \|-\mathbf{V}^{\pi^*}_\tau - \tau_w \log \kappa_t \|_\infty).
\end{align}

\subsection{Linear system for optimality gaps}\label{subsubsection:linear_system_exact}

For simplicity, we denote the three optimality gaps and supplementary term as follows.
\begin{align}
    G(\pi_{t}) := & \|Q^{\pi^*}_{w^*,\tau} - \tau\log\xi_t\|_\infty\\
    G(Q_{t}) := & \|Q^{\pi^*}_{w^*,\tau} - Q^{\pi_{t}}_{w_{t},\tau}\|_\infty \\
    G(w_{t}) := & \|-\mathbf{V}^{\pi^*}_{\tau} - \tau_w\log \kappa_{t}\|_\infty\\
    H_{t} := & \max\{0,-\min_{s,a} (Q^{\pi_t}_{w_t,\tau} - \tau\log\xi_t)\} 
\end{align}

The recursive bound for these $G(\pi_{t}), G(Q_{t}), G(w_{t})$ and $H_{t}$ are summarized below.
$
\left\{
\begin{aligned}
    G(Q_{t+1}) \le & \frac{2KQ_{\tau,max}}{\tau_w}G(w_{t+1})
     + \gamma G(\pi_{t+1})
     + \gamma H_{t+1} & (\because\eqref{eq:q_gap_exact})\\
    G(\pi_{t+1}) \le & \alpha G(\pi_{t}) + (1-\alpha) G(Q_{t}) & (\because\eqref{eq:pi_gap_exact})\\
    \le &(\alpha + (1-\alpha)\gamma) G(\pi_{t}) + (1-\alpha) (\frac{2KQ_{\tau,max}}{\tau_w}G(w_{t})
     + \gamma H_{t})\\
     {}&(\because\text{by the bound for $G(Q_{t+1})$ above})\\
    G(w_{t+1}) \le & \beta G(w_{t}) + (1-\beta)\frac{M}{\tau}G(\pi_{t}) & (\because\eqref{eq:w_gap_exact})\\
    H_{t+1} \le &  \alpha H_t + \frac{2KMQ_{\tau,max}(1-\beta)}{\tau\tau_w} G(\pi_t) + \frac{2KQ_{\tau,max}(1-\beta)}{\tau_w} G(w_t) & (\because\eqref{eq:supp_exact})
\end{aligned}
\right.
$\\
This can be written into the following linear system.
\begin{align}
    \begin{bmatrix}
        G(\pi_{t+1})\\
        G(w_{t+1})\\
        H_{t+1}
    \end{bmatrix}
    \le 
    \begin{bmatrix}
        \alpha + (1-\alpha)\gamma &  \frac{2KQ_{\tau,max}(1-\alpha)}{\tau_w} & (1-\alpha)\gamma\\
        \frac{M(1-\beta)}{\tau} & \beta & 0 \\
        \frac{2KMQ_{\tau,max}(1-\beta)}{\tau\tau_w} & \frac{2KQ_{\tau,max}(1-\beta)}{\tau_w} & \alpha
    \end{bmatrix}
    \begin{bmatrix}
        G(\pi_{t})\\
        G(w_{t})\\
        H_{t}
    \end{bmatrix}\label{eq:linear_system_exact}
\end{align}
where $M = \frac{r_{max}(1+\gamma) + 2\tau(1-\gamma)\log|A|}{(1-\gamma)^2}, \alpha=1-\frac{\tau\eta}{1-\gamma}, \beta = \frac{1}{\lambda\tau_w+1}$.

Let the transition matrix above be $A(\eta,\lambda)$.
Since $A(\eta,\lambda)$ is non-negative matrix, by Perron-Frobenius Theorem \cite{horn2012matrix}, $A(\eta,\lambda)$ has an eigenvalue which is equal to spectral radius.
The characteristic polynomial is simplified as follows.
\begin{align}
    f(x) = {} & (x - \alpha)\{\underbrace{(x - (\alpha + (1-\alpha)\gamma))(x - \beta)}_{=:f_1(x)}\\
    & - \underbrace{\frac{2KMQ_{\tau,max}}{\tau\tau_w}}_{=:X}(1-\alpha)(1-\beta) \}\\
    & - \underbrace{\gamma \frac{2KMQ_{\tau,max}}{\tau\tau_w}(1-\alpha)(1-\beta) ( x+1-2\beta )}_{=:g(x)}& 
\end{align}
Take step size $\eta$ and $\lambda$ to satisfy $\alpha = 1-\epsilon$ and $\beta = 1-\epsilon^2$. I.e., take $\eta = \frac{\epsilon(1-\gamma)}{\tau}$ and $\lambda = \frac{\epsilon^2}{\tau_w(1-\epsilon^2)}$.\\
Note that the polynomial $f_1$ has zeros at $\alpha + (1-\alpha)\gamma, \beta<1$.
Let $f_2(x) = f_1(x)-X(1-\alpha)(1-\beta)$, then $f_2(1) = (1-\gamma-X)(1-\alpha)(1-\beta)$. 
\begin{align}
    X =  \frac{2KMQ_{\tau,max}}{\tau\tau_w} < & \frac{2K}{\tau\tau_w} \frac{2(r_{max} + \tau\log|A|)}{(1-\gamma)^2} Q_{\tau,max}\\
    = & \frac{4KQ_{\tau,max}^2}{\tau\tau_w(1-\gamma)}\\
    \le & \frac{1-\gamma}{3} \label{eq:assumption_X}
\end{align}
by assumption on $\tau_w$.
Therefore, $f_2(1)>0$ which implies that the zeros of $f_2$, namely $x_1\le x_2$ are less than 1. This implies that the first term of $f$, i.e. $(x-\alpha)f_2(x)$, has three distinct zeros at $\alpha=1-\epsilon < \alpha + (1-\alpha)\gamma = 1-\epsilon(1-\gamma) < \beta=1-\epsilon^2$ for $\epsilon<1-\gamma$.
Furthermore, $(x-\alpha)(f_1(x)-X(1-\alpha)(1-\beta))|_{x=1} = \epsilon f_2(1) = \epsilon^4 (1-\gamma-X)$ where $X$ is independent of $\epsilon$.
For the last term, $g(1)$ has value $\gamma \frac{4KMQ_{\tau,max}}{\tau\tau_w}\epsilon^5<\frac{2}{3}\gamma(1-\gamma)\epsilon^5$. For $\epsilon < \frac{1}{\gamma}$, $f(1) = \epsilon^4 (1-\gamma-X) - g(1)> \epsilon^4 (1-\gamma-\frac{1-\gamma}{3}) - \frac{2}{3}\gamma(1-\gamma)\epsilon^5
> \epsilon^4 (\frac{2(1-\gamma)}{3} - \frac{2}{3}\gamma(1-\gamma)\epsilon)>0$. Note that since $\gamma,\epsilon<1$, $\epsilon$ already satisfies $\epsilon < \frac{1}{\gamma}$. Therefore, $x_2$, the maximal zero of $(x-\alpha)f_2(x)$ moves to $x_2'$, the maximal zero of $f(x)=(x-\alpha)f_2(x)-g(x)$ which is still $x_2'<1$(Otherwise, $f(x)\le 0$).
Denote this $x_2'<1$ as $\rho(\eta,\lambda)$, which is the spectral radius of $A(\eta,\lambda)$.

Specifically, for $x=1-\frac{\epsilon^2}{2}$, direct calculation results in 
\begin{align}
    f(x)=&(\frac{1-\gamma}{2}-X)\epsilon^4 - (\frac{2-\gamma}{4}+\frac{3\gamma-1}{2}X)\epsilon^5 + \frac{1}{8}\epsilon^6\\
    >& (\frac{1-\gamma}{6}- (\frac{2-\gamma}{4}+\frac{3\gamma-1}{2}\frac{1-\gamma}{3})\epsilon)\epsilon^4~(\because X<\frac{1-\gamma}{3})\\
    =&(2(1-\gamma)-(4+5\gamma-6\gamma^2)\epsilon)\frac{\epsilon^4}{12} >0 
\end{align}
by assumption $\epsilon<\frac{2(1-\gamma)}{4+5\gamma-6\gamma^2}$. Therefore, $\rho(\eta,\lambda)<1-\frac{\epsilon^2}{2}<1$.

For simplicity, let $\mathbf{x}_t = \begin{bmatrix}
        G(\pi_{t})\\
        G(w_{t})\\
        H_{t}
    \end{bmatrix}$.
The linear system (\ref{eq:linear_system_exact}) leads to 
\begin{align}
    G(\pi_t),G(w_t),H_t\le \|\mathbf{x}_t\| \le  \rho(\eta,\lambda)^t\|\mathbf{x}_0\|
\end{align}
for some norm. Finally, by \eqref{eq:policy_bound_exact}, \eqref{eq:w_bound_exact_full}, \eqref{eq:q_gap_exact} and \eqref{eq:assumption_X}, conclude the following.
\begin{align}
    \|\log\pi^* - \log\pi_t\|_\infty 
    \le & \frac{2}{\tau}\|\mathbf{x}_0\|\rho(\eta,\lambda)^t\\
    \|w^* - w\|_\infty 
    \le & \frac{2}{\tau_w}\|\mathbf{x}_0\|\rho(\eta,\lambda)^t\\
    \|Q^{\pi^*}_{w^*,\tau} - Q^{\pi_{t}}_{w_{t},\tau}\|_\infty
    \le & (\frac{\tau(1-\gamma)}{3M}+2\gamma)\|\mathbf{x}_0\|\rho(\eta,\lambda)^t
\end{align}
where $\rho(\eta,\lambda)<1-\frac{\epsilon^2}{2}$ with choice of $\eta = \frac{\epsilon(1-\gamma)}{\tau}$ and $\lambda = \frac{\epsilon^2}{\tau_w(1-\epsilon^2)}$.

\begin{remark}\label{rmk:epsilon}
    The condition for $\epsilon$ is $\epsilon<\epsilon_0 := \min\{\frac{2(1-\gamma)}{4+5\gamma-6\gamma^2}, 1-\gamma\}$ can be simplified with tighter bound.
    Since $3\le 4+5\gamma-6\gamma^2\le\frac{121}{24}$ for $\gamma\in(0,1)$, $\frac{2(1-\gamma)}{4+5\gamma-6\gamma^2}\le \frac{2(1-\gamma)}{3}< 1-\gamma$ and thus, 
    $\epsilon_0 := \min\{\frac{2(1-\gamma)}{4+5\gamma-6\gamma^2}, 1-\gamma\}=\frac{2(1-\gamma)}{4+5\gamma-6\gamma^2}$.
    In particular, we can take stronger condition $\epsilon<\epsilon_0 := \frac{48(1-\gamma)}{121}$.
    Condition for $\lambda$ becomes $\lambda<\frac{\epsilon_0^2}{\tau_w}$, then, from $\lambda = \frac{\epsilon^2}{\tau_w(1-\epsilon^2)}>\frac{\epsilon^2}{\tau_w}$, $\lambda$ satisfies the condition for $\epsilon$. 
    Condition for $\eta$ becomes $\eta<\frac{\epsilon_0(1-\gamma)}{\tau}$, which also satisfies the condition for $\epsilon$ by $\epsilon = \frac{\eta\tau}{1-\gamma}<\epsilon_0$.
\end{remark}



\section{Proof of Theorem \ref{thm:convergence_inexact}}
\label{convergence_pf_inexact}
Now suppose that we can access only an approximate policy evaluation (i.e. estimate for soft Q-function). Let an approximate policy evaluation for given $\pi$ with reward $\langle w,\mathbf{r}\rangle$ be $\widehat{Q}^\pi_{w,\tau}$. 
We assume that $\|\widehat{Q}^\pi_{w,\tau}-Q^\pi_{w,\tau}\|_\infty\le \delta,~~\forall \pi,w,s,a$ (Assumption \ref{assumption:estimate_error}). 

We conduct similar analysis in Appendix \ref{convergence_pf_exact} with approximate policy evaluation and derive a linear system.

Similar to the appendix \ref{convergence_pf_exact}, define auxiliary variables.
\begin{align}
    \widehat{\xi}_0(s,a) := &\|\exp(Q^{\pi^*}_{w^*,\tau}(s,\cdot)/\tau)\|_1\pi_0(a|s)\\
    \widehat{\xi}_{t+1}(s,a) := &(\widehat{\xi}_{t}(s,a))^\alpha e^{\frac{1-\alpha}{\tau}\widehat{Q}^{\pi_t}_{w_t,\tau}(s,a)} \label{eq:xi_update_inexact}\\ 
    \widehat{\kappa}_0(k)  = &\|\exp(-\mathbf{V}^{\pi^*}_\tau/\tau_w)\|_1 w_0(k)\\
    \widehat{\kappa}_{t+1}(k)  = &(\widehat{\kappa}_t(k))^\beta e^{-\frac{1-\beta}{\tau_w}\widehat{\mathbf{V}}^{\pi_t}_\tau} \label{eq:kappa_update_inexact}\\
    \pi_t,w_t: & \text{ policy and weight at $t$-th iteration}\\
    (\pi^*,w^*): &\text{ Nash equilibrium in }\mathcal{RG}\\
    Q^{\pi}_{w,\tau}, V^{\pi}_{w,\tau} :& \text{ exact policy evaluations}\\
    \widehat{Q}^{\pi}_{w,\tau}, \widehat{V}^{\pi}_{w,\tau}:& \text{ approximate policy evaluations}\\
    &\text{ I.e., estimators for soft optimal value for given policy $\pi$ and reward $\langle w,\mathbf{r}\rangle$.}\\
    \widehat{Q}^{\pi}_{\tau}, \widehat{V}^{\pi}_{\tau}: &\text{ approximate policy evaluations}\\
    &\text{ I.e., estimators for soft optimal value for given policy $\pi$ and vector reward $\mathbf{r}$}.\\
    Q^*_{w,\tau} :=& Q^{\pi^*(w)}_{w,\tau} \triangleq \max_\pi Q^{\pi}_{w,\tau}~(\text{soft optimal $Q$-value w.r.t. reward $\langle w,\mathbf{r}\rangle$})\\
    V^*_{w,\tau} := &V^{\pi^*(w)}_{w,\tau} \triangleq \max_\pi V^{\pi}_{w,\tau}~(\text{soft optimal $V$-value w.r.t. reward $\langle w,\mathbf{r}\rangle$})
\end{align}
Note that $\pi_t(\cdot|s)=\widehat{\xi}_t(s,\cdot)/\|\widehat{\xi}_t(s,\cdot)\|_1$ also holds for the case with approximate evaluation by the definition of $\widehat{\xi}_t$.

\textbf{Performance difference of NPG with approximate evaluation}\\
Before deriving the bound for the equation above, we first adapt performance difference lemma with approximate evaluation to our setting.

Performance difference for $V^{\pi_t}_{w_t,\tau}$ with approximate evaluation becomes
\begin{align}
    &V^{\pi_{t+1}}_{w_{t+1},\tau}(s) - V^{\pi_{t}}_{w_{t},\tau}(s) \label{eq:perf_diff_v_inexact}\\
    = & V^{\pi_{t+1}}_{w_{t+1},\tau}(s) -V^{\pi_{t+1}}_{w_{t},\tau}(s) + V^{\pi_{t+1}}_{w_{t},\tau}(s) - V^{\pi_{t}}_{w_{t},\tau}(s)\\
    \ge & \langle w_{t+1} - w_{t}, \mathbf{V}^{\pi_{t+1}}_{\tau}(s)\rangle - \frac{2}{1-\gamma}\|\widehat{Q}^{\pi_t}_{w_t,\tau}-Q^{\pi_t}_{w_t,\tau}\|_\infty~(\because\text{\cite{npg_ent}, lemma 4})\\
    \ge & -V_{\tau,max,l_1}\|w_{t+1}-w_t\|_\infty - \frac{2}{1-\gamma}\delta \\
    = & -Q_{\tau,max,l_1}\|w_{t+1}-w_t\|_\infty - \frac{2}{1-\gamma}\delta, 
\end{align}

and performance difference for $Q^{\pi_t}_{w_t,tau}$ with approximate evaluation becomes
\begin{align}
    &Q^{\pi_{t+1}}_{w_{t+1},\tau}(s,a) - Q^{\pi_{t}}_{w_{t},\tau}(s,a) \label{eq:perf_diff_q_inexact} \\
    =&\langle w_{t+1} - w_{t}, \mathbf{r}(s,a)\rangle + \gamma\mathbb{E}_{s'}[V^{\pi_{t+1}}_{w_{t+1},\tau}(s') - V^{\pi_{t}}_{w_{t},\tau}(s')]\\
    \ge& -Kr_{max}\|w_{t+1} - w_{t}\|_\infty -\gamma Q_{\tau,max,l_1}\|w_{t+1}-w_t\|_\infty - \frac{2\gamma\delta}{1-\gamma}\\
    = &-Q_{\tau,max,l_1}\|w_{t+1} - w_{t}\|_\infty- \frac{2\gamma\delta}{1-\gamma}. 
\end{align}

\subsection{Recursive bounds for with approximate evaluation}
\textbf{Recursive bounds with approximate evaluation: optimality gap for policy}
For each $(s,a)$,
\begin{align}
    & Q^{\pi^*}_{w^*,\tau}(s,a)-\tau\log\widehat{\xi}_{t+1}(s,a) \\
    =  & Q^{\pi^*}_{w^*,\tau}(s,a) - \tau\alpha\log\widehat{\xi}_{t}(s,a) - (1-\alpha)\widehat{Q}^{\pi_t}_{w_t,\tau}(s,a)~(\because (\ref{eq:xi_update_inexact}))\\
     = & \alpha(Q^{\pi^*}_{w^*,\tau}(s,a) - \tau\log\widehat{\xi}_{t}(s,a)) 
     + (1-\alpha)(Q^{\pi^*}_{w^*,\tau}(s,a) - \widehat{Q}^{\pi_t}_{w_t,\tau}(s,a))\\
     = & \alpha(Q^{\pi^*}_{w^*,\tau}(s,a) - \tau\log\widehat{\xi}_{t}(s,a))\\
    + &(1-\alpha)(Q^{\pi^*}_{w^*,\tau}(s,a) - Q^{\pi_t}_{w_t,\tau}(s,a))
    + (1-\alpha) (Q^{\pi_t}_{w_t,\tau}(s,a) - \widehat{Q}^{\pi_t}_{w_t,\tau}(s,a))
\end{align}
Thus, 
\begin{align}
    \| Q^{\pi^*}_{w^*,\tau}-\tau\log\widehat{\xi}_{t+1}\|_\infty & \le \alpha\|Q^{\pi^*}_{w^*,\tau} - \tau\log\widehat{\xi}_{t}\|_\infty + (1-\alpha)\|Q^{\pi^*}_{w^*,\tau} - Q^{\pi_t}_{w_t,\tau}\|_\infty + (1-\alpha)\delta.
    \label{eq:pi_gap_inexact}
\end{align}

\textbf{Recursive bounds with approximate evaluation: optimality gap for soft $Q$ function}
1) Upper bound\\
Similar the case with exact evaluation, start from the following identity.
\begin{align}
    &Q^{\pi^*}_{w^*,\tau}(s,a) - Q^{\pi_{t+1}}_{w_{t+1},\tau}(s,a)\\
    = & \langle w^*-w_{t+1}, \mathbf{r}(s,a)\rangle \label{eq:rwd_term_inexact}\\
    +& \gamma \mathbb{E}_{s'\sim P(\cdot|s,a)}[\tau\log\|e^{Q^{\pi^*}_{w^*,\tau}(s',\cdot)/\tau}\|_1 - \tau\log \|\widehat{\xi}_{t+1}(s',\cdot)\|_1] \label{eq:A_inexact}\\
    -& \gamma \mathbb{E}_{s'\sim P(\cdot|s,a),a'\sim\pi_{t+1}(\cdot|s')}[Q^{\pi_{t+1}}_{w_{t+1},\tau}(s',a') - \tau\log\widehat{\xi}_{t+1}(s',a')] \label{eq:B_inexact}\\
    \le & \frac{2Kr_{max}}{\tau_w}\|-\mathbf{V}^{\pi^*}_{\tau} - \tau_w\log \widehat{\kappa}_{t+1}\|_\infty\\
     + &\gamma\| Q^{\pi^*}_{w^*,\tau} - \tau\log\widehat{\xi}_{t+1}\|_\infty\\
    + &\gamma\max\{0,-\min_{s,a}\{Q^{\pi_{t+1}}_{w_{t+1},\tau}(s,a) - \tau\log\widehat{\xi}_{t+1}(s,a)\}\}
\end{align}
where the last inequality is derived by the similar logic in the case with exact evaluation.\\

2) Lower bound\\
Similar to the case with exact evaluation,
\begin{align}
    Q^{\pi^*}_{w^*,\tau}(s,a) - Q^{\pi_{t+1}}_{w_{t+1},\tau}(s,a)
    \ge - \frac{2Q_{\tau, max,l_1}}{\tau_w} \|-\mathbf{V}^{\pi^*}_{\tau} - \tau_w\log \widehat{\kappa}_{t+1}\|_\infty.
\end{align}

Combining the upper bound and the lower bound, obtain the following upper bound for the optimality gap.
\begin{align}
    & \|Q^{\pi^*}_{w^*,\tau} - Q^{\pi_{t+1}}_{w_{t+1},\tau}\|_\infty \label{eq:q_gap_inexact}\\
    \le & \frac{2KQ_{\tau,max}}{\tau_w}\|-\mathbf{V}^{\pi^*}_{\tau} - \tau_w\log \widehat{\kappa}_{t+1}\|_\infty + \gamma\| Q^{\pi^*}_{w^*,\tau} - \tau\log\widehat{\xi}_{t+1}\|_\infty \\
    +& \gamma\max\{0,-\min_{s,a}\{Q^{\pi_{t+1}}_{w_{t+1},\tau}(s,a) - \tau\log\widehat{\xi}_{t+1}(s,a)\}\}
\end{align}

\textbf{Recursive bounds with approximate evaluation: optimality gap for $w$}

For each $k=1,\ldots,K$,
\begin{align}
    & -V^{\pi^*}_{k,\tau} - \tau_w\log \widehat{\kappa}_{t+1}(k)\\
    = & \beta(-V^{\pi^*}_{k,\tau}) + (1-\beta)(-V^{\pi^*}_{k,\tau}) - \tau_w\beta\log\widehat{\kappa}_t(k) -(1-\beta)(-\widehat{V}^{\pi_t}_{k,\tau})~(\because (\ref{eq:kappa_update_inexact}))\\
    = & \beta(-V^{\pi^*}_{k,\tau} - \tau_w\log\widehat{\kappa}_t(k)) + (1-\beta)(-V^{\pi^*}_{k,\tau}+V^{\pi_t}_{k,\tau})
     +(1-\beta)(-V^{\pi_t}_{k,\tau} + \widehat{V}^{\pi_t}_{k,\tau}).
\end{align}

Then, for the same logic in the case with exact evaluation, obtain the following bound.
\begin{align}
    & \|-\mathbf{V}^{\pi^*}_{\tau} - \tau_w\log \widehat{\kappa}_{t+1}\|_\infty\\
    \le &\beta \|-\mathbf{V}^{\pi^*}_{\tau} - \tau_w\log \widehat{\kappa}_{t}\|_\infty + (1-\beta)\frac{M}{\tau}\|Q^{\pi^*}_{w^*,\tau}-\tau\log\widehat{\xi}_t\|_\infty
    + (1-\beta)\delta
    \label{eq:w_gap_inexact}
\end{align}

\textbf{Recursive bounds with approximate evaluation: supplementary term}
\begin{align}
    &Q^{\pi_{t+1}}_{w_{t+1},\tau}(s,a) - \tau\log\widehat{\xi}_{t+1}(s,a)\\
    = & Q^{\pi_{t+1}}_{w_{t+1},\tau}(s,a) - \tau\alpha\log\widehat{\xi}_{t}(s,a) - (1-\alpha)\widehat{Q}^{\pi_{t}}_{w_{t},\tau}(s,a)~(\because(\ref{eq:xi_update_inexact}))\\
    = & Q^{\pi_{t+1}}_{w_{t+1},\tau}(s,a) - \tau\alpha\log\widehat{\xi}_{t}(s,a) - (1-\alpha)(\widehat{Q}^{\pi_{t}}_{w_{t},\tau}(s,a)-Q^{\pi_{t}}_{w_{t},\tau}(s,a)+Q^{\pi_{t}}_{w_{t},\tau}(s,a))\\
    = & Q^{\pi_{t+1}}_{w_{t+1},\tau}(s,a) - Q^{\pi_{t}}_{w_{t},\tau}(s,a)\\
    +&\alpha(Q^{\pi_{t}}_{w_{t},\tau}(s,a) - \tau\alpha\log\widehat{\xi}_{t}(s,a))
    - (1-\alpha)(\widehat{Q}^{\pi_{t}}_{w_{t},\tau}(s,a)-Q^{\pi_{t}}_{w_{t},\tau}(s,a))
    \label{eq:supp_bd_intermediate_inexact}
\end{align}

Modify (\ref{eq:q_difference_start}) with approximate evaluation as follows.
\begin{align}
    &Q^{\pi_{t+1}}_{w_{t+1},\tau}(s,a) -  Q^{\pi_{t}}_{w_{t},\tau}(s,a)\\
    = & (Q^{\pi_{t+1}}_{w_{t+1},\tau}(s,a) - Q^{\pi_{t+1}}_{w_{t},\tau}(s,a))
    + (Q^{\pi_{t+1}}_{w_{t},\tau}(s,a)
    -  Q^{\pi_{t}}_{w_{t},\tau}(s,a))\\
    \ge & Q^{\pi_{t+1}}_{w_{t+1},\tau}(s,a) - Q^{\pi_{t+1}}_{w_{t},\tau}(s,a)
    {-Q_{\tau,max,l_1}\|w_{t+1} - w_{t}\|_\infty- \frac{2\gamma\delta}{1-\gamma}}~(\because(\ref{eq:perf_diff_q_inexact}))\\
    = & \langle w_{t+1}-w_t, Q^{\pi_{t+1}}_\tau \rangle
    {-Q_{\tau,max,l_1}\|w_{t+1} - w_{t}\|_\infty- \frac{2\gamma\delta}{1-\gamma}}\\
    \ge & - Q_{\tau,max,l_1}\|w_{t+1}-w_t\|_\infty
    {-Q_{\tau,max,l_1}\|w_{t+1} - w_{t}\|_\infty- \frac{2\gamma\delta}{1-\gamma}}
    ~(\because H\Ddot{o}lder \text{ inequality})\\
    \ge & - {2} Q_{\tau,max,l_1}\|\log w_{t+1}-\log w_t\|_\infty
    {- \frac{2\gamma\delta}{1-\gamma}}
    ~(\because \text{lemma \ref{lem:log_bd}})\\
    \ge & - {4}Q_{\tau,max,l_1}\|\log \widehat{\kappa}_{t+1}-\log \widehat{\kappa}_t\|_\infty
    {- \frac{2\gamma\delta}{1-\gamma}}
    ~(\because w_t = \frac{\widehat{\kappa}_t}{\|\widehat{\kappa}_t\|_1}~\forall t\text{ and 1-Lipschitzness of lse})\\
    = & -\frac{{4}Q_{\tau,max,l_1}(1-\beta)}{\tau_w}\|-\widehat{\mathbf{V}}^{\pi_t}_\tau - \tau_w \log \widehat{\kappa}_t \|_\infty 
    {- \frac{2\gamma\delta}{1-\gamma}}\\
    \ge & -\frac{{4}Q_{\tau,max,l_1}(1-\beta)}{\tau_w} (\|-\widehat{\mathbf{V}}^{\pi_t}_\tau + \mathbf{V}^{\pi_t}_\tau \|_\infty
    +
    \|-\mathbf{V}^{\pi_t}_\tau + \mathbf{V}^{\pi^*}_\tau \|_\infty
    +
    \|-\mathbf{V}^{\pi^*}_\tau - \tau_w \log \widehat{\kappa}_t \|_\infty)\\
    -& \frac{2\gamma\delta}{1-\gamma}\\
    \ge & -\frac{{4}Q_{\tau,max,l_1}(1-\beta)}{\tau_w} (\delta + \frac{M}{\tau}\|Q^{\pi^*}_{w^*,\tau} - \tau\log\widehat{\xi}_t \|_\infty
    +
    \|-\mathbf{V}^{\pi^*}_\tau - \tau_w \log \widehat{\kappa}_t \|_\infty)
    {- \frac{2\gamma\delta}{1-\gamma}}
    \label{eq:q_difference_inexact}
\end{align}

Plugging (\ref{eq:q_difference_inexact}) into (\ref{eq:supp_bd_intermediate_inexact}), obtain
\begin{align}
    &Q^{\pi_{t+1}}_{w_{t+1},\tau}(s,a) - \tau\log\widehat{\xi}_{t+1}(s,a)\\
    \ge & -\frac{{4}Q_{\tau,max,l_1}(1-\beta)}{\tau_w} (\delta + \frac{M}{\tau}\|Q^{\pi^*}_{w^*,\tau} - \tau\log\widehat{\xi}_t \|_\infty
    +
    \|-\mathbf{V}^{\pi^*}_\tau - \tau_w \log \widehat{\kappa}_t \|_\infty)
    {- \frac{2\gamma\delta}{1-\gamma}}\\
     +&\alpha(Q^{\pi_{t}}_{w_{t},\tau}(s,a) - \tau\alpha\log\widehat{\xi}_{t}(s,a))
    -(1-\alpha)\delta\\
    = & -\frac{{4}Q_{\tau,max,l_1}(1-\beta)}{\tau_w} (\frac{M}{\tau}\|Q^{\pi^*}_{w^*,\tau} - \tau\log\widehat{\xi}_t \|_\infty
    +
    \|-V^{\pi^*}_\tau - \tau_w \log \widehat{\kappa}_t \|_\infty)\\
     +& \alpha(Q^{\pi_{t}}_{w_{t},\tau}(s,a) - \tau\alpha\log\widehat{\xi}_{t}(s,a))\\
    -&((1-\alpha)(1+\frac{2\gamma}{\tau\eta})-\frac{{4}Q_{\tau,max,l_1}(1-\beta)}{\tau_w})\delta~ (\because \alpha = 1-\frac{\eta\tau}{1-\gamma})
\end{align}

By reversing the sign of both sides, taking $\max_{s,a}$ and $\max\{0,\cdot\}$ as in the case with exact evaluation, we obtain the following bound.
\begin{align}
    &\max\{0,-\min_{s,a}\{Q^{\pi_{t+1}}_{w_{t+1},\tau}(s,a) - \tau\log\widehat{\xi}_{t+1}(s,a)\}\}\\
    \le & \alpha\max\{0,-\min_{s,a}\{Q^{\pi_{t}}_{w_{t},\tau}(s,a) - \tau\log\widehat{\xi}_{t}(s,a)\}\}\\
     + &\frac{4KQ_{\tau,max}(1-\beta)}{\tau_w} (\frac{M}{\tau}\|Q^{\pi^*}_{w^*,\tau} - \tau\log\widehat{\xi}_t \|_\infty
    +
    \|-\mathbf{V}^{\pi^*}_\tau - \tau_w \log \widehat{\kappa}_t \|_\infty)\\
    +&\max\{0, ((1-\alpha)(1+\frac{2\gamma}{\tau\eta})-\frac{4KQ_{\tau,max}(1-\beta)}{\tau_w})\}\delta
    \label{eq:supp_inexact}
\end{align}

\subsection{Linear system with approximate evaluation}

For simplicity, we denote the three optimality gaps and supplementary term as follows, with slight abuse of notation.
\begin{align}
    G(\pi_{t}) := & \|Q^{\pi^*}_{w^*,\tau} - \tau\log\widehat{\xi}_t\|_\infty\\
    G(Q_{t}) := & \|Q^{\pi^*}_{w^*,\tau} - Q^{\pi_{t}}_{w_{t},\tau}\|_\infty \\
    G(w_{t}) := & \|-\mathbf{V}^{\pi^*}_{\tau} - \tau_w\log \widehat{\kappa}_{t}\|_\infty\\
    H_{t} := & \max\{0,-\min_{s,a} (Q^{\pi_t}_{w_t,\tau} - \tau\log\widehat{\xi}_t)\} 
\end{align}

$
\left\{
\begin{aligned}
    G(Q_{t+1}) \le {} & \frac{2KQ_{\tau,max}}{\tau_w}G(w_{t+1})
     + \gamma G(\pi_{t+1})
     + \gamma H_{t+1} & (\because\eqref{eq:q_gap_inexact})\\
    G(\pi_{t+1})  \le {} & \alpha G(\pi_{t}) + (1-\alpha) G(Q_{t}) + (1-\alpha)\delta & (\because\eqref{eq:pi_gap_inexact})\\
     \le {}& (\alpha + (1-\alpha)\gamma) G(\pi_{t}) + (1-\alpha) (\frac{2KQ_{\tau,max}}{\tau_w}G(w_{t})
     + \gamma H_{t}) + (1-\alpha)\delta\\
     {}&(\because\text{by the bound for $G(Q_{t+1})$ above})\\
    G(w_{t+1}) \le {}& \beta G(w_{t}) + (1-\beta)\frac{M}{\tau}G(\pi_{t}) + (1-\beta)\delta& (\because\eqref{eq:w_gap_inexact})\\
    H_{t+1} \le {}&  \alpha H_t + \frac{4KMQ_{\tau,max}(1-\beta)}{\tau\tau_w} G(\pi_t) + \frac{4KQ_{\tau,max}(1-\beta)}{\tau_w} G(w_t) & (\because\eqref{eq:supp_inexact})\\
    +&\max\{0, ((1-\alpha)(1+\frac{2\gamma}{\tau\eta})-\frac{4KQ_{\tau,max}(1-\beta)}{\tau_w})\}\delta & {} 
\end{aligned}
\right.
$\\
Therefore, the resulting linear system becomes the following.
\begin{align}
    \begin{bmatrix}
        G(\pi_{t+1})\\
        G(w_{t+1})\\
        H_{t+1}
    \end{bmatrix}
    \le 
    \begin{bmatrix}
        \alpha + (1-\alpha)\gamma &  \frac{2KQ_{\tau,max}(1-\alpha)}{\tau_w} & (1-\alpha)\gamma\\
        \frac{M(1-\beta)}{\tau} & \beta & 0 \\
        \frac{4KMQ_{\tau,max}(1-\beta)}{\tau\tau_w} & \frac{4KQ_{\tau,max}(1-\beta)}{\tau_w} & \alpha
    \end{bmatrix}
    \begin{bmatrix}
        G(\pi_{t})\\
        G(w_{t})\\
        H_{t}
    \end{bmatrix}
    + \delta y \label{eq:linear_system_inexact}
\end{align}
where
\begin{align}
    y = {}& \begin{bmatrix}
        1-\alpha\\
        1-\beta\\
        \max\{0, (1-\alpha)(1+\frac{2\gamma}{\tau\eta})-\frac{4KQ_{\tau,max}(1-\beta)}{\tau_w}\}
    \end{bmatrix},\\
    M = {}& \frac{r_{max}(1+\gamma) + 2\tau(1-\gamma)\log|A|}{(1-\gamma)^2}, \alpha=1-\frac{\tau\eta}{1-\gamma}, \text{ and } \beta = \frac{1}{\lambda\tau_w+1}.
\end{align}

Let the transition matrix above be $\widehat{A}(\eta,\lambda)$.
The characteristic polynomial is simplified as follows. 
\begin{align}
    \widehat{f}(x) = {} & (x - (\alpha + (1-\alpha)\gamma))(x - \beta)(x - \alpha)\\
    & - \frac{2KMQ_{\tau,max}(1-\alpha)(1-\beta)}{\tau\tau_w} (x - \alpha)\\
    & - \gamma \frac{4KMQ_{\tau,max}(1-\alpha)(1-\beta)}{\tau\tau_w} ( x+1-2\beta )
\end{align}
For the same analysis in Appendix \ref{convergence_pf_exact}, $\widehat{\rho}(\eta,\lambda)$, the spectral radius of $\widehat{A}(\eta,\lambda)$ satisfies $\widehat{\rho}(\eta,\lambda)<1-\frac{\epsilon^2}{2}<1$ with assumption 
$X = \frac{2KMQ_{\tau,max}}{\tau\tau_w}
    \le \frac{1-\gamma}{3}$ and $\eta = \frac{\epsilon(1-\gamma)}{\tau}$, $\lambda = \frac{\epsilon^2}{\tau_w(1-\epsilon^2)}$ with $\epsilon < \epsilon_0 := \min\{\frac{2(1-\gamma)}{4+11\gamma-12\gamma^2}, \frac{1}{2\gamma}, 1-\gamma\}$. 
More tightly, since $3\le 4+11\gamma-12\gamma^2 \le \frac{313}{48}$ for $\gamma\in(0,1)$, $\frac{2(1-\gamma)}{4+11\gamma-12\gamma^2} < 1-\gamma$ and we can take tighter condition 
$\epsilon \le \epsilon+0:=\min\{\frac{96(1-\gamma)}{313},\frac{1}{2\gamma}\}$.
To satisfy the condition for $\epsilon$, required condition for it suffice for $\eta$ and $\lambda$ to satisfy the range $\lambda<\frac{\epsilon_0^2}{\tau_w}$ and $\eta<\frac{\epsilon_0(1-\gamma)}{\tau}$.

Using $\widehat{\rho}(\eta,\lambda)<1-\frac{\epsilon^2}{2}$ and by letting $\mathbf{x}_t=\begin{bmatrix}
        G(\pi_{t})\\
        G(w_{t})\\
        H_{t}
    \end{bmatrix}$(slightly different from \ref{convergence_pf_exact} since this $\mathbf{x}_t$ contains different auxiliary variables), recursive application of the linear system (\ref{eq:linear_system_inexact}) results in the following.
\begin{align}
    \mathbf{x}_{t} \le \widehat{A}(\eta,\lambda)^t \mathbf{x}_{0} + (I+\cdots+\widehat{A}(\eta,\lambda)^{t-1})\delta y    
\end{align}
which results in
\begin{align}
    G(\pi_t),G(w_t),H_t\le \|\mathbf{x}_t\| 
    \le & \widehat{\rho}(\eta,\lambda)^t\|\mathbf{x}_0\| + \frac{1-\widehat{\rho}(\eta,\lambda)^t}{1-\widehat{\rho}(\eta,\lambda)}\|y\|\delta\\
    \le & \widehat{\rho}(\eta,\lambda)^t\|\mathbf{x}_0\| + \frac{1}{1-\widehat{\rho}(\eta,\lambda)}\|y\|\delta\\
    \le & \widehat{\rho}(\eta,\lambda)^t\|\mathbf{x}_0\| + \frac{2\|y\|}{\epsilon^2}\delta
\end{align}
for some norm. Therefore, 
\begin{align}
    \|\log\pi^* - \log\pi_t\|_\infty 
    \le & \frac{2}{\tau}( \|\mathbf{x}_0\|\widehat{\rho}(\eta,\lambda)^t + \frac{2\|y\|}{\epsilon^2}\delta)\\
    \|w^* - w\|_\infty 
    \le & \frac{2}{\tau_w}( \|\mathbf{x}_0\|\widehat{\rho}(\eta,\lambda)^t + \frac{2\|y\|}{\epsilon^2}\delta)\\
    \|Q^{\pi^*}_{w^*,\tau} - Q^{\pi_{t}}_{w_{t},\tau}\|_\infty
    \le & (\frac{\tau(1-\gamma)}{3M}+2\gamma)( \|\mathbf{x}_0\|\widehat{\rho}(\eta,\lambda)^t + \frac{2\|y\|}{\epsilon^2}\delta)
\end{align}
where $\widehat{\rho}(\eta,\lambda)<1-\frac{\epsilon^2}{2}$ with choice of $\eta = \frac{\epsilon(1-\gamma)}{\tau}$ and $\lambda = \frac{\epsilon^2}{\tau_w(1-\epsilon^2)}$.


\section{Proof of Corollaries \ref{cor:Cor42} and \ref{coro:sample_complexity}}\label{append:coro_pf}

\paragraph{Proof of Corollary \ref{cor:Cor42}}

For each $i=1,2,3$, to achieve $C_i [\rho(\eta,\lambda)]^t\le \epsilon_{acc}$, it suffices to find $t$ which satisfies
$C_i [\rho(\eta,\lambda)]^t < C_i [1-\frac{\epsilon^2}{2}]^t=\epsilon_{acc}$. 
Then, $\log C_i + t\log (1-\frac{\epsilon^2}{2}) = \log\epsilon_{acc}$. Since $\log (1-\frac{\epsilon^2}{2})\approx-\frac{\epsilon^2}{2}$ for small $\epsilon$, this results in 
$t=\frac{2}{\epsilon^2}(\log C_i - \log\epsilon_{acc})=O(\frac{1}{\epsilon^2}\log\frac{1}{\epsilon_{acc}})$.

\paragraph{Proof of Corollary \ref{coro:sample_complexity}}
For each $i=1,2,3$, to achieve $\widehat{C}_i [\widehat{\rho}(\eta,\lambda)]^t + \widehat{D}_i\delta/\epsilon^2\le 2\epsilon_{acc}$ with $\delta\le\frac{\epsilon^2\epsilon_{acc}}{\widehat{D}_i}$, it suffices to find $t$ which satisfies
$\widehat{C}_i [\widehat{\rho}(\eta,\lambda)]^t + \widehat{D}_i\delta/\epsilon^2 
< \widehat{C}_i [1-\frac{\epsilon^2}{2}]^t + \epsilon_{acc}
=2\epsilon_{acc}$, i.e., $\widehat{C}_i [1-\frac{\epsilon^2}{2}]^t
=\epsilon_{acc}$.
For the same logic in the case of exact policy evaluation, iteration complexity becomes at most $O(\frac{1}{\epsilon^2}\log\frac{1}{\epsilon_{acc}})$.

The derivation and analysis for sample complexity is in Appendix~\ref{append:sample_complexity_inexact}.

\section{Sample Complexity under Approximate Policy Evaluation}
\label{append:sample_complexity_inexact}

\begin{assumption}\label{assumption:generative_model}
    We assume access to a generative model that allows sampling next states from $P(\cdot \mid s,a)$ for any $(s,a)$, as in \citet{sample_complexity_model_based_generative_model}.
\end{assumption}

\begin{assumption}\label{assumption:estimate_error}
    For any given error bound $\delta$, policy $\pi$ and weight $w\in\Delta^K$, we have access to an estimate $\widehat{Q}^{\pi}_{w,\tau}$ and $\widehat{Q}_k^{\pi}$ of value functions ${Q}^{\pi}_{w,\tau}$ and ${Q}_k^{\pi}$, 
    which satisfy $\|\widehat{Q}^{\pi}_{w,\tau} - {Q}^{\pi}_{w,\tau}\|_\infty<\delta$ and $\|\widehat{Q}_k^{\pi} - {Q}_k^{\pi}\|_\infty<\delta$,
    respectively.
\end{assumption}

As discussed in  \citet{sample_complexity_model_based_generative_model}, Assumption~\ref{assumption:estimate_error} can be achieved 
for any fixed $\pi$ and any $k$-th objective
within error $\|\widehat{Q}^\pi_{k,\tau}-Q^\pi_{k,\tau}\|_\infty\le \delta$ with high probability if the number of samples per each state-actions pair exceeds the order of $\tilde{O}\left(\frac{1}{(1-\gamma)^3\delta^2}\right)$, under Assumption \ref{assumption:generative_model}. 
By employing fresh samples for the policy evaluation for each objective at every iteration and using union bound, 
$\|\widehat{Q}^\pi_{\tau}-Q^\pi_{\tau}\|_\infty=\max_k \|\widehat{Q}^\pi_{k,\tau}-Q^\pi_{k,\tau}\|_\infty\le \delta$ is achieved with high probability if the number of samples per each state-actions pair is exceeds the order of $\tilde{O}\left(\frac{K}{(1-\gamma)^3\delta^2}\right)$.
In particular, since $Q^\pi_{w,\tau}$ is a soft value function with a scalar reward $\langle w, \mathbf{r}\rangle$, we need samples exceeding the order of $\tilde{O}\left(\frac{1}{(1-\gamma)^3\delta^2}\right)$ to estimate.
Since $\tilde{O}\left(\frac{K}{(1-\gamma)^3\delta^2}\right)$($\because~w$ update requires vector value $V^{\pi_t}_\tau$) + $\tilde{O}\left(\frac{1}{(1-\gamma)^3\delta^2}\right)$($\because~\pi$ update requires scalar value $Q^{\pi_t}_{w_t,\tau}$) = $\tilde{O}\left(\frac{K}{(1-\gamma)^3\delta^2}\right)$, we used $\tilde{O}\left(\frac{K}{(1-\gamma)^3\delta^2}\right)$ in Corollary \ref{coro:sample_complexity}.

\section{Discussion on Two-player Zero-sum Markov Games}\label{append:2p0sMG}

We provide a detailed discussion on how our framework relates to the literature on two-player zero-sum Markov games (2p0s MGs).
\paragraph{Problem Settings.}
Our target problem, entropy-regularized max--min MORL, can be viewed as a two-player zero-sum Markov game in which 
the minimization player (the adversary) has a state-independent policy $w$, and the transition is determined solely by the maximization player (the learner), since $P(s'|s,a)$ is independent of $w$. 
We also aim to find an $\epsilon_{\mathrm{acc}}$-QRE, as considered in several prior works.
\paragraph{Algorithmic Settings.}
Table~\ref{tab:2p0s_comparison} summarizes the algorithmic settings of our method and existing studies on two-player zero-sum Markov games. 
Here, ``symm'' denotes symmetric regularization and learning-rate structures, while ``asym'' denotes asymmetric ones. 
In 2p0s MGs, symmetric regularization is natural because both players use policies as their strategies with cumulative entropy $\tilde{H}(\pi)$ on the order of $\frac{\log|\mathcal{A}|}{1-\gamma}$. 
In contrast, in ERAM, the adversary uses a weight $w$ as its strategy with non-cumulative entropy $\tilde{H}(w)$ on the order of $\log K$, 
which justifies the use of asymmetric regularization coefficients.
\begin{table}[h]
\centering
\caption{Comparison of algorithmic settings in two-player zero-sum Markov games.}
\label{tab:2p0s_comparison}
\small
\begin{tabular}{lccccc}
\toprule
{Reference} & {Entropy Reg.} & {Reg. Coef.} & {Base Method} & {Policy Param.} & {Learning Rates} \\
\midrule
\citet{daskalakis2020} & $\times$ & none  & PG     & direct   & asym  \\
\citet{2p0smg}       & $\checkmark$ & symm & PG     & softmax  & asym  \\
\citet{wei2021}        & $\checkmark$ & none & OGDA   & direct   & symm  \\
\citet{cen2023}       & $\checkmark$ & symm & OMWU   & direct   & symm  \\
\textbf{ERAM (Ours)}     & $\checkmark$ & asym & NPG, MD & softmax  & asym  \\
\bottomrule
\end{tabular}
\end{table}
\paragraph{Techniques for Proof.}
We carefully reviewed the proof of Theorem~1 in \citet{cen2023} and found it to be applicable to the MORL setting under symmetric learning rates and regularization ($\eta = \lambda$, $\tau = \tau_w$) by using 
$\frac{1}{1-\gamma}\tilde{H}(w)$ or by absorbing $\frac{1}{1-\gamma}$ into $\tau_w$, thereby yielding asymmetric coefficients. 
Nevertheless, a direct application remains challenging because our setting lacks these symmetric conditions. 
Our proof instead relies on asymmetric learning rates for last-iterate convergence (see Section~\ref{section_tabular}) and asymmetric regularization (see Algorithmic Settings above).

The result of Cen et al.~(2023) applies to MORL in the symmetric case because its key step, Eq.~(14), has a state-dependent right-hand side (RHS) that is ultimately bounded by $\| \cdot \|_{\Gamma_{(p)}}$, removing all state dependence. 
Hence, both the state-dependent policy in MGs and the state-independent $w$ in MORL satisfy the same recursive bounds as in Lemma~2--4 of Cen et al.~(2023). 
In our setting, however, since $w$ is already state-independent and there is only one RL learner, we find that adapting the proof in Cen et al.~(2022) is sufficient for our analysis.
\paragraph{Convergence Speed.}
Cen et al.~(2023) provides a convergence error bound of 
$O((1-(1-\gamma)\eta\gamma/4)^t)$,
while our method achieves at most 
$O((1-\epsilon^2/2)^t) \le \mathcal{O}((1-\lambda\tau_w/4)^t)$.
The proof in Appendix~\ref{convergence_pf_exact} shows that if we take $\epsilon \ge (1-\gamma)^2$, we have 
$1-\lambda\tau_w/4 \le 1-(1-\gamma)\eta\gamma/4$.
Hence, our convergence rate is comparable to that of prior work and does not conflict with their results.
(Note that $\epsilon \ge (1-\gamma)^2$ does not violate $\epsilon \in (0, \epsilon_0 = 48(1-\gamma)/121)$ (Remark~\ref{rmk:epsilon}), for $\gamma$ sufficiently close to~1.)
This discussion highlights that while the mathematical foundation of ERAM aligns closely with existing analyses in two-player zero-sum Markov games, 
our asymmetric learning-rate structure and the non-monotone nature of value gradients in RL motivate a distinct line of analysis tailored to the MORL framework.

\section{Supplementary Material for Numerical Results in Tabular MOMDPs}\label{append:tabular}
For each tabular MOMDP, transition matrix and reward function is randomly generated. The reward function has range $[1,20]$. 
For all experiment, we used $\gamma=0.95$, $\tau=\tau_w=0.05$, $\eta=0.01$ and $\lambda=0.0001$. The following graphs show last-iterate convergence behaviors for value $V^{\pi_t}_{w_t,\tau}$ and weight $w_t$.
For simplicity, we report $w_t(1)$, the first element of the weight $w_t$ at iteration $t$. 

The figures below show the last-iterate convergence behavior of values ($V^{\pi_t}_{w_t,\tau}$) and the first component of weights ($w_t(1)$) resulting from Algorithm~\ref{alg:pseudo_code_inexact}.

\begin{figure}[H]
    \centering
    \begin{minipage}{0.32\textwidth}
        \centering
        \includegraphics[width=\linewidth]{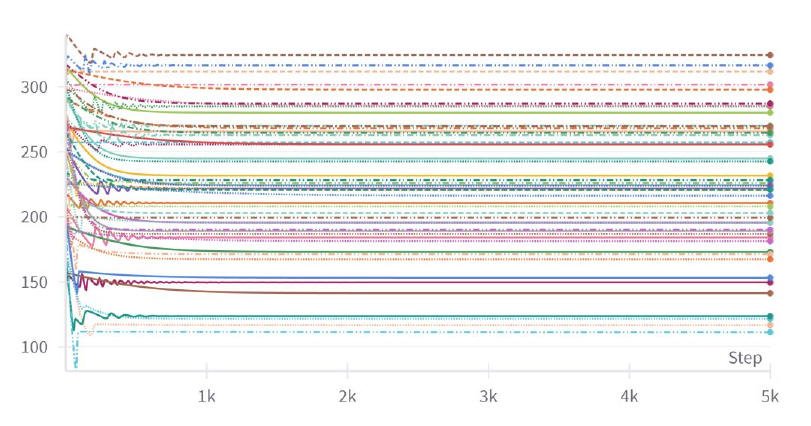}
        \caption*{$(|S|,|A|,K)=(2,2,2)$}
    \end{minipage}
    \hfill
    \begin{minipage}{0.32\textwidth}
        \centering
        \includegraphics[width=\linewidth]{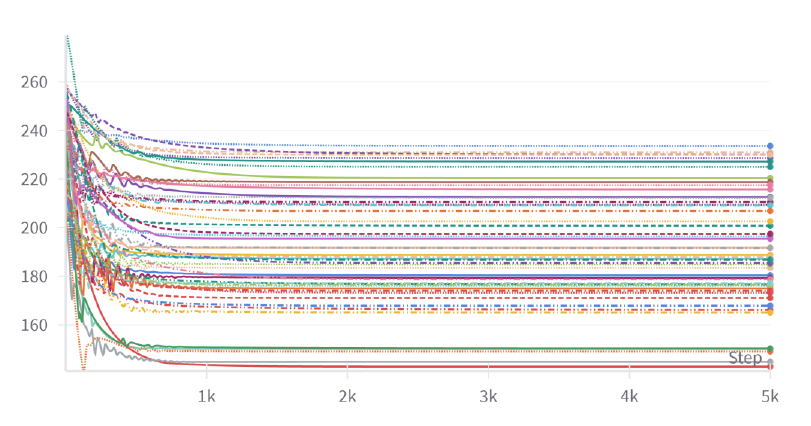}
        \caption*{$(|S|,|A|,K)=(3,3,6)$}
    \end{minipage}
    \hfill
    \begin{minipage}{0.32\textwidth}
        \centering
        \includegraphics[width=\linewidth]{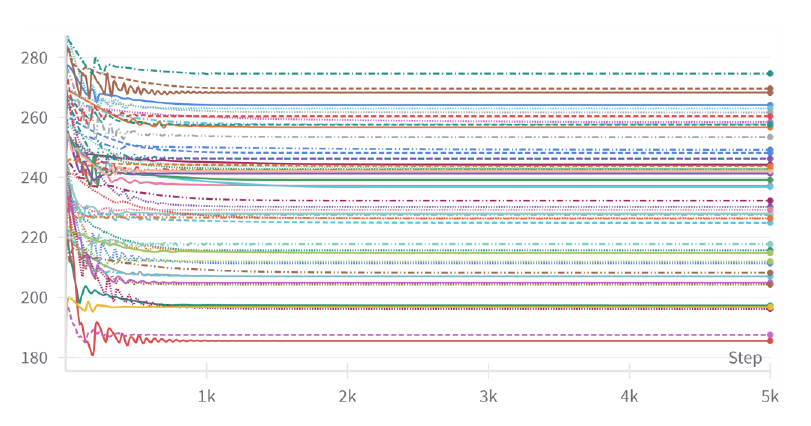}
        \caption*{$(|S|,|A|,K)=(4,4,4)$}
    \end{minipage}
    \caption{Last-iterate convergence of values in three types of MOMDPs, each with 50 randomly generated instances.}
\end{figure}

\begin{figure}[H]
    \centering
    \begin{minipage}{0.32\textwidth}
        \centering
        \includegraphics[width=\linewidth]{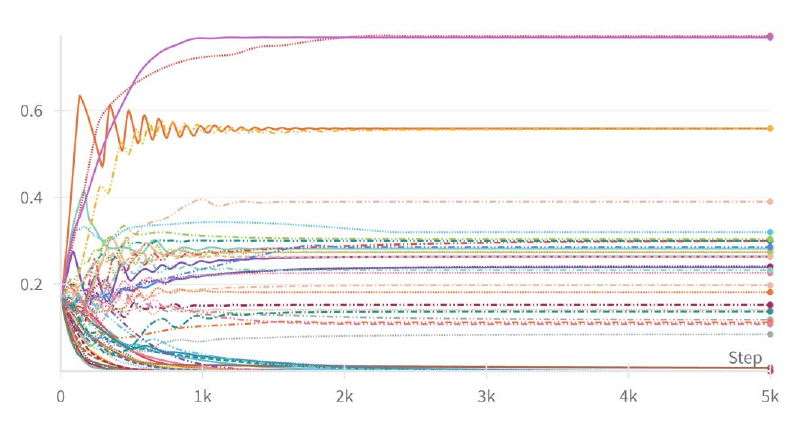}
        \caption*{$(|S|,|A|,K)=(2,2,2)$}
    \end{minipage}
    \hfill
    \begin{minipage}{0.32\textwidth}
        \centering
        \includegraphics[width=\linewidth]{images/w_336_r20_g95.pdf}
        \caption*{$(|S|,|A|,K)=(3,3,6)$}
    \end{minipage}
    \hfill
    \begin{minipage}{0.32\textwidth}
        \centering
        \includegraphics[width=\linewidth]{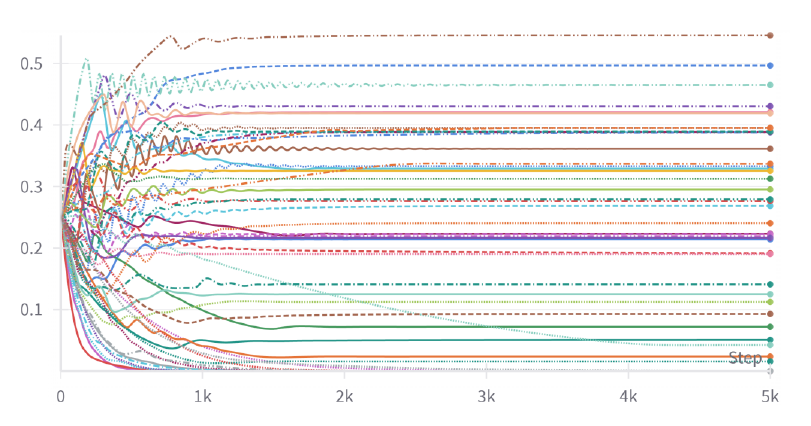}
        \caption*{$(|S|,|A|,K)=(4,4,4)$}
    \end{minipage}
    \caption{Last-iterate convergence of $w(1)$ in three types of MOMDPs, each with 50 randomly generated instances.}
\end{figure}

\begin{wrapfigure}[9]{r}{0.4\textwidth}
\vspace{-15pt}
  \centering
  \includegraphics[width=0.9\linewidth]{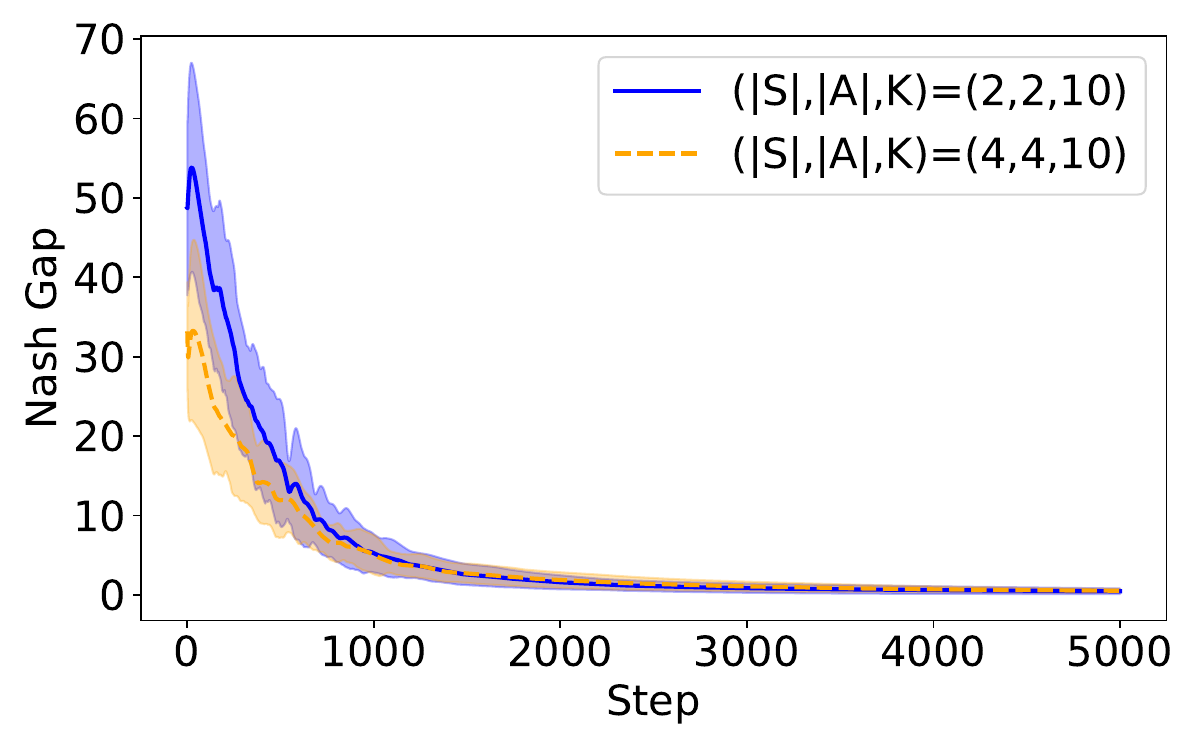}
  \captionsetup{aboveskip=2pt, belowskip=0pt}
  \caption{Nash gap for ARAM}
  \label{fig:aram_ng}
\end{wrapfigure}

We also evaluated ARAM in tabular settings. To demonstrate the effectiveness of ARAM, we considered two types of tabular MOMDPs with an increased number of objectives ($K=10$): $(|S|,|A|,K) = (2,2,10)$ and $(4,4,10)$. 
We used $\gamma=0.95$, $\tau=\tau_w=0.05$, $\eta=0.01$ and $\lambda=0.0001$.
Similar to ERAM, the Nash gap of ARAM decreases rapidly over time in both environments, as shown in Figure~\ref{fig:aram_ng}.
Each curve represents the average over 50 randomly generated instances, with shaded areas showing standard deviation.

\vspace{3em}

\section{More on the Traffic Signal Control Experiment}\label{append:traffic}

\subsection{Traffic signal control environment}\label{append:tsc_env}

We evaluate our method in a traffic signal control simulation environment with three scenarios: Base-4, Asym-4, and Asym-16. Each scenario simulates a four-way intersection with four roads (North, East, South, and West), where each road consists of four lanes. Vehicles arrive from the four directions with specified inflow proportions and can proceed straight, turn left, or turn right.

In the Base-4 scenario, 75\% of arriving vehicles proceed straight, and the remaining 25\% make left or right turns with equal probability. Among the straight-going vehicles, the proportions from West to East, East to West, North to South, and South to North are $[0.1, 0.1, 0.4, 0.4]$. The reward is 4-dimensional, where each element represents the negative total waiting time on a road. The simulation includes 10,000 vehicles and is trained for 100,000 time steps.

The Asym-4 scenario shares the same reward structure as Base-4 but introduces asymmetry in the traffic inflow, which is set to $[0.4, 0.1, 0.4, 0.1]$ across the four directions. Additionally, the turning probabilities vary by incoming direction to better reflect realistic traffic patterns. The scenario uses 4,000 vehicles and is also trained for 100,000 time steps.

The Asym-16 scenario uses the same asymmetric inflow as Asym-4 but increases the granularity of the reward by assigning one reward per lane, resulting in a 16-dimensional reward vector. Each element corresponds to the negative waiting time of a specific lane. This scenario includes 4,000 vehicles and is trained for 200,000 time steps.


For completeness, we provide the max-min performance under the metric of \citet{park} in the following table (see Table~\ref{tab:metric_park}).
Let $\pi^*$ denote the max-min optimal policy, and let $\pi^{seed_i}$ denote the final policy obtained from training with a fixed seed $seed_i$.
The metric used in the prior work \cite{park} averages over all seeds and then takes the minimum, i.e.,
$\min_k\frac{1}{n}\sum_{i=1}^n\mathbb{E}_{\pi^{seed_i}}[\sum_t \gamma^t r_k(s_t,a_t)].$
In contrast, our metric computes the max-min performance for each seed and then averages the results, i.e.,
$\frac{1}{n}\sum_{i=1}^n\min_k \mathbb{E}_{\pi^{seed_i}}[\sum_t \gamma^t r_k(s_t,a_t)].$
If $\pi^{seed_1}=\cdots=\pi^{seed_n}=\pi^*$, the value vectors $\mathbf{V}^{\pi^{seed_i}}$ are identical to $\mathbf{V}^{\pi^{*}}$, and both metrics yield the same measurement.
However, if $\pi^{seed_1},\cdots,\pi^{seed_n}$ are not exactly the same, the metric in the prior work averages returns that are from different policies, which does not accurately reflect each learned policy’s true performance. Thus, we evaluate the max-min performance of each policy separately and then average the results.

\begin{table}[!h]
\vspace{-5pt}
\centering
\begin{tabular}{cccccccc}
\toprule
Environments & ARAM & ERAM & \citet{park} & GGF-PPO & GGF-DQN & Avg-DQN  \\ 
\midrule
Base-4  & \textbf{-1140}  & \underline{-1387} & -1455    & -1603    & -1838      & -2774     \\ 
Asym-4  & \underline{-2589}  & \textbf{-2568} & -3510    & -3094    & -2670      & -4245     \\ 
Asym-16  & \textbf{-14399} & \underline{-15259} & -17754   & -19569   & -16477     & -27499   \\ 
\bottomrule
\end{tabular}
\captionsetup{aboveskip=2pt, belowskip=0pt}
\caption{Max-min performance in traffic signal control. Bold: best; underline: second-best.}
\label{tab:metric_park}
\vspace{-10pt} 
\end{table}

\subsection{Experimental Setup}\label{append:hyperparam}

The PPO hyperparameters are listed in the table below. We use the default network architecture and optimizer settings provided by Stable-Baselines3~\cite{stable-baselines3}. For entries with multiple values, the best-performing one was selected based on validation performance.
\begin{table}[H]
    \centering
    \begin{tabular}{cc}
    \toprule
    PPO hyperparameter & value \\
    \midrule
        entropy coefficient & 1e-6 \\
        value loss coefficient & 0.5 \\
        gae coefficient & 0.95 \\
        clip range & 0.2 \\
        optimizer & Adam \\
        hidden layer sizes for actor network & [64, 64] \\
        hidden layer sizes for critic network & [64, 64] \\
        activation function & Tanh \\
        epochs & 2, 4, 6, 8 \\
        rollout steps & 64, 128 \\
        batch size & 16, 32 \\
        learning rate & 0.001, 0.002, 0.003 \\
    \bottomrule
    \end{tabular}
    \caption{PPO hyperparameters for ERAM and ARAM}
    \label{tab:hyperparam}
\end{table}

The main hyperparameters for ERAM and ARAM are $\lambda$ and $\beta = \frac{1}{\lambda\tau_w + 1}$ (i.e. $\tau_w$), which appear in the closed-form updates of $w$ in both ERAM~\eqref{eq:w_closedform} and ARAM~\eqref{eq:appendARAMwt}. The best-performing hyperparameters were selected for each traffic scenario from the search over $\beta \in \{0.01, 0.25, 0.33, 0.5, 0.67, 0.75, 0.99\}$ and $\lambda \in \{0.001, 0.002, 0.003, 0.01, 0.02, 0.03, 0.1, 0.2, 0.3\}$, and their influence is analyzed in the ablation study.
In addition, all experiments were conducted on a machine equipped with two Intel Xeon Gold 6238R CPUs.

In each traffic scenario, the selected hyperparameters are listed in the following order: \\
(epochs, rollout steps, batch size, learning rate, $\lambda$, $\beta$).
\begin{table}[H]
    \centering
    \begin{tabular}{ccc}
    \toprule
    {} & ERAM & ARAM \\
    \midrule
        Base-4 &  (8, 128, 32, 0.001, 0.2, 0.67)& (8, 128, 32, 0.001, 0.1, 0.67)\\
        Asym-4 &  (4, 128, 32, 0.003, 0.03, 0.67)& (4, 128, 32, 0.003, 0.03, 0.25)\\
        Asym-16 &  (8, 128, 32, 0.001, 0.2, 0.5)& (8, 128, 32, 0.001, 0.2, 0.01)\\
    \bottomrule
    \end{tabular}
    \caption{Selected hyperparameters for ERAM and ARAM in each traffic scenario}
    \label{tab:hyperparam_opt}
\end{table}

\subsection{Ablation study}\label{append:ablation}

\begin{figure}[htbp!]
  \centering

  \begin{subfigure}[t]{0.45\textwidth}
    \includegraphics[width=\textwidth]{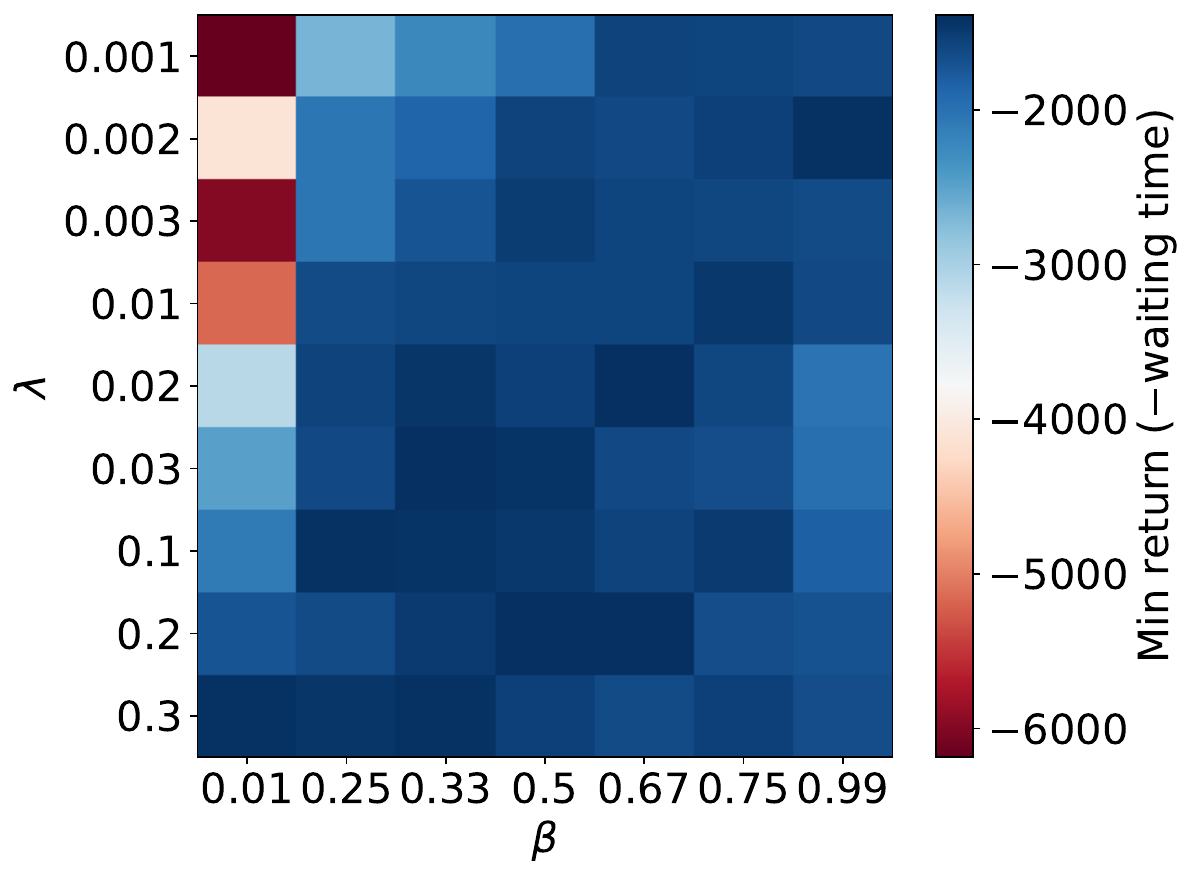}
    \caption*{ERAM in Base-4}
  \end{subfigure}
  \hspace{1em}
  \begin{subfigure}[t]{0.45\textwidth}
    \includegraphics[width=\textwidth]{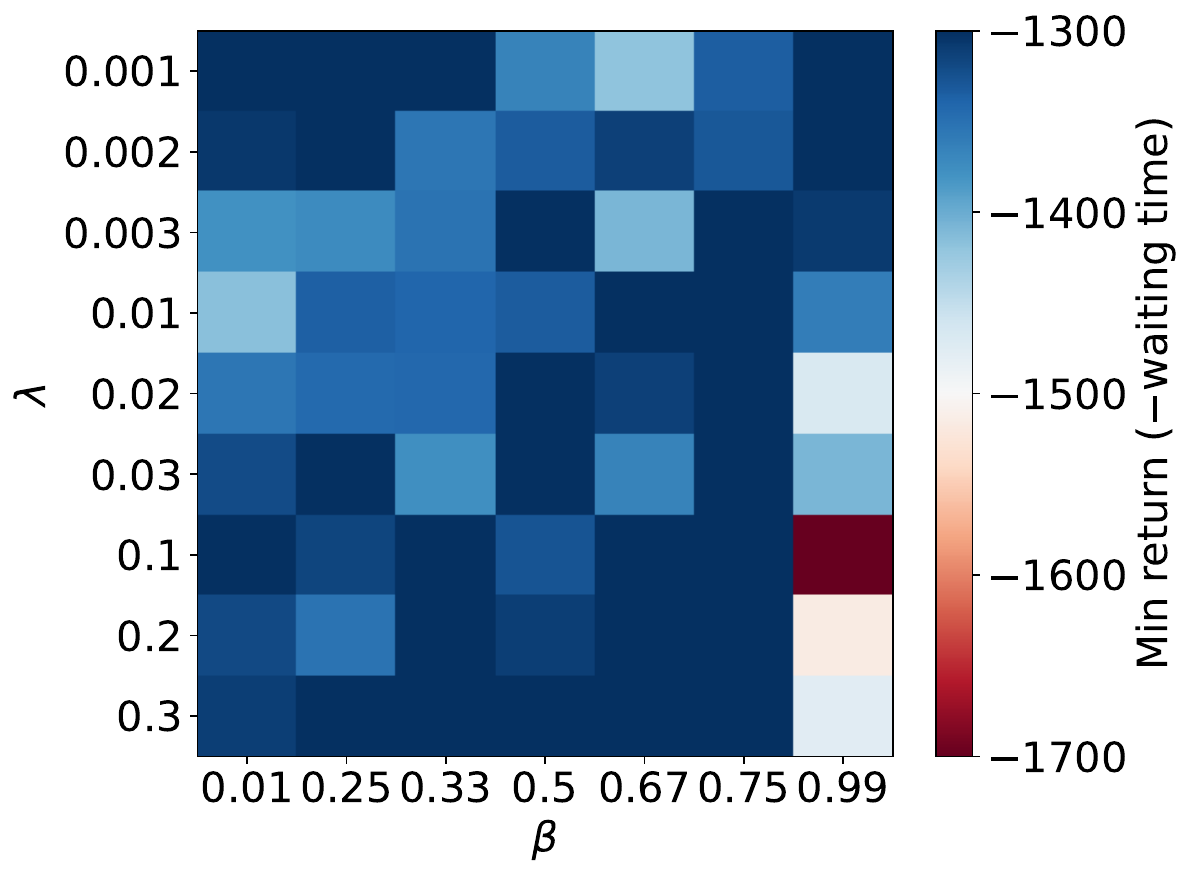}
    \caption*{ARAM in Base-4}
  \end{subfigure}

  \vspace{1em}

  \begin{subfigure}[t]{0.45\textwidth}
    \includegraphics[width=\textwidth]{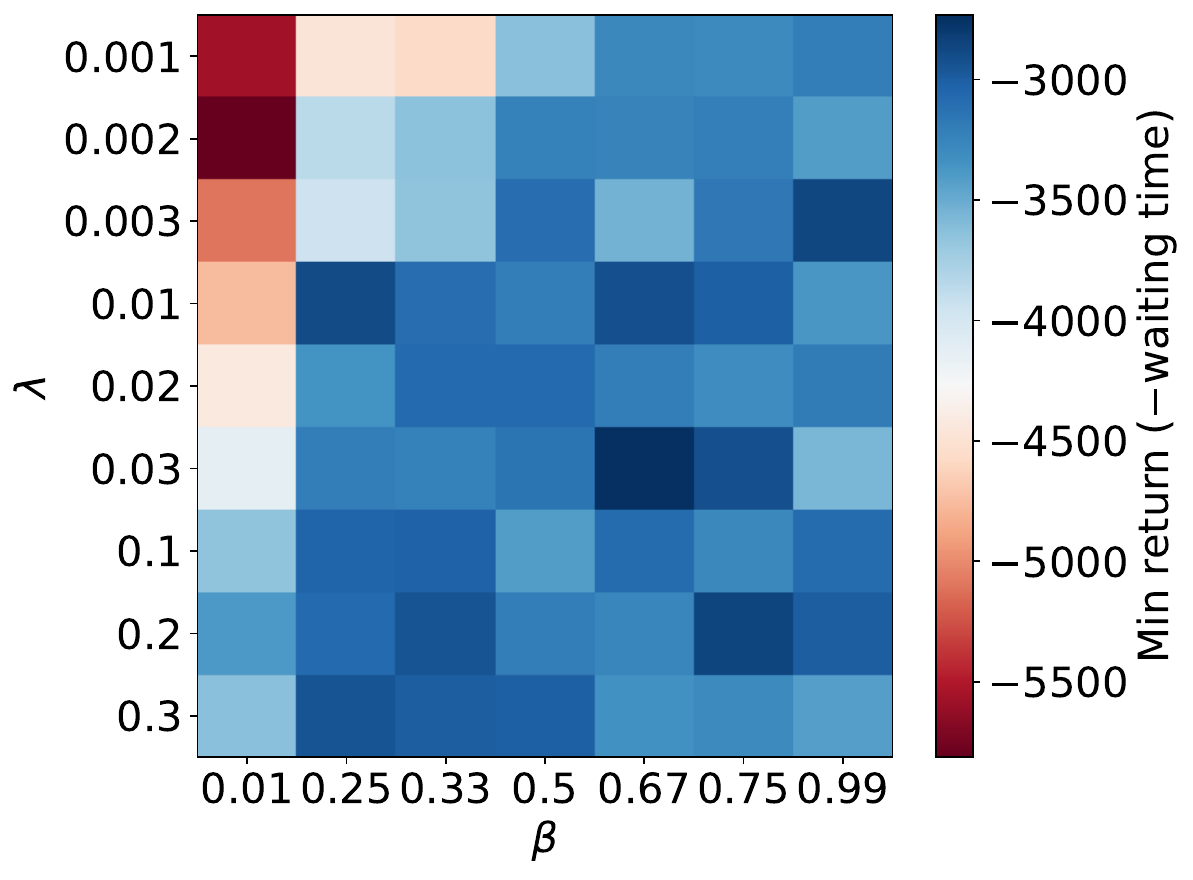}
    \caption*{ERAM in Asym-4}
  \end{subfigure}
  \hspace{1em}
  \begin{subfigure}[t]{0.45\textwidth}
    \includegraphics[width=\textwidth]{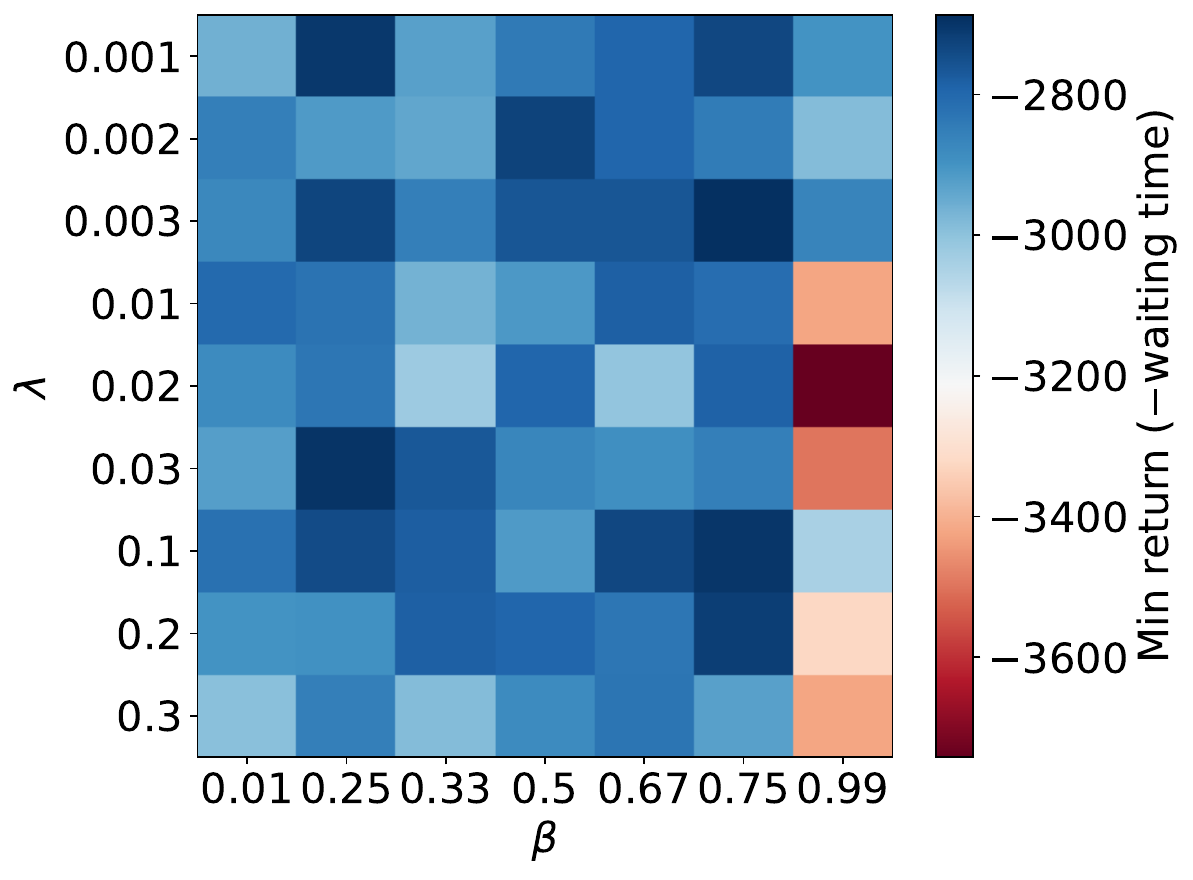}
    \caption*{ARAM in Asym-4}
  \end{subfigure}

  \vspace{1em}

  \begin{subfigure}[t]{0.45\textwidth}
    \includegraphics[width=\textwidth]{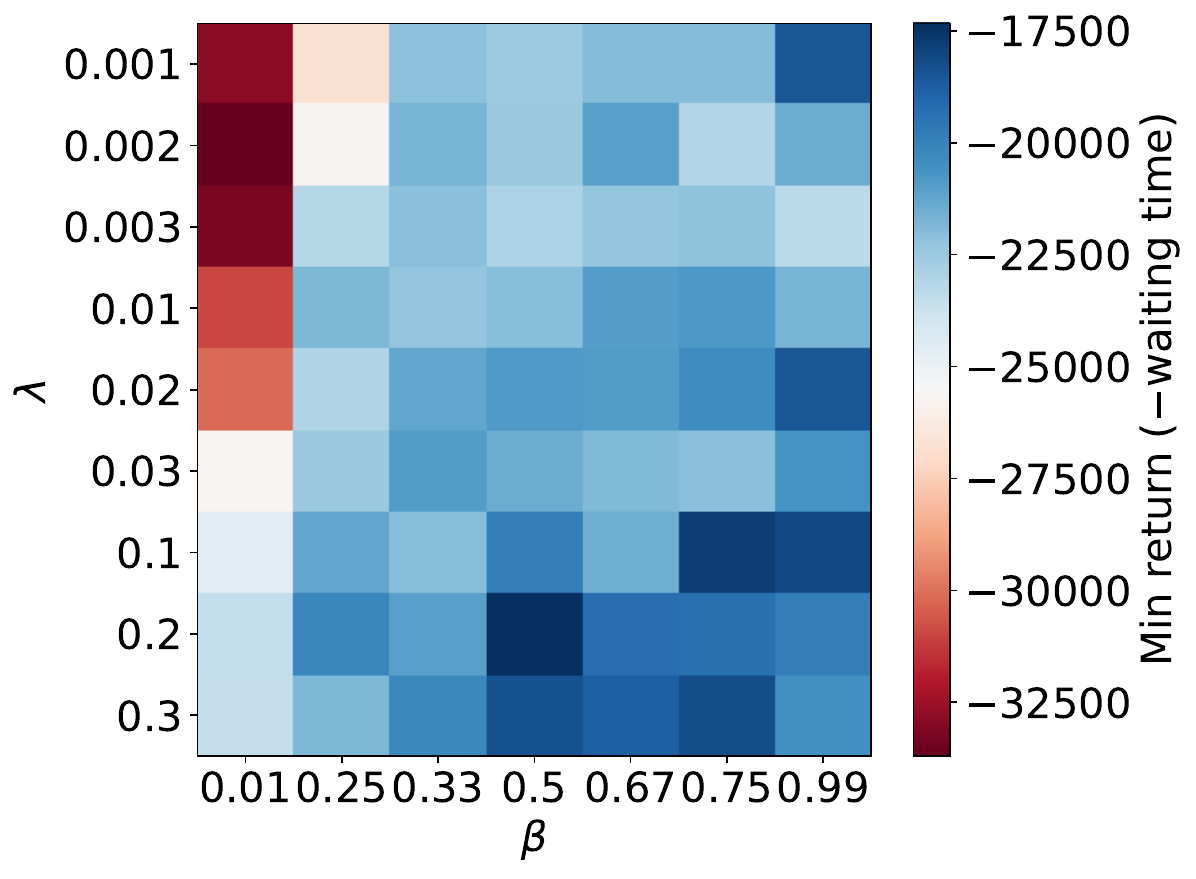}
    \caption*{ERAM in Asym-16}
  \end{subfigure}
  \hspace{1em}
  \begin{subfigure}[t]{0.45\textwidth}
    \includegraphics[width=\textwidth]{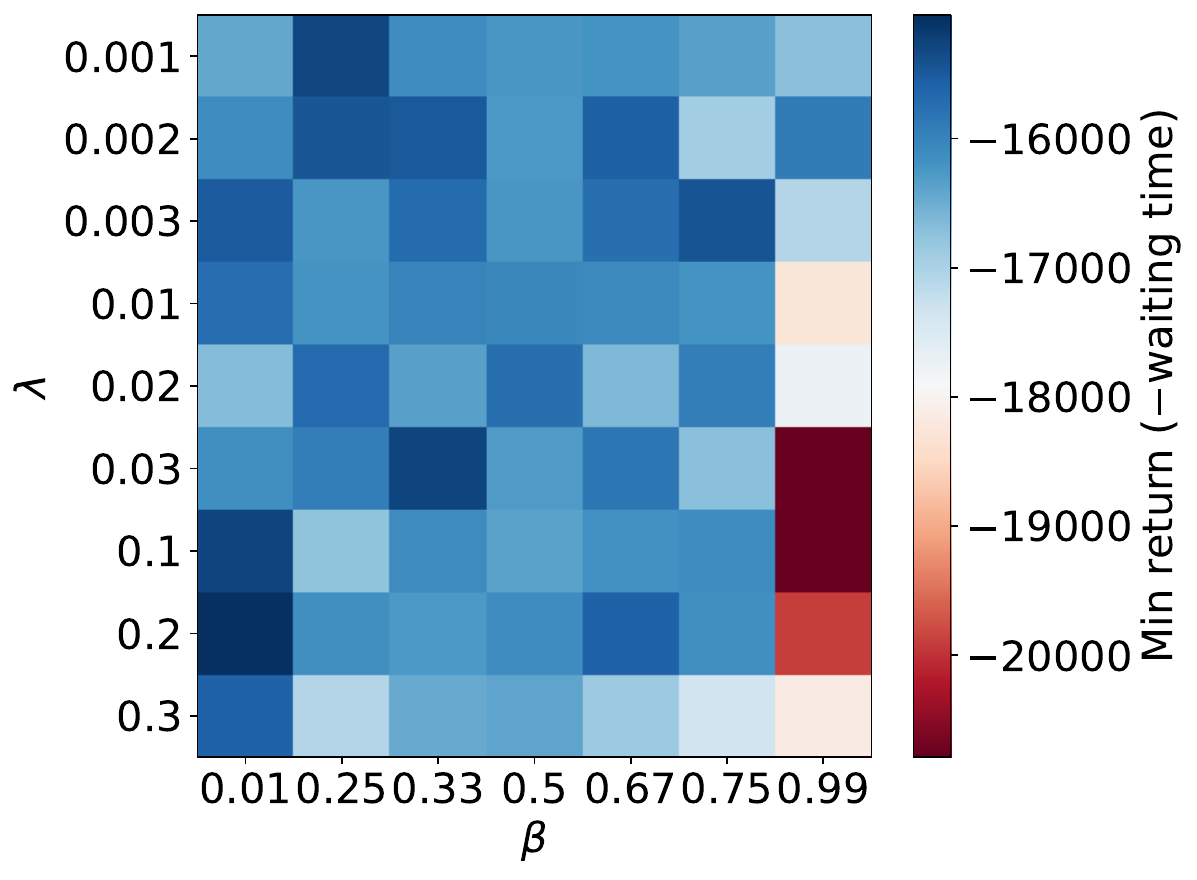}
    \caption*{ARAM in Asym-16}
  \end{subfigure}

  \caption{Ablation results on $\lambda$ and $\beta$ for ERAM and ARAM across traffic scenarios}
  \label{fig:ablation_star_heatmaps}
\end{figure}

As mentioned above, we conducted an ablation study on $\lambda$ and $\beta = \frac{1}{\lambda\tau_w + 1}$ (then, $\tau_w=(\frac{1}{\beta}-1)/\lambda$ is automatically determined), which are the main components of the closed-form update rules for $w$ in both ERAM~\eqref{eq:w_closedform} and ARAM~\eqref{eq:appendARAMwt}.
Figure~\ref{fig:ablation_star_heatmaps} presents the ablation study on $\lambda$ and $\beta$ for ERAM and ARAM across different traffic scenarios. Each heatmap visualizes the minimum return obtained for varying $(\lambda, \beta)$ combinations. In each heatmap, red indicates lower minimum return, while blue indicates higher minimum return. 

\newpage
\section{Additional Experimental Results}\label{append:additional_experiments}

\begin{table}[H]
\centering
\begin{tabular}{cccccccc}
\toprule
 & ARAM & ERAM & \citet{park} & GGF-PPO & GGF-DQN & Avg-DQN \\ 
 \midrule
Spec. Cons.    & 31 & 27 & 27 & 27 & 22 & 4 \\ 
MO-Reacher   & 25.27 & 25.13 & 23.54 & 24.32 & 23.90 & 22.44 \\
Four Room    & 1.80 & 1.56 & 1.02 & 1.47 & 0.02 & 0.12 \\
\bottomrule
\end{tabular}
\caption{Max-min performance in species conservation environment, MO-Reacher environment, and Four-room environment}
\label{tab:additional_exp}
\end{table}

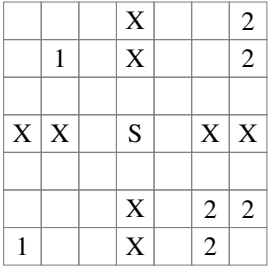
\begin{wrapfigure}{r}{0.4\textwidth}  
    \centering
\begin{tikzpicture}[scale=0.5]
    \foreach \x in {0,...,6} {
        \foreach \y in {0,...,6} {
            \draw[gray] (\x, \y) rectangle ++(1,1);
        }
    }

    \def\maze{
        { , , ,X, , ,2},
        { ,1, ,X, , ,2},
        { , , , , , , },
        {X,X, ,S, ,X,X},
        { , , , , , , },
        { , , ,X, ,2,2},
        {1, , ,X, ,2, },
    }

    \foreach \row [count=\i from 0] in \maze {
        \foreach \cell [count=\j from 0] in \row {
            \node at (\j+0.5, 6.5-\i) {\cell};
        }
    }
\end{tikzpicture}
    \caption{The map for Four-Room environment}
    \label{fig:fourroom_map}
\end{wrapfigure}

The Species Conservation environment (SC)~\cite{species_conservation}, a commonly used benchmark in MORL, aims to promote the conservation of species in an ecological simulation via multi-objective reinforcement learning. This environment includes two species: sea otters (an endangered predator) and northern abalone (their prey). The state space captures population information, and the action space consists of five discrete actions. The agent aims to achieve fair conservation of both species by treating their population levels as a vector-valued reward.

MO-Reacher environment (MR)~\cite{mo_gym} is a multi-objective extension of Reacher~\cite{gym}. This environment has a 6-dimensional observation space containing the sine and cosine values of the central and elbow joint angles, as well as their angular velocities. It has a discrete action space consisting of torques applied to the central and elbow joints, where each torque can take one of three values: $-1$, $0$, or $1$. As an extension of the standard Reacher environment, MO-Reacher includes four targets. The reward function is defined based on the distance between the tip of the arm and each target as follows: $r_i = 1 - 4\|\text{(tip's position)} - \text{(target $i$'s position)}\|^2,~i=1,2,3,4$.

The Four-Room (FR) environment~\cite{mo_gym} contains two types of collectible items, labeled 1 and 2. 
Figure~\ref{fig:fourroom_map} shows the map used in the Four-Room environment.
The agent starts at the center of the map, marked as "S". Cells marked "X" represent walls. Each episode terminates after a maximum of 200 steps. The agent collects items 1 and 2 throughout the episode, and the numbers of each collected item constitute the two-dimensional objective reward.

In all environments, we trained for a total of 100{,}000 timesteps. Table~\ref{tab:additional_exp} demonstrates the superior max-min performance of our algorithms in all environments.

\section{Limitations and Broader Impacts}\label{append:limitation}

The max-min criterion works best when objectives are homogeneous, such as having the same units. Extending it to heterogeneous objectives is an interesting direction for future work. While our theory focuses on softmax policy parameterization, analyzing last-iterate convergence under more general settings, such as linear function approximation, remains a valuable research direction.

Our work on max-min MORL may have several broader impacts. The proposed algorithm can be applied to real-world resource allocation problems where fairness or robustness across competing objectives is critical, such as transportation, healthcare, and public infrastructure. Moreover, the max-min criterion may contribute to more robust preference modeling in RL fine-tuning of large language models (LLMs), especially in the context of preference alignment. In such applications, max-min training can mitigate the influence of outlier preferences and help ensure consistency across diverse feedback signals. Because our algorithms are both memory-efficient and computation-efficient, it may be particularly suitable for training large-scale models under practical resource constraints. While our method is general and does not directly involve deployment, we note that any reinforcement learning system deployed in sensitive domains should be carefully audited for fairness, safety, and long-term behavior. We believe our approach supports these goals by improving robustness in multi-objective decision-making.

\end{document}